%% file: main_sisc_template.tex
\documentclass[review,hidelinks,onefignum,onetabnum]{siamart220329}

\usepackage{amssymb}
\usepackage{latexsym}
\usepackage{amstext,bbm,amsbsy} 
\usepackage{mathtools}
\usepackage{comment}
\usepackage{tabularx}
\usepackage{multirow}
\usepackage{url}

\usepackage{subcaption}

\usepackage{tikz}
\usetikzlibrary{shapes.geometric, arrows}
\usepackage{adjustbox}
\usepackage{siunitx}

\input{ex_shared}

\ifpdf
\hypersetup{
  pdftitle={Generalized Hopf Formula, Learning, and Optimal Control - LQR direction},
  pdfauthor={todo}
}
\fi

\definecolor{myorange}{rgb}{0.883, 0.473, 0.180}
\definecolor{myblue}{rgb}{0.277, 0.621, 0.969}
\definecolor{mygreen}{rgb}{0.324, 0.680, 0.230}

\newcommand{\be}{\begin{equation}}
\newcommand{\ee}{\end{equation}}
\newcommand{\benn}{\begin{equation*}}
\newcommand{\eenn}{\end{equation*}}

\newtheorem{prop}{Proposition}[section]

\newtheorem{example}{Example}[section]
\DeclareMathOperator*{\argmin}{arg\,min}
\DeclareMathOperator*{\argmax}{arg\,max}

\def\bx {\boldsymbol{x}}
\def\by {\boldsymbol{y}}
\def\bv {\boldsymbol{v}}
\def\bu {\boldsymbol{u}}

\def\bb {\boldsymbol{b}}
\def\ba {\boldsymbol{a}}
\def\bz {\boldsymbol{z}}

\def\R {\mathbb{R}}

\def\dom {\mathrm{dom~}}

\def\L {\mathcal{L}}

\def\Rn {\R^n}

\def\indexRiccati{j}
\def\pdhgdual{\boldsymbol{w}}

\def\HHJ {H}
\def\JHJ {J}

\newcommand{\Cpp}{C_{pp}} 
\newcommand{\Cxp}{C_{xp}} 
\newcommand{\Cxx}{C_{xx}} 
\newcommand{\Sxx}{P} 
\newcommand{\Sx}{\mathbf{q}} 
\newcommand{\Sc}{r} 

\newcommand{\update}[1]{\textcolor{red}{#1}} 

\newcommand{\updateone}[1]{\textcolor{black}{#1}}  
\newcommand{\updatetwo}[1]{\textcolor{black}{#1}} 
\newcommand{\updatethree}[1]{\textcolor{black}{#1}} 
\newcommand{\updateeditor}[1]{\textcolor{black}{#1}} 

\def\lossfunc{\mathcal{L}}

\def\weight{\theta}
\def\weightvec{{\boldsymbol{\weight}}}

\def\regfunc{R}
\def\param{\lambda}
\def\HJmom{\mathbf{p}}
\def\HJx{\mathbf{x}}
\def\HJt{t}
\def\HJu{\mathbf{u}}
\def\Hamiltonian{\HHJ}
\def\HJIC{\JHJ}
\def\LPinspace{\R^M}
\def\LPoutspace{\R^m}
\def\weightspace{\Rn}
\def\HJstatespace{\weightspace}
\def\numt{N}
\def\LQRxx{\mathcal{Q}}
\def\LQRuu{\mathcal{R}}
\def\LQRxu{\mathcal{N}}
\def\LQRTC{\mathcal{Q}_f}
\def\LQRA{A} 
\def\LQRB{B}
\def\MLregmat{\Gamma}
\def\MLreg{\gamma}    
\def\MLbasismat{\Phi}
\def\MLbasis{\phi}
\def\MLcentervec{\weightvec^0}
\def\MLcenter{\weight^0}
\def\MLcenternewvec{\tilde\weightvec^0}

\def\MLy{\by}

\begin{document}
\nolinenumbers
\renewcommand{\thefootnote}{\fnsymbol{footnote}}

\footnotetext[1]{Division of Applied Mathematics, Brown University, Providence, RI 02912, USA (paula\_chen@alumni.brown.edu, zongren\_zou@brown.edu, jerome\_darbon@brown.edu, george\_karniadakis@brown.edu).}
\footnotetext[2]{Department of Mathematics, UCLA, Los Angeles, CA 90025, USA (\email{tingwei@math.ucla.edu}).}
\footnotetext[3]{Pacific Northwest National Laboratory, Richland, WA 99354, USA}
\footnotetext[4]{Paula Chen, Tingwei Meng, and Zongren Zou contributed equally to this work.}
\footnotetext[5]{Corresponding author.}
\renewcommand{\thefootnote}{\arabic{footnote}}

\maketitle 

\begin{abstract}
Hamilton-Jacobi partial differential equations (HJ PDEs) have deep connections with a wide range of fields, including optimal control, differential games, and imaging sciences. By considering the time variable to be a higher dimensional quantity, HJ PDEs can be extended to the multi-time case. In this paper, we establish a novel theoretical connection between specific optimization problems arising in machine learning and the multi-time Hopf formula, which corresponds to a representation of the solution to certain multi-time HJ PDEs. Through this connection, we increase the interpretability of the training process of certain machine learning applications by showing that when we solve these learning problems, we also solve a multi-time HJ PDE and, by extension, its corresponding optimal control problem. As a first exploration of this connection, we develop the relation between the regularized linear regression problem and the Linear Quadratic Regulator (LQR). We then leverage our theoretical connection to adapt standard LQR solvers (namely, those based on the Riccati ordinary differential equations) to design new training approaches for machine learning. Finally, we provide some numerical examples that demonstrate the versatility and possible computational advantages of our Riccati-based approach in the context of continual learning, post-training calibration, transfer learning, and sparse dynamics identification. 
\end{abstract}



\tikzstyle{box} = [rectangle, rounded corners, minimum width=0cm, minimum height=1cm,text centered, draw=black, fill=none]
\tikzstyle{boxsmall} = [rectangle, rounded corners, minimum width=0cm, minimum height=0cm,text centered, draw=black, fill=none]
\tikzstyle{nobox} = [rectangle, rounded corners, minimum width=0cm, minimum height=1cm,text centered, draw=none, fill=none]
\tikzstyle{doublearrow} = [thick,<->,>=stealth]
\tikzstyle{dottedarrow} = [thick,<->,>=stealth,dotted]

\input{introduction}

\input{Hopf}

\input{LQR}

\input{methodology}

\section{Numerical examples}\label{sec:numerics}
\input{numerics}
\input{example_continual}
\input{example_calibration}

\input{example_poisson}
\input{example_sindy}

\input{conclusion}

\section*{Acknowledgments}
P.C. is supported by the SMART Scholarship, which is funded by the Under Secretary of Defense/Research and Engineering (USD/R\&E), National Defense Education Program (NDEP) / BA-1, Basic Research. J.D., G.E.K., and Z.Z. are supported by the MURI/AFOSR FA9550-20-1-0358 project. We also acknowledge the support by award DOE-MMICS SEA-CROGS DE-SC0023191.

\bibliographystyle{siamplain}
\bibliography{references}

\appendix
\input{appendix_short}

\end{document}

%% file: ex_shared.tex
\usepackage{lipsum}
\usepackage{epstopdf}
\usepackage{algorithmic}
\ifpdf
  \DeclareGraphicsExtensions{.eps,.pdf,.png,.jpg}
\else
  \DeclareGraphicsExtensions{.eps}
\fi

\newsiamremark{remark}{Remark}
\newsiamremark{hypothesis}{Hypothesis}
\crefname{hypothesis}{Hypothesis}{Hypotheses}
\newsiamthm{claim}{Claim}

\headers{Leveraging Multi-time HJ PDEs for Certain SciML Problems}{P. Chen, T. Meng, Z. Zou, J. Darbon, G.E. Karniadakis}

\title{Leveraging Multi-time Hamilton-Jacobi PDEs for Certain Scientific Machine Learning Problems}

\author{Paula Chen\footnotemark[1] \footnotemark[4] \and Tingwei Meng\footnotemark[2] \footnotemark[4] 
\and Zongren Zou\footnotemark[1] \footnotemark[4] 
\and J\'er\^ome Darbon\footnotemark[1] 
\footnotemark[5]
\and George Em Karniadakis\footnotemark[1] \footnotemark[3]
  }

\usepackage{amsopn}


%% file: introduction.tex
\section{Introduction}

It is well-known that Hamilton-Jacobi partial differential \linebreak equations (HJ PDEs) have deep connections to optimal control \cite{Bardi1997Optimal}, differential games \cite{evans1984differentialgames}, and imaging sciences \cite{darbon2015convex,darbon2022imagedenoising}, among many other fields. 
When the Hamiltonians are convex and only depend on the momentum, the solution to the HJ PDEs can be represented by a Hopf formula, which converts the solution of the PDE to the solution of an optimization problem. Multi-time HJ PDEs were originally introduced in economics~\cite{rochet1985multitimeHJ}. The solution to certain multi-time HJ PDEs was then shown to be able to be represented by a multi-time Hopf formula~\cite{lions1986hopf}, which is a generalization of the single-time case and has been shown to have connections with imaging sciences~\cite{darbon2019decomposition}. 

\begin{figure}[t!]
\begin{adjustbox}{width=\textwidth}
\begin{tikzpicture}[node distance=2cm]
    \node (min) [nobox, yshift=-0.2cm] {$\min$};
    \node (minarg) [boxsmall, right of=min, xshift=-1.5cm, yshift=-0.4cm, draw=cyan!60, fill=cyan!5] {$\text{}_\text{weights}$};
    \node (sum) [nobox, right of=min, xshift=-0.6cm] {$\sum$};
    \node (param) [box, right of=sum, xshift=-0.8cm, text width=1.6cm, draw=magenta!60, fill=magenta!5] {hyper- parameter};
    \node (loss) [box, right of=param, xshift=0.1cm, draw=green!60, fill=green!5, text width=1.7cm] {data fitting loss};
    \node (plus) [nobox, right of=loss, xshift=0.7cm] {$+$};
    \node (regLeft) [nobox, right of=loss, xshift=1cm] {};
    \node (regularization) [box, right of=regLeft, xshift=0.55cm, draw=blue!60, fill=blue!5] {regularization};
    \node (regRight) [nobox, right of=regularization, xshift=0.55cm] {};
    
    \node (sup) [nobox, below of=min, xshift=-0cm] {$\min$};
    \node (suparg) [boxsmall, right of=sup, xshift=-1.5cm, yshift=-0.37cm, draw=cyan!60, fill=cyan!5] {$\text{}_\text{momentum}$};
    \node (sum2) [nobox, below of=sum, xshift=0.2cm] {$\sum$};
    \node (time) [box, below of=param, xshift=-0cm, draw=magenta!60, fill=magenta!5] {time};
    \node (equals) [nobox, left of=sup, xshift=1.4cm] {$=$};
    \node (S) [box, left of=equals, xshift=0.6cm, text width=2cm, draw=red!60, fill=red!5] {solution to HJ PDE};
    \node (minus) [nobox, left of=S, xshift=0.65cm] {$-$};
    \node (Hamiltonian) [box, below of=loss, draw=green!60, fill=green!5] {Hamiltonian};
    \node (plus2) [nobox, right of=Hamiltonian, xshift=-0.55cm] {$+$};
    \node (IC) [box, below of=regLeft, text width=2cm, draw=blue!60, fill=blue!5] {Fenchel transform of initial data};
    \node (linear) [box, below of=regRight, draw=blue!60, fill=blue!5] {linear term};
    \node (plus3) [nobox, below of=regularization, xshift=-0cm] {$+$};

    \node (minloss) [box, above of=S, text width=2cm, draw=red!60, fill=red!5] {minimal loss value};
    \node (LPequals) [nobox, above of=equals] {$=$};
    
    \node (OCmin) [nobox, below of=sup] {$\min$};
    \node (OCminarg) [boxsmall, right of=OCmin, xshift=-1.5cm, yshift=-0.37cm, draw=cyan!60, fill=cyan!5] {$\text{}_\text{control}$};
    \node (OCS) [box, below of=S, xshift=0cm, draw=red!60, fill=red!5] {value function};
    \node (OCequals) [nobox, below of=equals, xshift=0cm] {$=$};
    \node (OCint) [nobox, below of=time, xshift=-0.5cm] {$\int$};
    \node (OCt) [boxsmall, right of=OCint, xshift=-1.5cm, yshift=-0.35cm, draw=magenta!60, fill=magenta!5] {time};
    \node (OCHamiltonian) [box, below of=Hamiltonian, draw=green!60, fill=green!5] {running cost};
    \node (OCds) [nobox, right of=OCHamiltonian, xshift=-0.7cm] {$ds$};
    \node (OCplus) [nobox, below of=plus2, xshift=0.2cm] {$+$};
    \node (OCIC) [box, below of=IC, draw=blue!60, fill=blue!5] {terminal cost};
    \node (OCdynam) [box, right of=OCIC, xshift=0.6cm, text width=2cm, draw=green!60, fill=green!5] {dynamics};
    \node (OCinitialposition) [box, below of=linear, text width=2cm, draw=blue!60, fill=blue!5] {initial position};
    \node (comma) [nobox, right of=OCdynam, xshift=-0.8cm] {,};
    \node (leftbracket) [nobox, left of=OCint, xshift=1.6cm] {$\Bigg\{$};
    \node (colon) [nobox, right of=OCIC, xshift=-0.75cm] {$:$};
    \node (rightbracket) [nobox, right of=OCinitialposition, xshift=-0.55cm] {$\Bigg\}$};
    
    \draw [doublearrow] (loss) -- (Hamiltonian);
    \draw [dottedarrow] (regularization) -- (IC);
    \draw [dottedarrow] (regularization) -- (linear);
    \draw [doublearrow] (minarg) -- (suparg);
    \draw [doublearrow] (OCS) -- (S);
    \draw [doublearrow] (OCt) -- (time);
    \draw [dottedarrow] (OCHamiltonian) -- (Hamiltonian);
    \draw [dottedarrow] (OCIC) -- (IC);
    \draw [dottedarrow] (OCdynam) -- (Hamiltonian);
    \draw [doublearrow] (time) -- (param);
    \draw [dottedarrow] (OCinitialposition) -- (linear);
    \draw [dottedarrow] (S) -- (minloss);
    \draw [dottedarrow] (suparg) -- (OCminarg);
\end{tikzpicture}
\end{adjustbox}

    \caption{(See Section~\ref{sec:Hopf}) Illustration of a connection between a regularized learning problem (\textbf{top}), the multi-time Hopf formula (\textbf{middle}), and an optimal control problem (\textbf{bottom}). The colors indicate the associated quantities between each problem. For example, the optimal weights in the learning problem are equivalent to the momentum in the HJ PDE, which is related to the control in the optimal control problem (\textcolor{cyan}{cyan}), and the hyper-parameters for the data fitting terms correspond to the time variables in the HJ PDE and the time horizon in the optimal control problem (\textcolor{magenta}{magenta}). This color scheme is reused in the subsequent illustrations of our connection. The solid-line arrows denote direct equivalences. The dotted arrows represent additional mathematical relations.}
    \label{fig:intro_connection_in_words}
\end{figure}
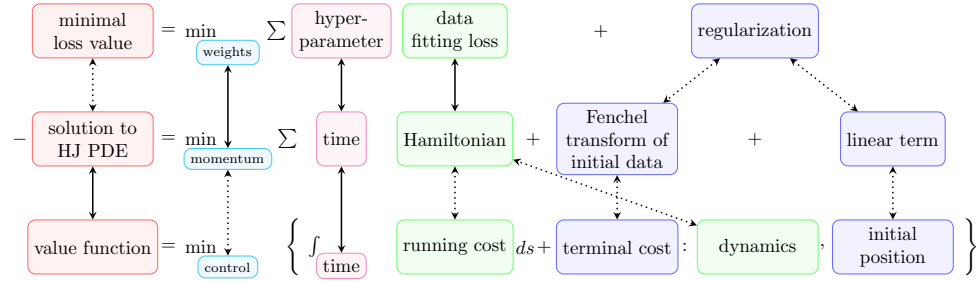

In this paper, we establish a novel theoretical connection between certain optimization problems arising in machine learning and the multi-time Hopf formula (Section~\ref{sec:Hopf}). Specifically, we show that there are one-to-one correspondences between the loss functions in \updateone{regularized} learning problems and the objective function of the multi-time Hopf formula, which in turn yields connections to optimal control. See Figure~\ref{fig:intro_connection_in_words} for an illustration of these correspondences. As such, our connection increases the interpretability of the training process of certain machine learning applications by showing that when we solve these learning problems, we actually solve a multi-time HJ PDE and by extension, its corresponding optimal control problem. In this paper, we show that our connection allows us to leverage HJ PDE and optimal control theory and algorithms to solve optimization problems arising from machine learning applications, and we reserve the reverse direction for future work.

As a first exploration of this connection, we focus on the connection between regularized linear regression problems and the Linear Quadratic Regulator (LQR) (Section~\ref{sec:LQR}). We use this case to gain a deeper understanding of our connection and illustrate its potential benefits. Linear regression consists of learning a linear prediction model and is one of the fundamental learning problems in supervised machine learning \cite{russell2010artificial, mohri2018foundations, weisberg2005applied}. LQR \cite{TrentelmanControlLinearSystems, anderson2007optimal} is a well-studied optimal control problem with quadratic running and terminal costs and linear dynamics and is typically solved using the Riccati ordinary differential equations (ODEs) \cite{mceneaney2006max,Darbon2023Neural,Nakamura2021QRnet}. Through our connection, we establish that solving regularized linear regression problems is equivalent to solving particular LQR problems. As a result, we can leverage standard LQR solvers to design new training approaches for machine learning. In particular, we develop new methodology for solving regularized linear regression problems by adapting solvers for the Riccati ODEs to these new settings (Section~\ref{sec:method}).

To highlight the versatility and possible computational advantages of our new Riccati-based approach, we apply our methodology to several test problems in machine learning (Section~\ref{sec:numerics}).
In the first example, we consider a function approximation problem to demonstrate the computational and memory advantages of our Riccati-based approach in the context of continual learning \cite{parisi2019continual, kirkpatrick2017overcoming, van2019three}. Specifically, we show that our Riccati-based approach naturally enables us to continually adapt the learned model to new data without having to store or retrain on the previous data (which is especially significant given the rise of big data), while also avoiding catastrophic forgetting~\cite{kirkpatrick2017overcoming, parisi2019continual}. 
In the second example, we demonstrate how our Riccati-based approach can be used to perform post-training calibrations. In particular, our approach gives us the flexibility to add or remove data points and tune hyper-parameters to increase the accuracy of the learned model without having to retrain it entirely, which again provides computational and memory advantages over conventional learning methods. 
In the third example, we use our Riccati-based methodology to fit the last layer of a physics-informed neural network (PINN) \cite{raissi2019physics} using transfer learning \cite{zou2023hydra, desai2021one}. In this application, we demonstrate that as we change the value of the hyper-parameters, solving the associated Riccati ODEs not only provides the solution to the updated problem, but also a \textit{continuum} of solutions along a 1D curve on the Pareto front of the data fitting losses and regularization. 
Finally, in the fourth example, we highlight the versatility of our Riccati-based approach by showing how it can be combined with existing optimization methods (e.g., the primal-dual hybrid gradient (PDHG) algorithm \cite{chambolle2011pdhg}) to perform sparse dynamics identification \cite{brunton2016discovering}.

\begin{figure}[t!]
\centering
\includegraphics[width = 0.8\textwidth]{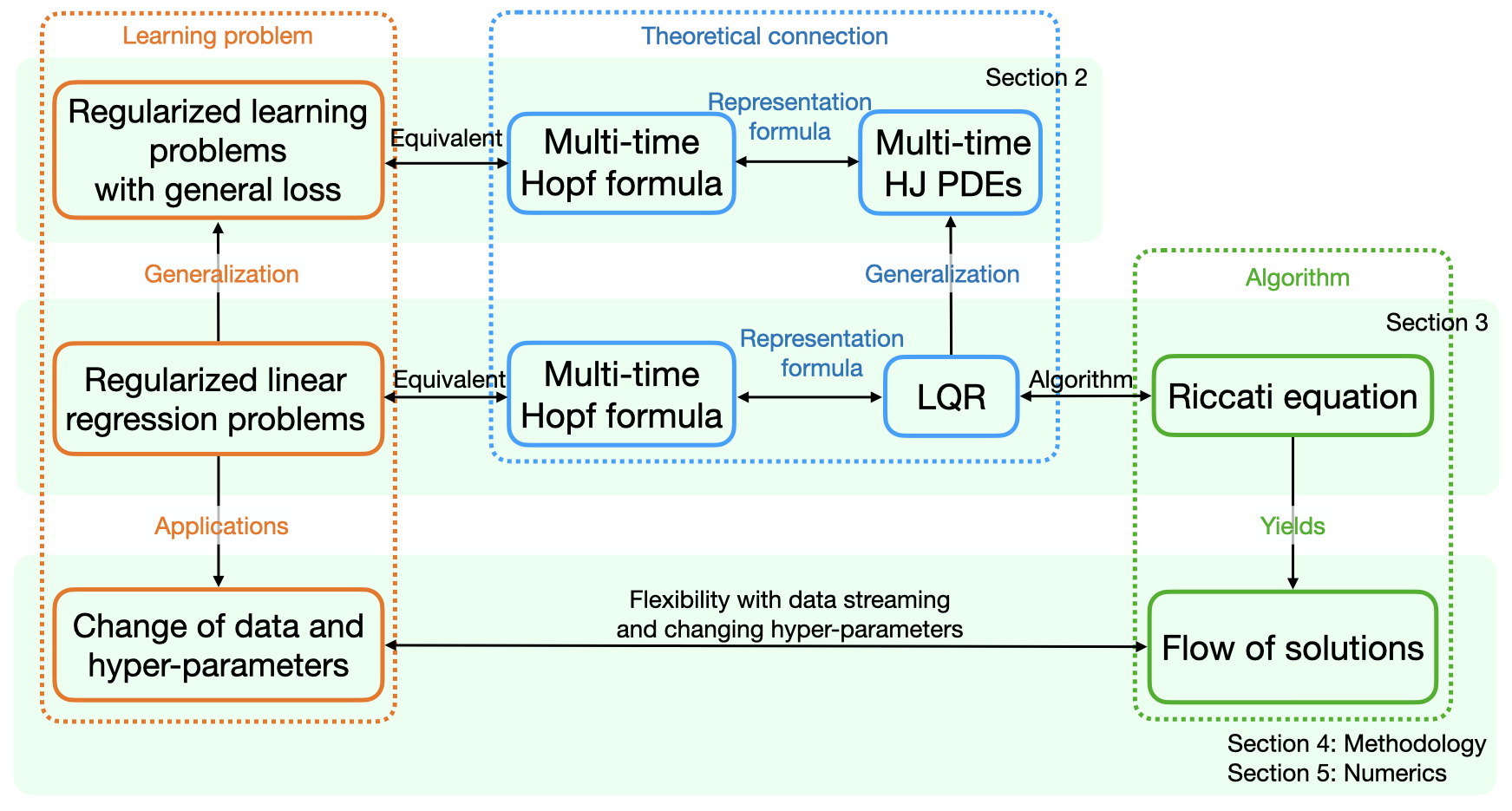}
\caption{
Overview of the overall structure of this paper, the learning problems considered, our new theoretical connection, and our new Riccati-based algorithm. We show the learning problem in the \textbf{left} (\textcolor{myorange}{orange}), the theoretical connection in the \textbf{middle} (\textcolor{myblue}{blue}), and the algorithm in the \textbf{right} (\textcolor{mygreen}{green}). The \textbf{top} row contains the general connection between learning problems and HJ PDEs (see Section~\ref{sec:Hopf}), the \textbf{middle} row contains the connection between the regularized linear regression problem and LQR (see Section~\ref{sec:LQR}), and the \textbf{bottom} row contains the connection between applications in machine learning and our Riccati-based algorithm (see Sections~\ref{sec:method} and~\ref{sec:numerics}).
}
\label{fig:illustration}
\end{figure}

The main contributions of this work are the development of a new theoretical connection between learning problems and the Hopf formula and a new Riccati-based approach for solving regularized linear regression problems.
While we demonstrate promising results, the work presented here has some limitations. For example, while we establish our theoretical connection between more general learning problems, multi-time HJ PDEs, and optimal control problems, we have yet to fully explore non-linear learning models, general convex Hamiltonians, or non-linear control dynamics. Additionally, as discussed previously, we have also not yet investigated what possible advantages our  connection provides for solving HJ PDEs and optimal control problems. In particular, many efficient solvers for high-dimensional problems in machine learning exist~\cite{lecun2015deep, goodfellow2016deep}; it would be desirable to be able to leverage our connection to reuse this machinery for HJ PDEs and optimal control. Thus, our novel connection presents many exciting opportunities.
We discuss some other possible future directions in Section~\ref{sec:conclusion}.
The overall structure of this paper is illustrated in Figure~\ref{fig:illustration}.

%% file: Hopf.tex
\section{Generalized Hopf formula}\label{sec:Hopf}
In this section, we provide some mathematical background on the single- and multi-time Hopf formulas. Specifically, we review the well-known connections between the Hopf formula, the solution to HJ PDEs with time- and space-independent Hamiltonians, and the solution to the corresponding optimal control problems. We then present a novel theoretical connection between the Hopf formula and \updateone{regularized} learning problems. Through this connection, we establish that when we solve these learning problems, we actually evaluate the solution to certain HJ PDEs and their corresponding optimal control problems, and vice versa.

\subsection{Introduction to the Hopf formula}

The single-time HJ PDE is 
\begin{equation}\label{eqt:singletimeHJPDE}
    \begin{dcases}
    \frac{\partial S(\HJx,\HJt)}{\partial \HJt} + \Hamiltonian(\nabla_\HJx S(\HJx,\HJt)) = 0 & \HJx\in \HJstatespace, \HJt > 0, \\
    S(\HJx,0) = \HJIC(\HJx) & \HJx\in\HJstatespace,
    \end{dcases}
\end{equation}
where $\Hamiltonian:\Rn\to \R$ is the Hamiltonian and $\HJIC:\Rn\to \R$ is the initial condition. Assume that $\Hamiltonian$ and $\HJIC$ are convex (although we note that these assumptions could be relaxed; e.g., see \cite{hopf1965hopfformula}). Then, the viscosity solution to the single-time HJ PDE~\eqref{eqt:singletimeHJPDE} is given by the Hopf formula \cite{hopf1965hopfformula}:
\begin{equation}\label{eqt:singletime_Hopf}
\begin{aligned}
S(\HJx,\HJt) & = \sup_{\HJmom\in\HJstatespace} \{\langle \HJx,\HJmom\rangle - \HJt \Hamiltonian(\HJmom) - \HJIC^*(\HJmom)\} = -\inf_{\HJmom\in\HJstatespace} \{ \HJt \Hamiltonian(\HJmom) + \HJIC^*(\HJmom)-\langle \HJx,\HJmom\rangle\},
\end{aligned}
\end{equation}
where $f^*$ denotes the Fenchel-Legendre transform of the function $f$; i.e., $f^*(p) = \sup_{x\in\Rn} \{\langle x,p\rangle - f(x)\}$.

\updatetwo{The value function of the following optimal control problem also solves~\eqref{eqt:singletimeHJPDE}}: 
\begin{equation}\label{eqt:optimal_control_standardform}
S(\HJx,\HJt)=\min_{\HJu(\cdot)} \left\{\int_{0}^{\HJt}L(\HJu(s)) ds + \HJIC(\bx(\HJt))\colon \dot\bx(s) = f(\HJu(s)) \forall s\in(0,\HJt], \bx(0) = \HJx\right\},
\end{equation}
where the running cost $L$ and the source term $f$ of the dynamics are related to the Hamiltonian $\Hamiltonian$ by $\Hamiltonian(\HJmom) = \sup_{\HJu\in\R^m} \{\langle -f(\HJu), \HJmom\rangle - L(\HJu)\}$ and, in this context, we interpret $\HJIC$ to be the terminal cost.

A natural generalization of this formulation to the multi-time case is as follows. Let $\Hamiltonian_1,\dots, \Hamiltonian_\numt$ be convex Hamiltonians, such that $\dom \Hamiltonian_i = \Rn$ for all $i = 1, \dots, \numt$, and let $\HJIC$ be a convex initial condition. (Again, we note that these assumptions could be relaxed; e.g., see \cite{lions1986hopf}.) Then, the multi-time HJ PDE is given by
\begin{equation}\label{eqt:multitimeHJPDE}
\begin{adjustbox}{width=\textwidth}
    $\begin{dcases}
     \frac{\partial S(\HJx,\updatetwo{\HJt_1, \dots, \HJt_\numt} )}{\partial \HJt_i} + \Hamiltonian_i(\nabla_\HJx S(\HJx,\updatetwo{\HJt_1, \dots, \HJt_\numt })) = 0 \text{ for } i\in\{1, \dots, \numt\} & \HJx\in \HJstatespace, \HJt_1, \dots, \HJt_\numt > 0, \\
    S(\HJx,0, \dots, 0) = \HJIC(\HJx) & \HJx\in\HJstatespace,
    \end{dcases}$
    \end{adjustbox}
\end{equation}
and the solution to the multi-time HJ PDE~\eqref{eqt:multitimeHJPDE} can be represented by the following generalized (multi-time) Hopf formula \cite{lions1986hopf}:
\begin{equation} \label{eqt:multitime_Hopf}
\begin{aligned}
S(\HJx,\HJt_1,\dots, \HJt_\numt) & = \sup_{\HJmom\in\HJstatespace} \left\{\langle \HJx, \HJmom\rangle - \sum_{i=1}^\numt \HJt_i\Hamiltonian_i(\HJmom) - \HJIC^*(\HJmom)\right\} \\
&= -\inf_{\HJmom\in\HJstatespace} \left\{ \sum_{i=1}^\numt \HJt_i\Hamiltonian_i(\HJmom) + \HJIC^*(\HJmom) -\langle \HJx, \HJmom\rangle \right\}.\\
\end{aligned}
\end{equation}

\updatetwo{Moreover, the value function of an optimal control problem in the form of~\eqref{eqt:optimal_control_standardform}}
with terminal time $\HJt = \sum_{j=1}^\numt \HJt_j$ and where the running cost $L$ and the source term $f$ of the dynamics are defined piecewise by $L(s, \HJu) = L_i(\HJu)$ and $f(s, \HJu) = f_i(\HJu)$, respectively, for $s \in \left(\sum_{j=1}^{i-1}\HJt_j, \sum_{j=1}^{i}\HJt_j\right]$ and $i= 1, \dots, \numt$ \updatetwo{also solves the HJ PDE~\eqref{eqt:multitimeHJPDE}}. The piecewise running costs $L_i$ and piecewise source terms $f_i$ of the dynamics are related to the Hamiltonians $\Hamiltonian_i$ by $\Hamiltonian_i(\HJmom) = \sup_{\HJu\in\R^m} \{\langle -f_i(\HJu), \HJmom\rangle - L_i(\HJu)\}$.

\subsection{Connection between the Hopf formula and learning problems}\label{sec:general_connection_Hopf}

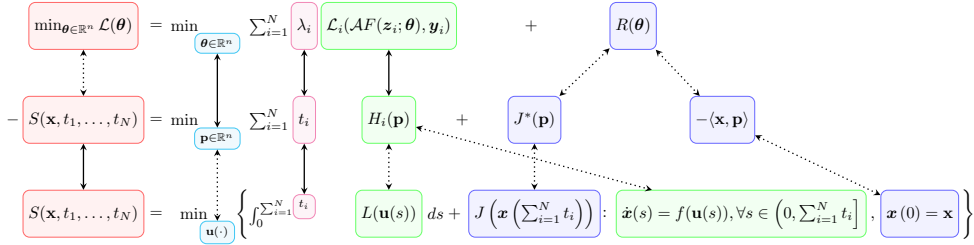
\begin{figure}[htbp]

    \centering   
    \begin{adjustbox}{width=\textwidth}
\begin{tikzpicture}[node distance=2cm]
    \node (min) [nobox, yshift=-0.2cm] {$\min$};
    \node (minarg) [boxsmall, right of=min, xshift=-1.3cm, yshift=-0.37cm, draw=cyan!60, fill=cyan!5] {$\text{}_{\weightvec\in\weightspace}$};
    \node (sum) [nobox, right of=min, xshift=-0.2cm] {$\sum_{i=1}^\numt$};
    \node (param) [box, right of=sum, xshift=-1.25cm, draw=magenta!60, fill=magenta!5]{$\param_i$};
    \node (loss) [box, right of=param, xshift=-0.2cm, draw=green!60, fill=green!5] {$\lossfunc_i(\mathcal{A}F(\bz_i;\weightvec), \by_i)$};
    \node (plus) [nobox, right of=loss, xshift=1.03cm] {$+$};
    \node (regLeft) [nobox, right of=loss, xshift=1.1cm] {};
    \node (regularization) [box, right of=regLeft, xshift=0.1cm, draw=blue!60, fill=blue!5] {$\regfunc(\weightvec)$};
    \node (regRight) [nobox, right of=regularization, xshift=-0.1cm] {};
    
    \node (sup) [nobox, below of=min, xshift=-0cm] {$\min$};
    \node (suparg) [boxsmall, right of=sup, xshift=-1.3cm, yshift=-0.4cm, draw=cyan!60, fill=cyan!5] {$\text{}_{\HJmom\in\HJstatespace}$};
    \node (sum2) [nobox, below of=sum] {$\sum_{i=1}^\numt$};
    \node (equal) [nobox, left of=sup, xshift=1.4cm] {$=$};
    \node (S) [box, left of=equal, xshift=0.45cm, draw=red!60, fill=red!5] {$S(\HJx, \HJt_1, \dots, \HJt_N)$};
    \node (minus) [nobox, left of=S, xshift=0.5cm] {$-$};
    \node (time) [box, below of=param, draw=magenta!60, fill=magenta!5]{$\HJt_i$};
    \node (Hamiltonian) [box, below of=loss, draw=green!60, fill=green!5] {$\Hamiltonian_i(\HJmom)$};
    \node (plus2) [nobox, right of=Hamiltonian, xshift=-0.44cm] {$+$};
    \node (IC) [box, below of=regLeft, draw=blue!60, fill=blue!5] {$\HJIC^*(\HJmom)$};
    \node (linear) [box, below of=regRight, draw=blue!60, fill=blue!5] {$-\langle \HJx, \HJmom\rangle$};

    \node (OCmin) [nobox, below of=sup, xshift=0.2cm] {$\min$};
    \node (OCminarg) [boxsmall, below of=suparg, yshift=-0cm, draw=cyan!60, fill=cyan!5] {$\text{}_{\HJu(\cdot)}$};
    \node (OCS) [box, below of=S, xshift=0cm, draw=red!60, fill=red!5] {$S(\HJx, \HJt_1, \dots, \HJt_N)$};
    \node (leftbracket) [nobox, right of=OCmin, xshift=-0.97cm] {$\Bigg\{$};
    \node (OCequal) [nobox, below of=equal] {$=$};
    \node (OCint) [nobox, below of=sum2, xshift=0.05cm] {$\int_{0}^{\sum_{i=1}^{\numt}}$};
    \node (OCintupperbd) [boxsmall, below of=time, yshift=0.15cm, draw=magenta!60, fill=magenta!5] {${}^{\HJt_i}$};
    \node (OCHamiltonian) [box, below of=Hamiltonian, draw=green!60, fill=green!5] {$L(\HJu(s)) $};
    \node (OCds) [nobox, right of=OCHamiltonian, xshift=-1cm] {$ds$};
    \node (OCplus) [nobox, right of=OCds, xshift=-1.6cm] {$+$};
    \node (OCIC) [box, below of=IC, draw=blue!60, fill=blue!5] {$\HJIC\left(\bx\left(\sum_{i=1}^{\numt}\HJt_i\right)\right)$};
    \node (colon) [nobox, right of=OCIC, xshift=-0.5cm] {:};
    \node(OCdynamics) [box, right of=OCIC, xshift=2.4cm, draw=green!60, fill=green!5] {$\dot\bx(s) = f(\HJu(s)), \forall s \in \left(0, \sum_{i=1}^{\numt}\HJt_i\right]$};
    \node(comma) [nobox, right of=OCdynamics, xshift=0.8cm, yshift=-0.11cm] {,};
    \node (terminalposition) [box, right of=OCdynamics, xshift=1.8cm, draw=blue!60, fill=blue!5] {$\bx\left(0\right) = \HJx$};
    \node (rightbracket) [nobox, right of=terminalposition, xshift=-0.97cm] {$\Bigg\}$};

    \node (LPequal) [nobox, above of=equal] {$=$};
    \node (minloss) [box, above of=S, draw=red!60, fill=red!5] {$\min_{\weightvec\in\weightspace}\mathcal{L}(\weightvec)$};
    
    \draw [doublearrow] (param) -- (time);
    \draw [doublearrow] (loss) -- (Hamiltonian);
    \draw [dottedarrow] (regularization) -- (IC);
    \draw [dottedarrow] (regularization) -- (linear);
    \draw [doublearrow] (minarg) -- (suparg);
    \draw [doublearrow] (S) -- (OCS);
    \draw [doublearrow] (time) -- (OCintupperbd);
    \draw [dottedarrow] (IC) -- (OCIC);
    \draw [dottedarrow] (Hamiltonian) -- (OCHamiltonian);
    \draw [dottedarrow] (Hamiltonian) -- (OCdynamics);
    \draw [dottedarrow] (linear) -- (terminalposition);
    \draw[dottedarrow](minloss) -- (S);
    \draw[dottedarrow] (suparg) -- (OCminarg);
\end{tikzpicture}
\end{adjustbox}

    \caption{(See Section~\ref{sec:Hopf}) Mathematical formulation describing the connection between a regularized learning problem (\textbf{top}), the multi-time Hopf formula (\textbf{middle}), and an optimal control problem (\textbf{bottom}). The content of this illustration matches that of Figure~\ref{fig:intro_connection_in_words} by replacing each term in Figure~\ref{fig:intro_connection_in_words} with its corresponding mathematical expression. The colors indicate the associated quantities between each problem. The solid-line arrows denote direct equivalences. The dotted arrows represent additional mathematical relations.}
    \label{fig:connection_multitimeHopf}
\end{figure}

In this section, we connect the Hopf formulas~\eqref{eqt:singletime_Hopf} and~\eqref{eqt:multitime_Hopf} with \updateone{regularized} learning problems. Consider a learning problem with data points $\{(\bz_i,\by_i)\}_{i=1}^\numt \subset \LPinspace\times \LPoutspace$. The goal of the learning problem is to find a function $F(\bz;\weightvec)$ with input $\bz\in\LPinspace$ and unknown parameter $\weightvec\in\weightspace$, such that $\mathcal{A}F(\bz_i;\weightvec)$ approximately equals $\by_i$ at every $\bz_i$. Here $\mathcal{A}$ is an operator acting on the function $F$. 
For instance, $\mathcal{A}$ could be the identity operator (as in regression problems \cite{weisberg2005applied}) or a differential operator (as in PINNs \cite{raissi2019physics}).
Then, the learning problem is given by the following optimization problem:
\begin{equation}\label{eqt:general_learning}
\min_{\weightvec\in \weightspace} \sum_{i=1}^\numt \lambda_i\lossfunc_i(\mathcal{A}F(\bz_i;\weightvec), \by_i) + \regfunc(\weightvec).
\end{equation}
The above loss function consists of two parts: each  $\lossfunc_i(\mathcal{A}F(\bz_i;\weightvec), \by_i)$ is a data fitting term at $(\bz_i, \by_i)$ (where $\lossfunc_i(\ba,\bb)$ is a function measuring the discrepancy between $\ba$ and $\bb$) and $\regfunc$ is a regularization term. In this paper, we assume the functions $\weightvec \mapsto \lossfunc_i(\mathcal{A}F(\bz_i;\weightvec), \by_i)$ are convex for all $i=1,\dots, \numt$.

Then, the connection between the learning problem~\eqref{eqt:general_learning}, the Hopf formulas~\eqref{eqt:singletime_Hopf} and~\eqref{eqt:multitime_Hopf}, and the optimal control problem~\eqref{eqt:optimal_control_standardform} is illustrated in Figure~\ref{fig:connection_multitimeHopf}. 
Specifically, if there is only one data point ($\numt =1$), the learning problem~\eqref{eqt:general_learning} is related to the single-time Hopf formula~\eqref{eqt:singletime_Hopf} by setting \updateone{$\weightvec = \HJmom$}, $\Hamiltonian(\HJmom) = \lossfunc_1(\mathcal{A}F(\bz_1;\HJmom), \by_1)$, $\HJt = \param_1$, and $\regfunc(\HJmom) = \HJIC^*(\HJmom) - \langle \HJx,\HJmom\rangle + c(\HJx)$, where $c(\HJx)$ is a constant (possibly 0) that is independent of $\HJmom$ but may depend on $\HJx$. \updateone{In other words, the variables $\HJx, \HJt$ in the HJ PDE become hyper-parameters in the learning problem, and we can treat them as constants when optimizing the learning problem~\eqref{eqt:general_learning} with respect to $\weightvec = \HJmom$.} Hence, when we solve these single-point learning problems, we simultaneously evaluate the solution to the HJ PDE~\eqref{eqt:singletimeHJPDE} at the point $(\HJx,\HJt)$, or equivalently, we solve the corresponding optimal control problem~\eqref{eqt:optimal_control_standardform}. Conversely, when we solve the HJ PDE~\eqref{eqt:singletimeHJPDE}, the spatial gradient $\nabla_\HJx S(\HJx, \HJt)$ of the solution gives the minimizer $\weightvec^*$ to the single-point learning problem~\eqref{eqt:general_learning}.

If there are $\numt > 1$ data points, then the learning problem~\eqref{eqt:general_learning} is related to the multi-time Hopf formula~\eqref{eqt:multitime_Hopf} by setting \updateone{$\weightvec = \HJmom$}, $\Hamiltonian_i(\HJmom) = \lossfunc_i(\mathcal{A}F(\bz_i;\HJmom), \by_i)$, $\HJt_i = \param_i$, and $\regfunc(\HJmom) = \HJIC^*(\HJmom) - \langle \HJx,\HJmom\rangle+ c(\HJx)$, where $c(\HJx)$ is a constant (possibly 0) that is independent of $\HJmom$ but may depend on $\HJx$. 
Hence, solving these multi-point learning problems is equivalent to evaluating the solution to the HJ PDE~\eqref{eqt:multitime_Hopf} at the point $(\HJx,\HJt_1, \dots, \HJt_\numt)$, which is also equivalent to solving the corresponding optimal control problem~\eqref{eqt:optimal_control_standardform} with terminal time $\HJt = \sum_{j=1}^\numt \HJt_j$ and piecewise running costs $L_i$ and dynamics $f_i$ related via $\Hamiltonian_i(\HJmom) = \sup_{\HJu\in\R^m} \{\langle -f_i(\HJu), \HJmom\rangle - L_i(\HJu)\}$. Similarly to the single-time case, when we solve the multi-time HJ PDE~\eqref{eqt:multitimeHJPDE}, the spatial gradient $\nabla_\HJx S(\HJx, \HJt_1, \dots, \HJt_\numt)$ of the solution gives the minimizer $\weightvec^*$ to the multi-point learning problem~\eqref{eqt:general_learning}. \updateone{We provide a glossary of the mathematical relationships between these three problem formulations in Appendix~\ref{appendix:glossary}.}

%% file: LQR.tex
\section{Linear Quadratic Regulator}\label{sec:LQR}
In this section, we develop our theoretical connection from Section~\ref{sec:general_connection_Hopf} in the specific case where the optimal control problem~\eqref{eqt:optimal_control_standardform} corresponds to the LQR problem  \cite{TrentelmanControlLinearSystems, anderson2007optimal}. We show that solving certain LQR problems is equivalent to solving learning problems with linear models, quadratic data fitting losses, and quadratic regularization (i.e., an $\ell_2$-regularized linear regression problem). Although broader classes of learning problems are of interest, we restrict to linear regression problems as a starting point for demonstrating the potential of our theoretical connection. Specifically, we leverage our theoretical connection to show how established techniques for solving HJ PDEs (e.g., the Riccati ODEs \cite{mceneaney2006max}) can be reused to solve this class of learning problems. 

\subsection{Introduction to the Linear Quadratic Regulator and Riccati equation}\label{sec:intro_LQR}

The finite-horizon, continuous-time LQR is given by
\begin{multline}\label{eqt:LQR_general_control}
    S(\HJx,\HJt) = \min_{\updateone{\HJu(\cdot)}} \Bigg\{\int_0^\HJt \left(\frac{1}{2}\bx(s)^T\LQRxx\bx(s) + \frac{1}{2}\HJu(s)^T\LQRuu\HJu(s) + \bx(s)^T\LQRxu\HJu(s)\right) ds \\ + \frac{1}{2}\bx(\HJt)^T\LQRTC\bx(\HJt): 
    \dot\bx(s) = \LQRA\bx(s) + \LQRB\HJu(s) \forall s\in(0,t], \bx(0) = \HJx\Bigg\},
\end{multline}
where $\LQRxx, \LQRTC \in\R^{n\times n}$ and $\LQRuu\in\R^{m\times m}$ are symmetric positive definite, $\LQRxu\in\R^{n\times m}$, $\LQRA\in\R^{n\times n}$, and $\LQRB\in\R^{n\times m}$. The corresponding HJ PDE is 
\begin{equation}\label{eqt:LQR_HJPDE}
    \begin{dcases}
    \frac{\partial S(\HJx,\HJt)}{\partial \HJt} + \Hamiltonian(\HJx,\nabla_\HJx S(\HJx,\HJt)) = 0 & \HJx\in\R^n, \HJt > 0, \\
    S(\HJx,0) = \HJIC(\HJx)  & \HJx\in\R^n,
    \end{dcases}
\end{equation}
where the initial data of the HJ PDE is given by the terminal cost $\HJIC(\HJx) := \frac{1}{2}\HJx^T\LQRTC\HJx$ of the optimal control problem and the Hamiltonian $\Hamiltonian$ is defined by
\begin{equation}
    \begin{aligned}
    \Hamiltonian(\HJx,\HJmom) & = \sup_{\HJu\in\R^m} \langle -f(\HJx,\HJu), \HJmom \rangle - L(\HJx,\HJu) \\
     &=  -\langle \LQRA\HJx, \HJmom\rangle - \frac{1}{2}\langle \HJx, \LQRxx\HJx\rangle + \frac{1}{2}\langle \LQRB^T\HJmom + \LQRxu^T\HJx, \LQRuu^{-1}(\LQRB^T\HJmom + \LQRxu^T\HJx) \rangle,
    \end{aligned}
\end{equation}
where $f(\HJx,\HJu) = \LQRA\HJx + \LQRB\HJu$ is the source term of the dynamics and $L(\HJx,\HJu) = \frac{1}{2}\HJx^T\LQRxx\HJx + \frac{1}{2}\HJu^T\LQRuu\HJu + \HJx^T\LQRxu\HJu$ is the running cost. Note that because of the spatial dependence in the Hamiltonian $\Hamiltonian$, the Hopf formula cannot be applied directly to the above LQR problem without additional assumptions. In Sections~\ref{sec:LQR_1ptregression} and~\ref{sec:LQR_multiptregression}, we discuss some assumptions under which $\Hamiltonian$ is independent of the spatial variable $\HJx$ and the corresponding learning problems that can be solved via our connection through the Hopf formula (see, for example, Figure~\ref{fig:connection_LQR}).

It is well-known that this LQR problem can be solved using the Riccati equation as follows \cite{mceneaney2006max}. Define $\Cpp = \LQRB \LQRuu^{-1}\LQRB^T$, $\Cxx = -\LQRxu\LQRuu^{-1}\LQRxu^T + \LQRxx$, and $\Cxp = \LQRA - \LQRB \LQRuu^{-1}\LQRxu^T$. 
Then, the solution is also given by
$S(\HJx,\HJt) = \frac{1}{2} \HJx^T \Sxx(\HJt)\HJx$, 
where the function $\Sxx: [0,\infty)\to\R^{n\times n}$ takes values in the space of symmetric positive definite matrices and solves the following Riccati equation:
\begin{equation} \label{eqt: odeP}
{\small
    \begin{dcases}
    \dot{\Sxx}(\HJt) =  -\Sxx(\HJt)^T\Cpp\Sxx(\HJt) + \Sxx(\HJt)^T\Cxp + \Cxp^T\Sxx(\HJt) + \Cxx &\HJt\in(0,+\infty),\\
    \Sxx(0) = \LQRTC.
    \end{dcases}
    }
\end{equation}

When $\Hamiltonian$ does not depend on $\HJx$, we can use our connection to modify the corresponding LQR problem to consider different HJ PDEs and, hence, to accommodate different learning problems. For example, adding lower order terms in the running cost $L$ and/or the source term $f$ of the dynamics corresponds to adding lower order terms in the Hamiltonian of the HJ PDE or, equivalently, in the data fitting term of the learning problem. 
Similarly, adding lower order terms in the 
terminal cost $\HJIC$ corresponds to adding lower order terms in the initial condition of the HJ PDE or, equivalently, in the regularization term of the learning problem. \updatetwo{In Appendix~\ref{appendix:general_LQR}, we present a more general LQR problem with lower order terms that is more amenable to forming our connection to linear regression problems.}

\subsection{Connection to single-point regularized linear regression problems}\label{sec:LQR_1ptregression}

In this section, we establish a relation between the linear regression problem with only one data point and the LQR problem~\eqref{eqt:LQR_general_control} with $\LQRA = \LQRxx = 0$ and $\LQRxu = 0$. Note that, to do this, we also add in some lower order terms to the original LQR problem~\eqref{eqt:LQR_general_control} \updatetwo{(e.g., see Appendix~\ref{appendix:general_LQR})}.
In this case, the LQR problem becomes
    \begin{multline}\label{eqt:optctrl_1pt}
        S(\HJx,\HJt) = \min_{\updateone{\HJu(\cdot)}}  \Bigg\{\int_0^\HJt \left(\frac{1}{2}\HJu(s)^T\LQRuu\HJu(s) -\ba^T\HJu(s)\right)ds \\ + \frac{1}{2}\bx(\HJt)^T\LQRTC\bx(\HJt)
        + \bb^T \bx(\HJt)\colon 
         \dot\bx(s) = \LQRB\HJu(s) \forall s\in(0,\HJt], \bx(0) = \HJx\Bigg\},
    \end{multline}
    and the corresponding HJ PDE is given by~\eqref{eqt:LQR_HJPDE}, where $\HJIC(\HJx) = \frac{1}{2}\HJx^T\LQRTC\HJx + \bb^T \HJx$ is the initial data/terminal cost whose Fenchel transform is given by
    $$\HJIC^*(\HJmom) = \sup_{\HJx\in\Rn} \langle \HJx, \HJmom\rangle - \HJIC(\HJx) = \frac{1}{2}\left\|\LQRTC^{-1/2} (\HJmom - \bb)\right\|_2^2$$
    and the Hamiltonian $\Hamiltonian$ is given by 
    $$
    \Hamiltonian(\HJmom) 
    = \sup_{\HJu\in\R^m} \langle -\LQRB \HJu, \HJmom\rangle - \frac{1}{2}\HJu ^T\LQRuu\HJu + \ba^T \HJu = \frac{1}{2}\left\|\LQRuu^{-1/2}(\LQRB^T\HJmom- \ba)\right\|_2^2, $$
    where $f(\HJu) = \LQRB\HJu$ is the source term of the dynamics and $L(\HJu) = \frac{1}{2}\HJu^T\LQRuu\HJu$ is the running cost. Then, using the single-time Hopf formula~\eqref{eqt:singletime_Hopf}, we have that the solution to the HJ PDE is given by
    \begin{equation}\label{eqt:hopf_1ptregression}
        S(\HJx,\HJt) = \sup_{\HJmom\in\Rn} \left\{\langle \HJx, \HJmom\rangle - \frac{\HJt}{2}\left\|\LQRuu^{-1/2}( \LQRB^T \HJmom - \ba)\right\|_2^2 - \frac{1}{2}\left\|\LQRTC^{-1/2} (\HJmom - \bb)\right\|_2^2 \right\}.
    \end{equation}
    In this case, we can compute the maximizer in the above Hopf formula explicitly, which can be done numerically using the method of least squares.
   
    Alternatively, this LQR problem can also be solved via the Riccati ODEs, which are given by 
    \begin{equation}
    \dot{\Sxx}(\HJt) =  -\Sxx(\HJt)^T\LQRB\LQRuu^{-1}\LQRB^T\Sxx(\HJt), \quad 
    \dot{\Sx}(\HJt) = -\Sxx(\HJt)^T\LQRB\LQRuu^{-1}(\LQRB^T\Sx(\HJt) - \ba),
    \end{equation}
    $$\dot{\Sc}(\HJt) = -\frac{1}{2}\left\|\LQRuu^{-1/2}(\LQRB^T\Sx(\HJt)- \ba)\right\|_2^2$$
    with initial conditions $\Sxx(0) = \LQRTC$, $\Sx(0) = \bb$, and $\Sc(0) = 0$.

The above Hopf formula~\eqref{eqt:hopf_1ptregression} is related to the learning problem~\eqref{eqt:general_learning} with quadratic data fidelity term and quadratic regularization.
Specifically, let $\HJt = \param$ and $\HJmom = \weightvec$. Then, solving the above maximization problem~\eqref{eqt:hopf_1ptregression}  is equivalent to minimizing the following loss function with respect to $\weightvec = [\weight_1, \dots, \weight_n]^T$:
    \begin{equation}\label{eqt:loss_1ptregression}
        \lossfunc(\weightvec) = \frac{\param}{2}\left\|\LQRuu^{-1/2} (\LQRB^T \weightvec - \ba)\right\|_2^2 + \frac{1}{2}\left\|\LQRTC^{-1/2} (\weightvec - (\bb + \LQRTC\HJx))\right\|_2^2.
    \end{equation}
This loss function corresponds to a one-point linear regression problem, where the regularization term is given by $\frac{1}{2}\left\|\LQRTC^{-1/2} (\weightvec - (\bb + \LQRTC\HJx))\right\|_2^2$ and the data fitting term is given by $\frac{\param}{2}\left\|\LQRuu^{-1/2} (\LQRB^T \weightvec - \ba)\right\|_2^2$; i.e., set $\numt = 1$ in Figure~\ref{fig:connection_LQR}.
The minimizer $\weightvec^*$ of $\lossfunc$ is related to the solution of the Riccati equation via
\begin{equation}\label{eqt:singlept_minimizer}
\weightvec^* (=\HJmom^*) = \nabla_\HJx S(\HJx,\param) = \Sxx(\param)\bx + \Sx(\param),
\end{equation}
where $\HJmom^*$ is the minimizer in the Hopf formula~\eqref{eqt:hopf_1ptregression}.
Note that if we only need to recover the minimzer $\weightvec^*$, then we only need to solve two ODEs (namely, the ODEs for $\Sxx, \Sx$) since~\eqref{eqt:singlept_minimizer} does not depend on $\Sc$.

\subsection{Connection to multi-point regularized linear regression problems}\label{sec:LQR_multiptregression}

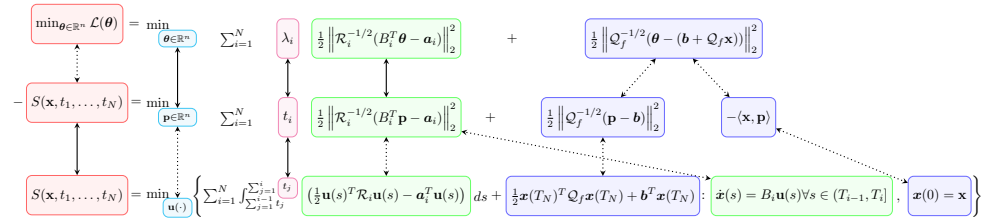
\begin{figure}[htbp]
    \centering
\begin{adjustbox}{width=\textwidth}
\begin{tikzpicture}[node distance=2cm]
    \node (min) [nobox, yshift=-0.2cm] {$\min$};
    \node (minarg) [boxsmall, right of=min, xshift=-1.5cm, yshift=-0.37cm, draw=cyan!60, fill=cyan!5] {$\text{}_{\weightvec\in\weightspace}$};
    \node (sum) [nobox, right of=minarg, xshift=-0.5cm] {$\sum_{i=1}^\numt$};
    \node (param) [box, right of=sum, xshift=-0.7cm, draw=magenta!60, fill=magenta!5] {$\param_i$};
    \node (loss) [box, right of=param, xshift=0.5cm, draw=green!60, fill=green!5] {$\frac{1}{2}\left\|\LQRuu_i^{-1/2}(\LQRB_i^T\weightvec - \ba_i)\right\|_2^2$ };
    \node (plus) [nobox, right of=loss, xshift=1.2cm] {$+$};
    \node (regLeft) [nobox, right of=loss, xshift=3.5cm] {};
    \node (regularization) [box, right of=regLeft, xshift=-0.2cm, draw=blue!60, fill=blue!5] { $\frac{1}{2}\left\|\LQRTC^{-1/2} (\weightvec - (\bb + \LQRTC\HJx))\right\|_2^2$};
    \node (regRight) [nobox, right of=regularization, xshift=-0.1cm] {};
    
    \node (sup) [nobox, below of=min, xshift=-0cm] {$\min$};
    \node (suparg) [boxsmall, right of=sup, xshift=-1.5cm, yshift=-0.37cm, draw=cyan!60, fill=cyan!5] {$\text{}_{\HJmom\in\HJstatespace}$};
    \node (sum) [nobox, below of=sum] {$\sum_{i=1}^\numt$};
    \node (time) [box, below of=param, xshift=-0cm, draw=magenta!60, fill=magenta!5] {$\HJt_i$};
    \node (equals) [nobox, left of=sup, xshift=1.45cm] {$=$};
    \node (S) [box, left of=equals, xshift=0.5cm, draw=red!60, fill=red!5] {$S(\HJx, \HJt_1, \dots, \HJt_\numt)$};
    \node (minus2) [nobox, left of=S, xshift=0.5cm] {$-$};
    \node (Hamiltonian) [box, below of=loss, draw=green!60, fill=green!5] {$\frac{1}{2}\left\|\LQRuu_i^{-1/2}(\LQRB_i^T\HJmom- \ba_i)\right\|_2^2$};
    \node (plus2) [nobox, right of=Hamiltonian,xshift=0.65cm] {$+$};
    \node (IC) [box, below of=regLeft, draw=blue!60, fill=blue!5] {$\frac{1}{2}\left\|\LQRTC^{-1/2}(\HJmom - \bb)\right\|_2^2$};
    \node (linear) [box, below of=regRight, draw=blue!60, fill=blue!5] {$-\langle \HJx, \HJmom\rangle$};

    \node (OCmin) [nobox, below of=sup, yshift=-0.3cm] {$\min$};
    \node (OCminarg) [boxsmall, right of=OCmin, xshift=-1.5cm, yshift=-0.37cm, draw=cyan!60, fill=cyan!5] {$\text{}_{\HJu(\cdot)}$};
    \node (OCS) [box, below of=S, xshift=0cm, yshift=-0.35cm, draw=red!60, fill=red!5] {$S(\HJx, \HJt_1, \dots, \HJt_\numt)$};
    \node (OCequals) [nobox, below of=equals, xshift=0cm, yshift=-0.35cm] {$=$};
    \node (OCint) [nobox, below of=time, xshift=-1.1cm] {$\sum_{i=1}^\numt \int_{\sum_{j=1}^{i-1}\HJt_j}^{\sum_{j=1}^{i}} $};
    \node (OCt) [boxsmall, right of=OCint, xshift=-0.9cm, yshift=0.2cm, draw=magenta!60, fill=magenta!5] {$\text{}^{\HJt_j}$};
    \node (OCHamiltonian) [box, below of=Hamiltonian, xshift=-0cm, draw=green!60, fill=green!5] {$\left(\frac{1}{2}\HJu(s)^T\LQRuu_i\HJu(s) -\ba_i^T\HJu(s)\right)$};
    \node (OCds) [nobox, right of=OCHamiltonian, xshift=0.4cm] {$ds$};
    \node (OCplus) [nobox, right of=OCds, xshift=-1.6cm] {$+$};
    \node (OCIC) [box, below of=IC, draw=blue!60, fill=blue!5] {$\frac{1}{2}\bx(T_\numt)^T\LQRTC\bx(T_\numt) + \bb^T \bx(T_\numt)$};
    \node (OCdynam) [box, right of=OCIC, xshift=3.05cm, draw=green!60, fill=green!5] {$\dot\bx(s) = \LQRB_i\HJu(s)\forall s\in \left(T_{i-1}, T_i\right]$};
    \node (OCinitialposition) [box, right of=OCdynam, xshift=1.5cm, draw=blue!60, fill=blue!5] {$\bx(0) = \HJx$};
    \node (leftbracket) [nobox, left of=OCint, xshift=0.8cm] {$\Bigg\{$};
    \node (colon) [nobox, right of=OCIC, xshift=0.55cm] {$:$};
    \node (comma) [nobox, right of=OCdynam, xshift=0.45cm, yshift=-0.11cm] {,};
    \node (rightbracket) [nobox, right of=OCinitialposition, xshift=-1.05cm] {$\Bigg\}$};

    \node (LPequal) [nobox, above of=equal] {$=$};
    \node (minloss) [box, above of=S, draw=red!60, fill=red!5] {$\min_{\weightvec\in\weightspace}\mathcal{L}(\weightvec)$};
    \draw [dottedarrow] (S) -- (minloss);
    
    \draw [doublearrow] (loss) -- (Hamiltonian);
    \draw [dottedarrow] (regularization) -- (IC);
    \draw [dottedarrow] (regularization) -- (linear);
    \draw [doublearrow] (minarg) -- (suparg);
    \draw [doublearrow] (OCS) -- (S);
    \draw [doublearrow] (OCt) -- (time);
    \draw [dottedarrow] (OCHamiltonian) -- ++(0,1.5cm);
    \draw [dottedarrow] (OCIC) -- (IC);
    \draw [dottedarrow] (OCdynam) -- (Hamiltonian);
    \draw [doublearrow] (time) -- (param);
    \draw [dottedarrow] (linear) -- (OCinitialposition);
    \draw [dottedarrow] (suparg) -- (OCminarg);
\end{tikzpicture}
\end{adjustbox}
    \caption{(See Section~\ref{sec:LQR}) Mathematical formulation describing the connection between a regularized linear regression problem  (\textbf{top}), the multi-time Hopf formula (\textbf{middle}), and a piecewise LQR problem with $\LQRA = \LQRxx = 0$ and $\LQRxu = 0$ on each piece (\textbf{bottom}). Note that $T_i = \sum_{j=1}^i t_j$. The content of this illustration is a special case of the connection in Figure~\ref{fig:connection_multitimeHopf} using quadratic data fitting losses and quadratic regularization. The colors indicate the associated quantities between each problem. The solid-line arrows denote direct equivalences. The dotted arrows represent additional mathematical relations.}
    \label{fig:connection_LQR}
\end{figure}

If the running cost is piecewise quadratic with $\LQRxx = 0, \LQRxu = 0$ on each piece and the dynamics are piecewise linear with $\LQRA = 0$ on each piece, then we get a more general piecewise LQR problem in the form of~\eqref{eqt:optimal_control_standardform}
with piecewise running costs $L_i(\HJu) = \frac{1}{2}\HJu ^T\LQRuu_i\HJu - \ba_i^T \HJu$, where each $\LQRuu_i\in \R^{m\times m}$ is symmetric positive definite, piecewise dynamics $f_i(\HJu) = \LQRB_i \HJu$, terminal cost $\HJIC(\HJx) = \frac{1}{2}\HJx^T\LQRTC \HJx + \bb^T\HJx$, and terminal time $\HJt = \sum_{j=1}^\numt \HJt_j$. Recall that this optimal control problem corresponds to the multi-time HJ PDE~\eqref{eqt:multitimeHJPDE} with Hamiltonians $\Hamiltonian_i(\HJmom) = \sup_{\HJu\in\R^m} \{\langle -f_i(\HJu), \HJmom\rangle - L_i(\HJu)\}$ and initial data $\HJIC$. Then, the solution to this LQR problem and corresponding HJ PDE is given by the following  multi-time Hopf formula:
\begin{equation} \label{eqt:piece_LQR_Hopf}
S(\HJx,\HJt_1, \dots, \HJt_\numt) = \sup_{\HJmom\in\Rn} \left\{\langle \HJx, \HJmom\rangle - \sum_{i=1}^\numt \frac{\HJt_i}{2}\|\LQRuu_i^{-1/2}(\LQRB_i^T\HJmom- \ba_i)\|_2^2 - \frac{1}{2}\|\LQRTC^{-1/2}(\HJmom - \bb)\|_2^2\right\}.
\end{equation}

Define $T_i$ to be
\begin{equation}
T_i = \sum_{j=1}^i t_j.
\end{equation}
Then, the multi-time HJ PDE and piecewise LQR problem can also be solved using the following Riccati equation: 
$S(\HJx,\HJt_1,\cdots, \HJt_N) = \frac{1}{2} \HJx^T \Sxx(T_N)\HJx + \Sx(T_N)^T\HJx + \Sc(T_N)$, where the function $\Sxx:[0,\infty)\to\R^{n\times n}$ takes values in the space of symmetric positive definite matrices and solves the following piecewise Riccati equation:
\begin{equation} \label{eqt: odeP_piecewise}
    \begin{dcases}
    \dot{\Sxx}(s) =  -\Sxx(s)^T\LQRB_i\LQRuu_i^{-1}\LQRB_i^T\Sxx(s) &s\in \left(T_{i-1}, T_i\right)\\
    \Sxx(0) = \LQRTC,
    \end{dcases}
\end{equation}
the function $\Sx:[0,\infty)\to\Rn$ solves the following piecewise linear ODE:
\begin{equation} \label{eqt: odeq_piecewise}
    \begin{dcases}
    \dot{\Sx}(s) = -\Sxx(s)^T\LQRB_i\LQRuu_i^{-1}(\LQRB_i^T\Sx(s) - \ba_i)&s\in \left(T_{i-1}, T_i\right),\\
    \Sx(0) = -\bb,
    \end{dcases}
\end{equation}
and the function $\Sc:[0,\infty)\to\R$ solves the following piecewise ODE:
\begin{equation} \label{eqt: oder_piecewise}
    \begin{dcases}
    \dot{\Sc}(s) = -\frac{1}{2}\left\|\LQRuu_i^{-1/2}(\LQRB_i^T\Sx(s)- \ba_i)\right\|_2^2 &s\in \left(T_{i-1}, T_i\right),\\
    \Sc(0) = 0.
    \end{dcases}
\end{equation}

The above multi-time Hopf formula~\eqref{eqt:piece_LQR_Hopf} can also be regarded as a linear regression problem with multiple data points, as illustrated in Figure~\ref{fig:connection_LQR}. Let  $\frac{1}{2}\|\LQRTC^{-1/2}(\HJmom - \bb)\|_2^2$ be the the regularization term and  $\frac{\HJt_i}{2}\|\LQRuu_i^{-1/2}(\LQRB_i^T\HJmom- \ba_i)\|_2^2$ be the data fitting term at the $i$-th data point. 
Then, the multi-time LQR problem~\eqref{eqt:piece_LQR_Hopf} is equivalent to the learning problem $\min_{\weightvec}\lossfunc(\weightvec)$, where the loss function $\lossfunc:\Rn\to\R$ is given by
\begin{equation} \label{eqt:regression_multidata}
\lossfunc(\weightvec) = \sum_{i=1}^\numt \frac{\param_i}{2}\left\|\LQRuu_i^{-1/2}(\LQRB_i^T\weightvec - \ba_i)\right\|_2^2 + \frac{1}{2}\left\|\LQRTC^{-1/2} (\weightvec - (\bb + \LQRTC\HJx))\right\|_2^2.
\end{equation}
In Section~\ref{subsec:method_1}, we discuss a specific example of the learning problem~\eqref{eqt:regression_multidata}, which is more readily recognizable as the standard linear regression problem. The minimizer $\weightvec^*$ of the learning problem~\eqref{eqt:regression_multidata} (and $\HJmom^*$ of the Hopf formula~\eqref{eqt:piece_LQR_Hopf}) is given by 
\begin{equation}\label{eqt:multipt_minimizer}
\weightvec^* (=\HJmom^*) = \nabla_{\HJx} S(\HJx,\HJt_1,\dots, \HJt_N) = \Sxx\left(T_N\right)\HJx + \Sx\left(T_N\right).
\end{equation}
\updatethree{For more details, we refer readers to~\cite{bardi1984hopf}. Note that one could also use the Pontryagin maximum principle to compute the gradient of the solution to the HJ PDE, which gives the same result as in~\eqref{eqt:multipt_minimizer}.}

%% file: methodology.tex
\section{Methodology}\label{sec:method}
In the previous section, we established a novel theoretical connection between multi-time HJ PDEs, LQR problems, and regularized linear regression problems. In this section, we leverage this theoretical connection to design new algorithms for solving various learning problems.
The connection between the learning problem of interest, the Hopf formula, and the corresponding optimal control problem is summarized in Figure~\ref{fig:connection_Riccati_learning} and Table~\ref{tab:learning_problems}.

\subsection{Solving the regularized linear regression problem using Riccati ODEs}\label{subsec:method_1}
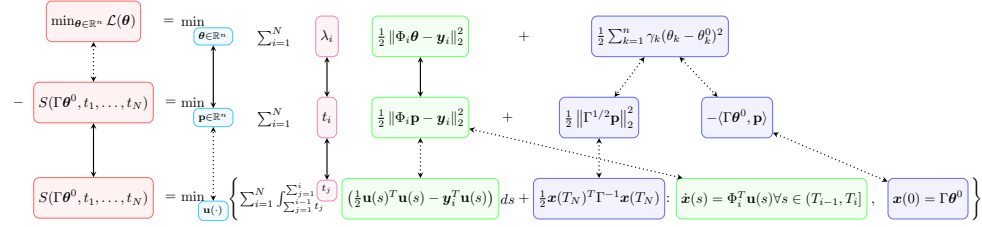
\begin{figure}[htbp]
    \centering
\begin{adjustbox}{width=\textwidth}
\begin{tikzpicture}[node distance=2cm]
    \node (min) [nobox, yshift=-0.2cm] {$\min$};
    \node (minarg) [boxsmall, right of=min, xshift=-1.5cm, yshift=-0.37cm, draw=cyan!60, fill=cyan!5] {$\text{}_{\weightvec\in\weightspace}$};
    \node (sum) [nobox, right of=minarg, xshift=-0.5cm] {$\sum_{i=1}^\numt$};
    \node (param) [box, right of=sum, xshift=-0.7cm, draw=magenta!60, fill=magenta!5] {$\param_i$};
    \node (loss) [box, right of=param, xshift=0.3cm, draw=green!60, fill=green!5] {$\frac{1}{2}\left\|\Phi_i \weightvec - \MLy_i\right\|_2^2$ };
    \node (plus) [nobox, right of=loss, xshift=0.55cm] {$+$};
    \node (regLeft) [nobox, right of=loss, xshift=2.4cm] {};
    \node (regularization) [box, right of=regLeft, xshift=-0.5cm, draw=blue!60, fill=blue!5] { $\frac{1}{2}\sum_{k=1}^n\MLreg_k (\weight_k - \MLcenter_k)^2$};
    \node (regRight) [nobox, right of=regularization, xshift=-0.1cm] {};
    
    \node (sup) [nobox, below of=min, xshift=-0cm] {$\min$};
    \node (suparg) [boxsmall, right of=sup, xshift=-1.5cm, yshift=-0.37cm, draw=cyan!60, fill=cyan!5] {$\text{}_{\HJmom\in\HJstatespace}$};
    \node (sum) [nobox, below of=sum] {$\sum_{i=1}^\numt$};
    \node (time) [box, below of=param, xshift=-0cm, draw=magenta!60, fill=magenta!5] {$\HJt_i$};
    \node (equals) [nobox, left of=sup, xshift=1.45cm] {$=$};
    \node (S) [box, left of=equals, xshift=0.1cm, draw=red!60, fill=red!5] {$S(\MLregmat\MLcentervec, \HJt_1, \dots, \HJt_\numt)$};
    \node (minus2) [nobox, left of=S, xshift=0.15cm] {$-$};
    \node (Hamiltonian) [box, below of=loss, draw=green!60, fill=green!5] {$\frac{1}{2}\left\|\Phi_i \HJmom - \MLy_i\right\|_2^2$};
    \node (plus2) [nobox, right of=Hamiltonian,xshift=0.15cm] {$+$};
    \node (IC) [box, below of=regLeft, draw=blue!60, fill=blue!5] {$\frac{1}{2}\left\|\MLregmat^{1/2}\HJmom\right\|_2^2$};
    \node (linear) [box, below of=regRight, draw=blue!60, fill=blue!5] {$-\langle \MLregmat\MLcentervec, \HJmom\rangle$};

    \node (OCmin) [nobox, below of=sup, yshift=-0.3cm] {$\min$};
    \node (OCminarg) [boxsmall, right of=OCmin, xshift=-1.5cm, yshift=-0.37cm, draw=cyan!60, fill=cyan!5] {$\text{}_{\HJu(\cdot)}$};
    \node (OCS) [box, below of=S, xshift=0cm, yshift=-0.35cm, draw=red!60, fill=red!5] {$S(\MLregmat\MLcentervec, \HJt_1, \dots, \HJt_\numt)$};
    \node (OCequals) [nobox, below of=equals, xshift=0cm, yshift=-0.35cm] {$=$};
    \node (OCint) [nobox, below of=time, xshift=-1.1cm] {$\sum_{i=1}^\numt \int_{\sum_{j=1}^{i-1}\HJt_j}^{\sum_{j=1}^{i}} $};
    \node (OCt) [boxsmall, right of=OCint, xshift=-0.9cm, yshift=0.2cm, draw=magenta!60, fill=magenta!5] {$\text{}^{\HJt_j}$};
    \node (OCHamiltonian) [box, below of=Hamiltonian, xshift=-0cm, draw=green!60, fill=green!5] {$\left(\frac{1}{2}\HJu(s)^T\HJu(s) -\MLy_i^T\HJu(s)\right)$};
    \node (OCds) [nobox, right of=OCHamiltonian, xshift=0.15cm] {$ds$};
    \node (OCplus) [nobox, right of=OCds, xshift=-1.6cm] {$+$};
    \node (OCIC) [box, below of=IC, draw=blue!60, fill=blue!5] {$\frac{1}{2}\bx(T_\numt)^T\MLregmat^{-1}\bx(T_\numt)$};
    \node (OCdynam) [box, right of=OCIC, xshift=2.25cm, draw=green!60, fill=green!5] {$\dot\bx(s) = \Phi_i^T\HJu(s)\forall s\in \left(T_{i-1}, T_i\right]$};
    \node (OCinitialposition) [box, right of=OCdynam, xshift=1.85cm, draw=blue!60, fill=blue!5] {$\bx(0) = \MLregmat\MLcentervec$};
    \node (leftbracket) [nobox, left of=OCint, xshift=0.8cm] {$\Bigg\{$};
    \node (colon) [nobox, right of=OCIC, xshift=-0.3cm] {$:$};
    \node (comma) [nobox, right of=OCdynam, xshift=0.5cm, yshift=-0.11cm] {,};
    \node (rightbracket) [nobox, right of=OCinitialposition, xshift=-0.8cm] {$\Bigg\}$};

    \node (LPequal) [nobox, above of=equal] {$=$};
    \node (minloss) [box, above of=S, draw=red!60, fill=red!5] {$\min_{\weightvec\in\weightspace}\mathcal{L}(\weightvec)$};
    \draw [dottedarrow] (S) -- (minloss);
    
    \draw [doublearrow] (loss) -- (Hamiltonian);
    \draw [dottedarrow] (regularization) -- (IC);
    \draw [dottedarrow] (regularization) -- (linear);
    \draw [doublearrow] (minarg) -- (suparg);
    \draw [doublearrow] (OCS) -- (S);
    \draw [doublearrow] (OCt) -- (time);
    \draw [dottedarrow] (OCHamiltonian) -- ++(0,1.5cm);
    \draw [dottedarrow] (OCIC) -- (IC);
    \draw [dottedarrow] (OCdynam) -- (Hamiltonian);
    \draw [doublearrow] (time) -- (param);
    \draw [dottedarrow] (linear) -- (OCinitialposition);
    \draw [dottedarrow] (OCminarg) -- (suparg);
\end{tikzpicture}
\end{adjustbox}
    \caption{(See Section~\ref{sec:method}) Mathematical formulation describing the connection between our model linear regression problem with quadratic regularization (\textbf{top}), the multi-time Hopf formula (\textbf{middle}), and a piecewise LQR problem (\textbf{bottom}). Note that $T_i = \sum_{j=1}^i t_j$. The content of this illustration is a special case of the connection in Figure~\ref{fig:connection_LQR} using $\LQRuu_i = I$, $B_i = \Phi_i^T$, $\ba_i = \MLy_i$, $\LQRTC = \MLregmat^{-1}$, $\bb = 0$, and $\HJx = \MLregmat\MLcentervec$. We use this connection to develop our Riccati-based methodology (Section~\ref{sec:method}) and perform our numerical examples (Section~\ref{sec:numerics}). The colors indicate the associated quantities between each problem. The solid-line arrows denote direct equivalences. The dotted arrows represent additional mathematical relations.}
    \label{fig:connection_Riccati_learning}
\end{figure}

\begin{table}[ht]
    \footnotesize
    \centering
    \begin{adjustbox}{width=\textwidth}
    \begin{tabular}{c|c|c|c}
    \hline\hline
        \begin{tabular}{c}
            \textbf{Terms in the} \\ \textbf{loss function~\eqref{eq:loss_function}}
        \end{tabular} & \begin{tabular}{c}
            \textbf{Terms in the} \\ \textbf{Hopf formula}
        \end{tabular} & \begin{tabular}{c}
            \textbf{Terms in the optimal} \\
            \textbf{ control problem}
        \end{tabular} & \textbf{Learning problems} \\
        \hline
        $\Phi_i, \MLy_i$ & Hamiltonian & Running cost, dynamics & \begin{tabular}{c}
            Continual learning (Section~\ref{subsec:example_1}); \\ Post-training calibration (Section~\ref{subsec:example_2})
        \end{tabular} \\
        \hline
        $\param_i$ & Time & Time & Post-training calibration (Section~\ref{subsec:example_2})\\
        \hline
        $\MLreg_k$ & \begin{tabular}{c}
            Hamiltonian; \\ Initial condition
        \end{tabular}
        & \begin{tabular}{c}
            Running cost, dynamics; \\ Terminal cost
        \end{tabular} & \begin{tabular}{c}
        Hyper-parameter tuning, \\ flow along the Pareto front (Section~\ref{subsec:example_3}) 
        \end{tabular} \\
         \hline
        $\MLcenter_k$ & Spatial variable & Initial position & Generalized regularization functions (Section~\ref{subsec:example_4}) \\
        \hline \hline
    \end{tabular} 
    
    \end{adjustbox}
    \caption{Summary of the learning problems and corresponding numerical examples presented in this work. Each row summarizes which terms in the loss function~\eqref{eq:loss_function}, the Hopf formula, and the optimal control problem must be changed to match the context of various learning problems.}
    \label{tab:learning_problems}
\end{table}

A linear regression problem with quadratic data fitting loss and quadratic regularization is formulated as follows.  
The goal of this learning problem is to fit $N$ data points $(\bz_i, \MLy_i)\in\R^M\times\R^m$ with the linear prediction model $\Phi_i \weightvec\approx \MLy_i$, where $\Phi_i = [\phi_1(\bz_i), ..., \phi_n(\bz_i)]\in\R^{m\times n}$ is the matrix whose columns are the basis functions $\phi_j:\R^M\to\R^m$, $j=1, \dots, n$ evaluated at $\bz_i$ and $\weightvec = [\weight_1, \dots, \weight_n]^T\in\Rn$ are unknown trainable coefficients. We learn $\weightvec\in\Rn$ by minimizing the following loss function:
\begin{align}\label{eq:loss_function}
    \mathcal{L}(\weightvec) = \frac{1}{2}\sum_{i=1}^N\lambda_i\|\Phi_i \weightvec - \MLy_i\|_2^2 + \frac{1}{2}\sum_{k=1}^n\MLreg_k (\weight_k - \MLcenter_k)^2,
\end{align}
where $\lambda_i\geq 0, i=1,...,N$ are tunable hyper-parameters for the data fitting losses, $\MLreg_k>0, k=1,...,n$ are tunable hyper-parameters for the regularization terms, and $\MLcenter_k \in \R$ is a prior on the unknown coefficients that biases $\weight_k$ to be close to $\MLcenter_k$. Note that the above loss function~\eqref{eq:loss_function} is strictly convex, and hence it has a unique global minimizer. Since this loss function is quadratic in $\weightvec$, it can be minimized exactly using the method of least squares. However, we explore a different approach for minimizing~\eqref{eq:loss_function}, which, in certain learning contexts, yields computational advantages over conventional approaches like the method of least squares.

Note that this learning problem is in the form of~\eqref{eqt:regression_multidata}; i.e., set $\LQRuu_i = I_{m\times m}$, $B_i = \Phi_i^T$, $\ba_i = \MLy_i$, $\LQRTC = \MLregmat^{-1}$, where $\MLregmat\in\R^{n\times n}$ is the diagonal matrix whose $i$-th entry is $\MLreg_i$, $\bb = 0$, and $\HJx = \MLregmat\MLcentervec$. Then, this learning problem is related to an LQR problem, and we summarize this connection in Figure~\ref{fig:connection_Riccati_learning}.
Thus, this learning problem can alternatively be solved via the following Riccati ODEs:
\begin{equation}\label{eqt:sequentialRiccatiODEs}
    \begin{dcases}
    \dot{\Sxx}(\HJt) =  -\Sxx(\HJt)^T\MLbasismat_i^T\MLbasismat_i\Sxx(\HJt) &\HJt\in \left(T_{i-1}, T_i\right),\\
    \dot{\Sx}(\HJt) = -\Sxx(\HJt)^T\MLbasismat_i^T(\MLbasismat_i\Sx(\HJt) - \MLy_i)&\HJt\in \left(T_{i-1}, T_i\right),
    \end{dcases}
\end{equation}
with initial condition $\Sxx(0) = \MLregmat^{-1}$, $\Sx(0) = 0$.
Note that here we disregard the ODE for $\Sc$ since, for the learning problem, we are only concerned with the value of the minimizer $\weightvec^*$, which only requires the values of $\Sxx,\Sx$ (e.g., see~\eqref{eqt:singlept_minimizer}, \eqref{eqt:multipt_minimizer}), whereas recovering the minimal objective value $\L(\weightvec^*)$ (which is generally not needed in the context of learning) would require all three values $\Sxx,\Sx,\Sc$. 
There are many numerical methods for solving the Riccati ODEs~\eqref{eqt:sequentialRiccatiODEs} in the literature. In this paper, we use the 4th-order Runge-Kutta method (RK4) \cite{butcher2016numerical}, but other methods could be used instead.

At first glance, solving this linear regression problem via the Riccati ODEs \eqref{eqt:sequentialRiccatiODEs} may seem unnecessary given the existence of other well-established methods for minimizing~\eqref{eq:loss_function} (e.g., the method of least squares). However, note that~\eqref{eqt:sequentialRiccatiODEs} is actually a sequence of Riccati ODEs. Thus, using our theoretical connection (and hence, these sequential Riccati ODEs to minimize~\eqref{eq:loss_function}) means that we have the flexibility to handle sequential changes to the learning problem. 
In the remainder of Section~\ref{sec:method}, 
we identify several applications in learning (summarized in Table~\ref{tab:learning_problems}), for which this Riccati-based approach yields computational advantages over traditional learning methods (especially when the number of data points $N$ is large), and we describe how standard Riccati solvers can be adapted for these contexts. \updatethree{In Appendix~\ref{appendix:comp_complexity}, we provide a more detailed comparison of the computational complexity of using RK4 and our Riccati-based methodology versus the method of least squares for these applications.}

\subsection{Adding or removing data}\label{subsec:method_2}
Since our theoretical connection gives us access to the Riccati ODEs~\eqref{eqt:sequentialRiccatiODEs}, which are solved sequentially, we have the flexibility to handle sequential changes to the learning problem~\eqref{eq:loss_function}. In this section, we focus on the sequential addition or removal of data. Some related machine learning examples are provided in Sections~\ref{subsec:example_1} and \ref{subsec:example_2}.

First, we discuss the addition of one data point; i.e., we increase the number of data points from $N$ to $N+1$. This case corresponds to updating our learned model as new data is collected, which is crucial for many practical machine learning applications. 
To add  one data point, we adapt the Riccati ODEs~\eqref{eqt:sequentialRiccatiODEs} as follows. 
Adding the $(N+1)$-th data point corresponds to adding the term $\frac{1}{2}\lambda_{N+1}\|\Phi_{N+1} \weightvec - \MLy_{N+1}\|_2^2 $ in the loss function~\eqref{eq:loss_function} or, equivalently, to adding the Hamiltonian $\frac{1}{2}\|\Phi_{N+1} \weightvec - \MLy_{N+1}\|_2^2$ to the multi-time HJ PDE and the pieces $L_{N+1}(s, \HJu) = \frac{1}{2}\HJu^T\HJu - \MLy_{N+1}^T\HJu$ and $f(s,\HJu) = \Phi_{N+1}^T\HJu, s\in(T_N, T_{N+1})$ to the running cost and dynamics, respectively, of the corresponding piecewise LQR problem.
Thus, minimizing this new loss function is equivalent to solving the following Riccati ODE:
\begin{equation}\label{eqt:regression_1Riccati}
    \begin{dcases}
    \dot{\tilde\Sxx}(\HJt) =  -\tilde\Sxx(\HJt)^T\MLbasismat_\indexRiccati^T\MLbasismat_\indexRiccati\tilde\Sxx(\HJt) &\HJt>0,\\
    \dot{\tilde\Sx}(\HJt) = -\tilde\Sxx(\HJt)^T\MLbasismat_\indexRiccati^T(\MLbasismat_\indexRiccati\tilde\Sx(\HJt) - \MLy_\indexRiccati)&\HJt>0,
    \end{dcases}
\end{equation}
where the index $\indexRiccati=N+1$ and
with initial condition $\tilde\Sxx(0) = \Sxx\left(T_N\right)$ and $\tilde\Sx(0) = \Sx\left(T_N\right)$, where $\Sxx\left(T_{N}\right)$ and $\Sx\left(T_{N}\right)$ are obtained from solving the learning problem~\eqref{eq:loss_function} with $N$ data points. Then, the solution to the new learning problem with an additional point is given by
\begin{equation}\label{eqt:sec42_newoptimizer}
 \tilde\weightvec^* = \tilde\Sxx\MLregmat\MLcentervec + \tilde\Sx,
\end{equation}
where $\tilde\Sxx = \tilde\Sxx(\lambda_{N+1}) (=\Sxx(T_{N+1}))$ and $\tilde \Sx = \tilde \Sx(\lambda_{N+1}) (=\Sx(T_{N+1}))$ are the solution to~\eqref{eqt:regression_1Riccati}; i.e., we have evolved the solution of the new corresponding multi-time HJ PDE in the time variable $t_{N+1}$ from $S(\HJx, \HJt_1, \dots, \HJt_N, 0)$ to $S(\HJx, \HJt_1, \dots, \HJt_N, \lambda_{N+1})$. Equivalently, we can also interpret adding one data point as solving a one-point linear regression problem~\eqref{eqt:loss_1ptregression} and its corresponding single-piece LQR problem~\eqref{eqt:optctrl_1pt}.

Removing one data point (i.e., decreasing the number of data points from $N$ to $N-1$) corresponds to calibrating our learned model by removing possible outliers and/or overly noisy data. 
To remove one data point, we reverse time and solve a terminal value Riccati ODE~\eqref{eqt:regression_2Riccati}. The solution to the new learning problem is then given by~\eqref{eqt:sec42_newoptimizer}, where $\tilde \Sxx$, $\tilde\Sx$ are given by the solution to the time-reversed Riccati ODEs. For more details and the mathematical derivation, see Appendix~\ref{appendix:method_delete_data}.

Note that in both cases, the above approach only requires information about the data point to be added or removed and the results of the previous training. 
Thus, we can add and remove data without retraining on the entire dataset or requiring access to all of the previous data. In contrast, traditional approaches to minimizing~\eqref{eq:loss_function} (e.g., the method of least squares) would require memory of all previous data and then retraining on the entire updated data set. Hence, our approach provides promising computational and memory savings over conventional methods. \updatetwo{While, in the numerics, we use RK4 to solve the corresponding Riccati ODEs in both of these cases, we note that other numerical methods could also be applied to solve these Riccati ODEs and hence, to achieve these computational and memory savings. For example, in Appendix~\ref{sec:RLS}, we discuss how recursive least squares can be applied to these cases.}

\subsection{Hyper-parameter tuning}\label{subsec:method_3}
In the loss function~\eqref{eq:loss_function}, we have three types of hyper-parameters: the weights $\lambda_i$ of the data fitting terms, the weights $\MLreg_k$ of the regularization term, and the bias $\MLcentervec$ on the trainable coefficients $\weightvec$. 
In this section, we discuss how we adapt the Riccati ODEs to tune each of these hyper-parameters.

First, we discuss tuning the weights of the data fitting terms; e.g., as in post-training calibration and federated learning (Section~\ref{subsec:example_2}). Consider changing the weight on the $i$-th data fitting term from $\lambda_i$ to $\tilde\lambda_i$, which corresponds to changing the time $\HJt_i = \lambda_i$ to $\HJt_i = \tilde\lambda_i$ in the multi-time HJ PDE and corresponding piecewise LQR problem. Then, the methodology is similar to that in Section~\ref{subsec:method_1}. 

If $\tilde\lambda_i > \lambda_i$, then the solution to the learning problem with the new data fitting weight $\tilde\lambda_i$ is given by~\eqref{eqt:sec42_newoptimizer}, where $\tilde\Sxx = \tilde\Sxx( \tilde\lambda_i - \lambda_i)$ and $\tilde\Sx = \tilde\Sx( \tilde\lambda_i - \lambda_i)$ are the solutions to the Riccati ODEs~\eqref{eqt:regression_1Riccati} with $j=i$, at time $(\tilde\lambda_i - \lambda_i)$, and with initial condition $\tilde\Sxx(0) = \Sxx\left(T_N\right)$ and $\tilde\Sx(0) = \Sx\left(T_N\right)$, where $\Sxx(T_N), \Sx(T_N)$ are obtained from solving the original learning problem~\eqref{eq:loss_function} with the original value of $\lambda_i$. Similarly, if $\lambda_i > \tilde \lambda_i$, then the solution to the learning problem with the new data fitting weight $\tilde\lambda_i$ is given by~\eqref{eqt:sec42_newoptimizer}, where $\tilde\Sxx = \tilde\Sxx(0)$ and $\tilde\Sx = \tilde\Sx(0)$ are the solutions to the time-reversed Riccati ODEs~\eqref{eqt:regression_2Riccati} with $j=i$, at time 0, and with terminal condition $\tilde\Sxx(\lambda_i - \tilde\lambda_i) = \Sxx\left(T_N\right)$ and $\tilde\Sx(\lambda_i - \tilde\lambda_i) = \Sx\left(T_N\right)$, where $\Sxx\left(T_N\right), \Sx\left(T_N\right)$ are obtained from solving the original learning problem~\eqref{eq:loss_function} with the original value of $\lambda_i$.

Next, we discuss tuning the weights of the regularization terms; e.g., as in the hyper-parameter tuning example in Section~\ref{subsec:example_3}. Note that in the original formulation of the Riccati ODEs~\eqref{eqt:sequentialRiccatiODEs}, the regularization weights appear in the initial condition for $\Sxx$. Changing a hyper-parameter $\MLreg_k$ by changing this initial condition would require re-solving the entire sequence of Riccati ODEs and hence retraining on the entire dataset. However, since we have freedom in how we formulate the corresponding multi-time HJ PDE, we can avoid this retraining as follows. Note that here, instead of sequentially changing each regularization weight $\MLreg_k$ one-by-one, we can actually update all of the regularization weights at once.  

We update the regularization weights in two steps. First, we update the weights whose values we increase by adding an additional Hamiltonian to the multi-time HJ PDE. This yields an initial value Riccati ODE~\eqref{eqt:riccati_increase_regweight}. Second, we update the remaining weights by reinterpreting the problem as a terminal condition, single-time HJ PDE. We then evolve this HJ PDE backward in time using a terminal condition Riccati ODE~\eqref{eqt:riccati_decrease_reg_weight}. Finally, the minimizer to the new learning problem resulting from changing all of the regularization weights is given by the solutions to the time-reversed Riccati ODE~\eqref{eqt:riccati_decrease_reg_weight} at time 0. For more details, see Appendix~\ref{appendix:hyperparam_tuning}.

Finally, we discuss tuning the bias $\MLcentervec$ on the trainable coefficients $\weightvec$; e.g., as in the generalized regularization example in Section~\ref{subsec:example_4}. In the context of the learning problem, we interpret $\MLcentervec$ as the center of a Gaussian prior on $\weightvec$. Using our theoretical connection, changing $\MLcentervec$ is also equivalent to evaluating the corresponding multi-time HJ PDE and piecewise LQR problem at a different point in space. Specifically, changing $\MLcentervec$ to $\MLcenternewvec$ means that instead of evaluating the multi-time HJ PDE and piecewise LQR problem at $\HJx = \MLregmat\MLcentervec$, we evaluate them at $\HJx = \MLregmat\MLcenternewvec$. Then, the solution to the learning problem with the new bias $\MLcenternewvec$ is given by 
\begin{equation}\label{eqt:minimizer_changeofbias}
    \Sxx(T_N)\MLregmat\MLcenternewvec + \Sx(T_N),
\end{equation}
where $\Sxx(T_N)$ and $\Sx(T_N)$ are obtained from solving the original learning problem~\eqref{eq:loss_function} with the original value of $\MLcentervec$. In other words, shifting the bias involves neither retraining nor access to any of the data points nor solving Riccati ODEs. Instead, we only require some matrix multiplication and addition involving the results of the previous training and the new bias $\MLcenternewvec$.

\subsection{General convex regularization functions}\label{subsec:method_4}
So far, we have only considered quadratic regularizations. In this section, we will consider the linear regression problem with an arbitrary convex regularization term.
Specifically, consider the general regularization term $R(\weightvec) - \langle \HJx, \weightvec\rangle$, where $R:\weightspace\to\R$ is a convex function. Then, the loss function of the learning problem is given by
\begin{align}  \label{eqt:loss_fn_general_reg}
    \mathcal{L}(\weightvec) = \frac{1}{2}\sum_{i=1}^N\lambda_i\|\Phi_i \weightvec - \MLy_i\|_2^2 + R(\weightvec) - \langle \HJx, \weightvec\rangle.
\end{align}
This learning problem corresponds to a multi-time HJ PDE~\eqref{eqt:multitimeHJPDE}, where the $i$-th Hamiltonian is $H_i(\HJmom) = \frac{1}{2}\lambda_i\|\Phi_i \HJmom - \MLy_i\|_2^2$ and the initial condition is $J = R^*$, which  is the Fenchel-Legendre transform of $R$. The corresponding optimal control problem is given by~\eqref{eqt:optimal_control_standardform}, with terminal time $\sum_{i=1}^N\lambda_i$, piecewise running costs  $L_i(\HJu) = \frac{1}{2}\|\HJu\|^2 - \langle \MLy_i, \HJu\rangle$ and piecewise dynamics $f_i(\HJu) = \MLbasismat_i^T\HJu$ on $(\sum_{j=1}^{i-1}\lambda_j, \sum_{j=1}^{i}\lambda_j]$, and terminal cost $R^*$.
Then, the solution $\weightvec^*$ to the learning problem~\eqref{eqt:loss_fn_general_reg}  is equivalent to the spatial gradient $\nabla_\HJx S(\HJx, \lambda_1, \dots, \lambda_N)$ of the solution to the multi-time HJ PDE, and the minimal value of the loss function~\eqref{eqt:loss_fn_general_reg} is equivalent to $-S(\HJx, \lambda_1, \dots, \lambda_N)$. 

Since the loss function~\eqref{eqt:loss_fn_general_reg} is convex, it can be minimized using any appropriate convex optimization algorithm.
In this paper, we demonstrate how our Riccati-based approach can be combined with PDHG \cite{chambolle2011pdhg} to solve this learning problem. Note that PDHG can be applied to this learning problem~\eqref{eqt:loss_fn_general_reg} as long as the proximal point of $R$ or $R^*$ is numerically computable. For instance, $R$ could be a quadratic function $\langle \cdot, \mathcal{M}\cdot\rangle$, a quadratic norm $\sqrt{\langle \cdot, \mathcal{M}\cdot\rangle}$, $\|\cdot\|_1$, $\|\cdot\|_1^2$, $\|\cdot\|_\infty^2$, a polynomial function, or the Fenchel-Legendre transformation of any of these functions \cite{Darbon2016Algorithms}. 
To compute the solution $\weightvec^*$ of this learning problem, PDHG iterates the following:
\begin{equation}\label{eq:pdhg}
\begin{dcases}
\weightvec^{\ell +1} = \argmin_{\weightvec\in\Rn}
\frac{1}{2}\sum_{i=1}^N\lambda_i\|\Phi_i \weightvec - \MLy_i\|_2^2 
+ \frac{1}{2\sigma_\weightvec}\left\|\weightvec - (\weightvec^\ell - \sigma_\weightvec (\pdhgdual^\ell-\HJx))\right\|_2^2,\\
\bar \weightvec^{\ell+1} = 2\weightvec^{\ell+1} - \weightvec^{\ell},\\
\pdhgdual^{\ell+1} = \argmin_{\pdhgdual\in\Rn}  R^*(\pdhgdual) + \frac{1}{2\sigma_{\pdhgdual}}\left\|\pdhgdual - (\pdhgdual^\ell + \sigma_{\pdhgdual} \bar \weightvec^{\ell+1})\right\|_2^2,
\end{dcases}
\end{equation}
where $\sigma_\weightvec$, $\sigma_{\pdhgdual}$ are positive step-size parameters satisfying $\sigma_\weightvec \sigma_{\pdhgdual} < 1$. 
In each iteration, we need to solve two minimization problems. Updating $\weightvec^{\ell}$ is equivalent to solving the learning problem with quadratic regularization.
Specifically, updating $\weightvec^{\ell}$ is the same as changing the bias $\MLcenter$ in the loss function~\eqref{eq:loss_function}. Therefore, we only need to solve the Riccati ODEs~\eqref{eqt:sequentialRiccatiODEs} once (to compute $\weightvec^{1}$) and every other iterate can be computed using~\eqref{eqt:minimizer_changeofbias}.  
Updating $\pdhgdual^{\ell}$ is equivalent to computing the proximal point of $R^*$. Depending on $R^*$, there may already exist efficient solvers or explicit formulas for computing its proximal point.

Note that with non-quadratic regularization, if we change the weights $\lambda_k$ of the data fitting losses, change $\HJx$ (which is related to the bias on $\weightvec$), or add or remove data points, then we will need to restart PDHG to solve the resulting learning problem with these new hyper-parameter values or updated datasets. However, we can reuse some of the computations between runs. Namely, we only need to solve the full sequence of Riccati ODEs~\eqref{eqt:sequentialRiccatiODEs} once (to compute $\weightvec^1$ in the first run of PDHG). Then, the solution to those Riccati ODEs can be reused in combination with the methods described in Sections~\ref{subsec:method_2} and~\ref{subsec:method_3} (according to how the learning problem is modified) in subsequent iterations and runs of PDHG.

%% file: numerics.tex
In this section, we apply the Riccati-based methodology presented in Section~\ref{sec:method} to four test problems from machine learning to demonstrate the versatility and potential computational advantages of our new approach.  
In each example, we use RK4 with double precision to solve the Riccati ODEs when applying our Riccati-based methodology. The connections between the examples presented in this section, the Hopf formula, and  the corresponding optimal control problems can be found in Table~\ref{tab:learning_problems}. 
We note that in this work we use two metrics to evaluate our results quantitatively: the $\ell_1$-norm (defined as $\|x\|_1 = \sum_{i=1}^n |x_i|$ for $x\in\Rn$) for the finite-dimensional minimizer and the $L^2$-norm (defined as $\|f\|_2 = \left(\int_{x\in\Rn} |f(x)|^2 dx\right)^{1/2}$ for $f:\Rn\to \R$) for the inferred functions. The $L^2$-norm for functions is approximated using trapezoidal rule with a uniform grid.
Supplementary details of the numerical experiments can be found in Appendix \ref{sec:details}. Code for all examples will be made publicly available at \url{https://github.com/ZongrenZou/HJPDE4SciML}.

%% file: example_continual.tex
\subsection{Function approximation in continual learning}\label{subsec:example_1}
In this section, we test our Riccati-based approach (as described in Section~\ref{subsec:example_1}) on a pedagogical function approximation example in continual learning \cite{parisi2019continual, kirkpatrick2017overcoming, van2019three} to demonstrate its computational and memory advantages.
Under the continual learning framework, data are accessed in a stream and the trainable model parameters are updated incrementally as new data become available. 
In some cases, the historical data may also become inaccessible after new data are received, which can often lead to catastrophic forgetting \cite{kirkpatrick2017overcoming, parisi2019continual}, which refers to the abrupt degradation in performance of learned models on previous tasks upon training on new tasks. 
In this example, we show how our Riccati-based approach naturally coincides with the continual learning framework, while also inherently avoiding catastrophic forgetting even if the historical data are inaccessible.

Our goal is to regress the function $y(x) = \sin(10 x)$ given noisy data $\{x_i, y_i\approx y(x_i)\}_{i=1}^N$, where the data are corrupted by additive Gaussian noise with relatively large noise scale. Hence, a large amount of data is required in order to obtain an accurate inference. In this example, we follow the continual learning framework and assume that the data are accessed in a stream, so that each new data point must be incorporated into the learned model as soon as it becomes available. Specifically, each new data point comes from a uniformly random sample point $x\in[0, 10]$ and is corrupted by additive Gaussian noise with noise scale one. We regress $y(x)$ using the linear model $y(x) = \sum_{k=1}^n \weight_k \phi_k(x)$, where $n=10$ and 
\begin{equation*}\label{eq:continual:basis}
    \{\phi_k(x)\}_{k=1}^n = \{1, x, x^2, x^3, \sin(\tau), \sin(5x), \sin(8x), \sin(9x), \sin(10x), \sin(12x)\}.
\end{equation*}
We learn the coefficients $\weightvec = [\weight_1, \dots, \weight_n]^T$ by minimizing the following loss function:
\begin{equation}\label{eq:continual:loss}
    \mathcal{L}(\weightvec) = \frac{1}{2}\sum_{i=1}^N \lambda_i \left|\sum_{k=1}^{n} \weight_k\phi_k(x_i) - y_i \right|^2 + \frac{1}{2}\sum_{k=1}^{n} \MLreg_k|\weight_k|^2,
\end{equation}
where $\lambda_i, i=1,...,N$ and $\MLreg_k, k=1,...,n$ are weights for the data loss and $\ell_2$ regularization terms, respectively, and we update our learned coefficients every time a new data point is available. In our numerical experiments, we set $\lambda_i=1, \forall i$ and $\MLreg_k=100, \forall k$.

\begin{figure}[htbp]
    \begin{subfigure}[b]{\textwidth}
        \centering
        \includegraphics[width = 0.38\textwidth]{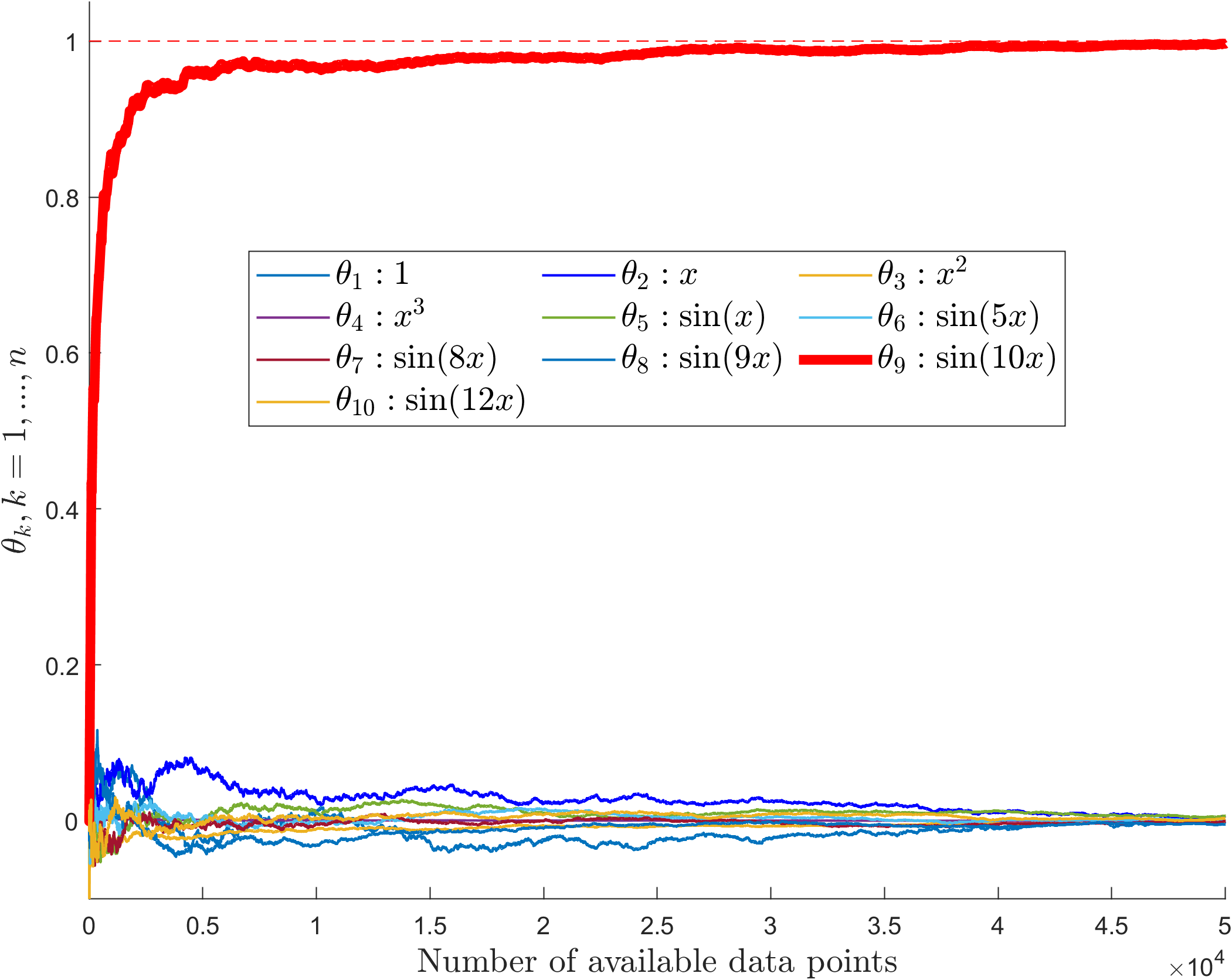}
        \includegraphics[width = 0.38\textwidth]{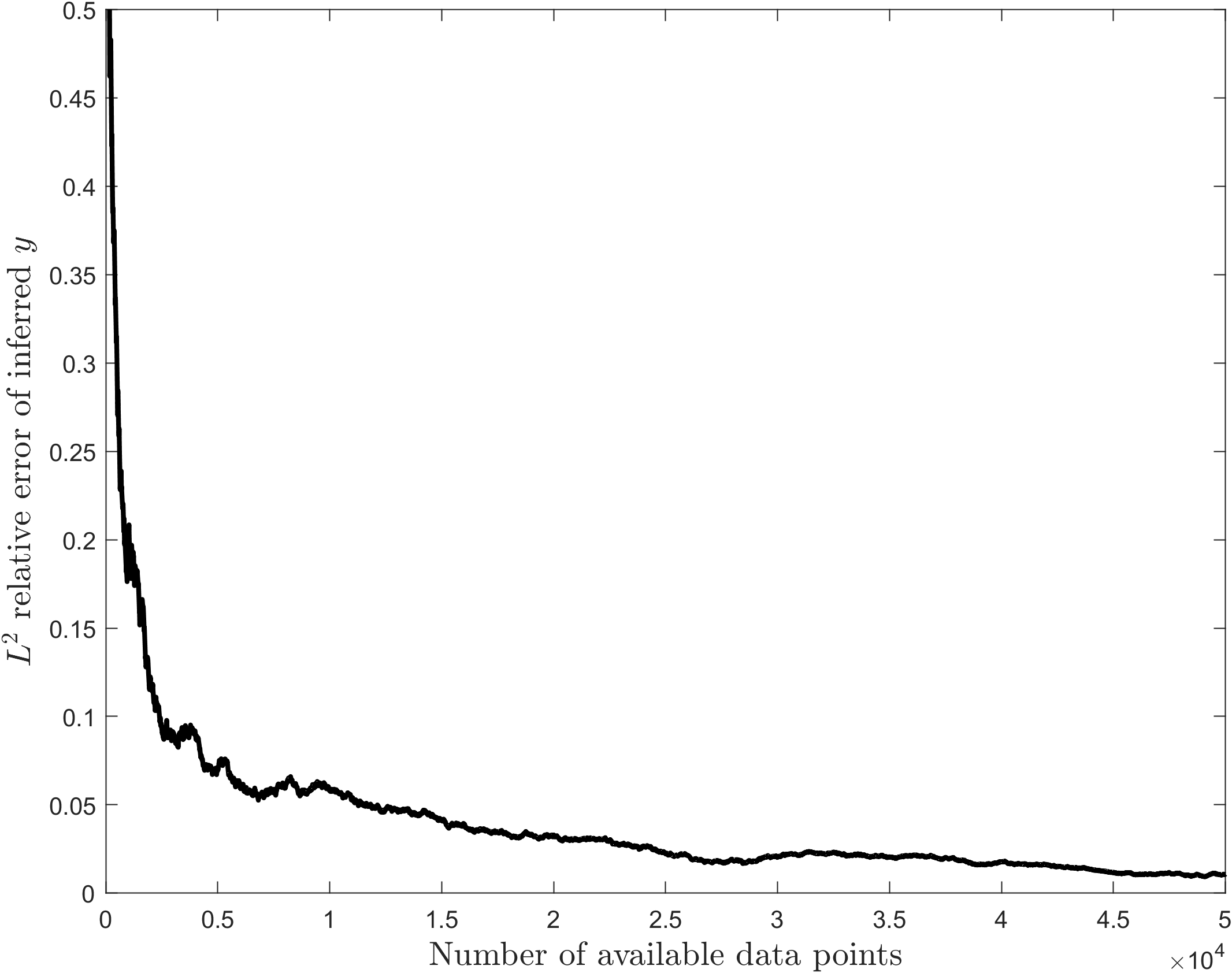}
        \caption{}
    \end{subfigure}
    \begin{subfigure}[b]{\textwidth}
        \centering
        \includegraphics[width = 0.3\textwidth]{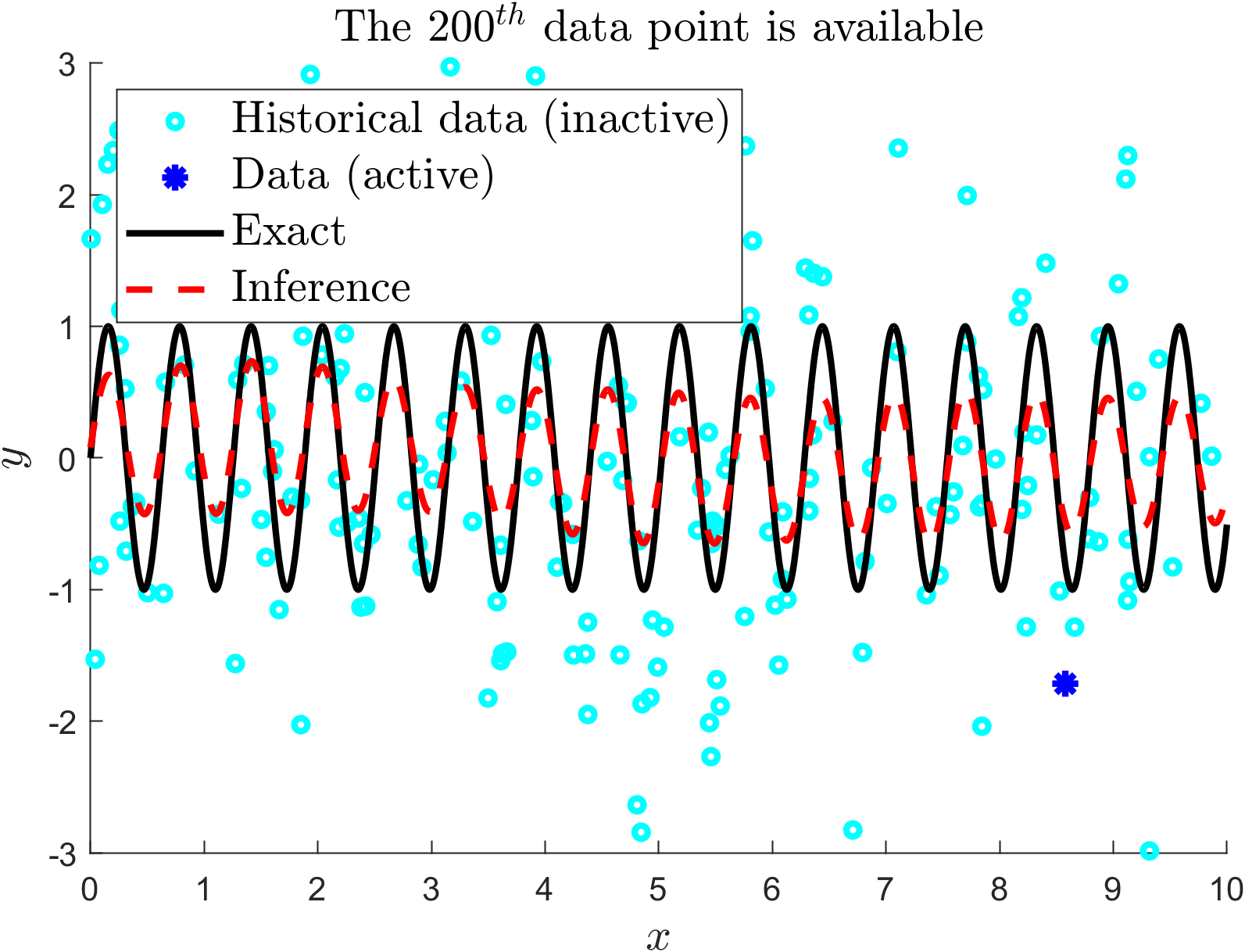}
        \includegraphics[width = 0.3\textwidth]{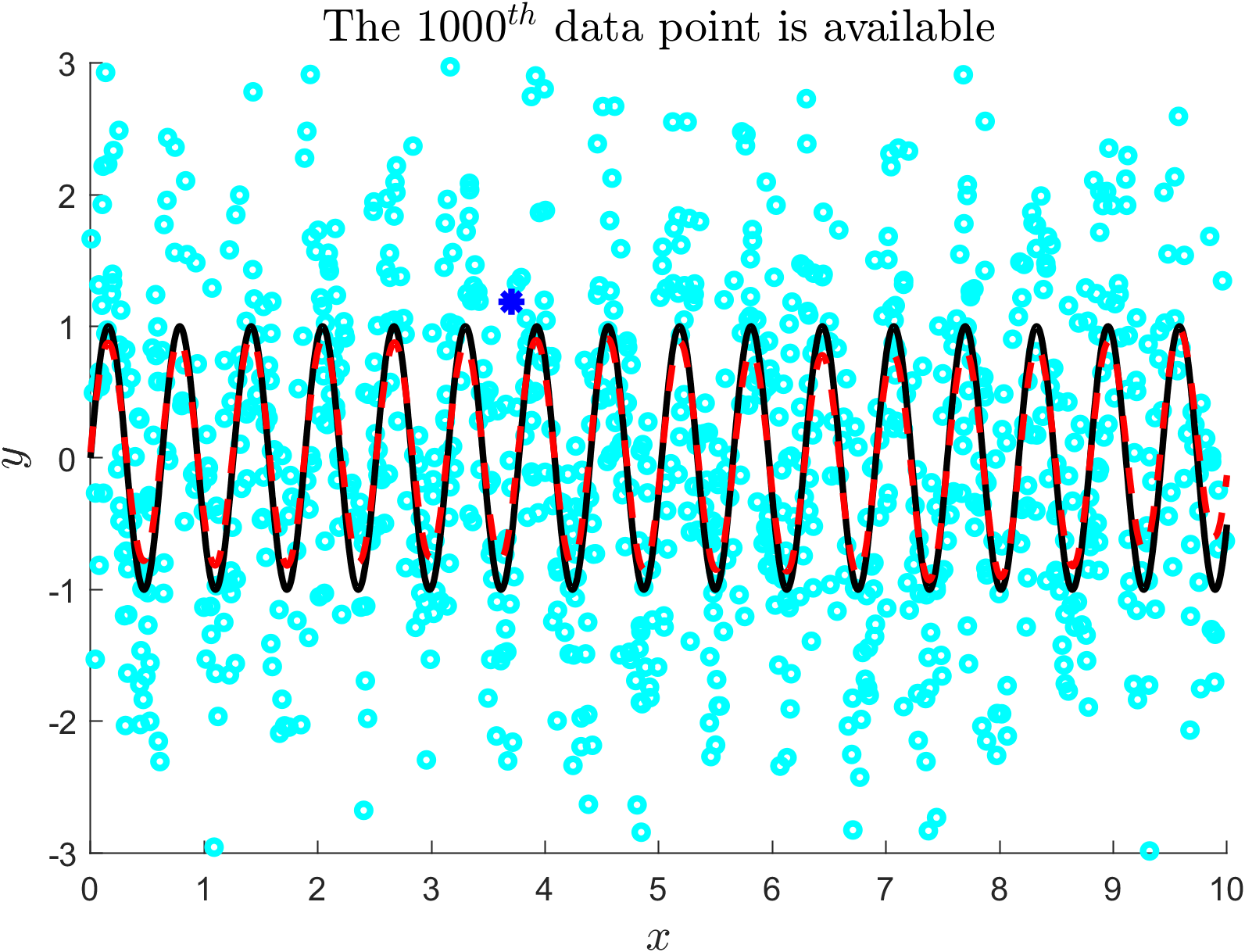}
        \includegraphics[width = 0.3\textwidth]{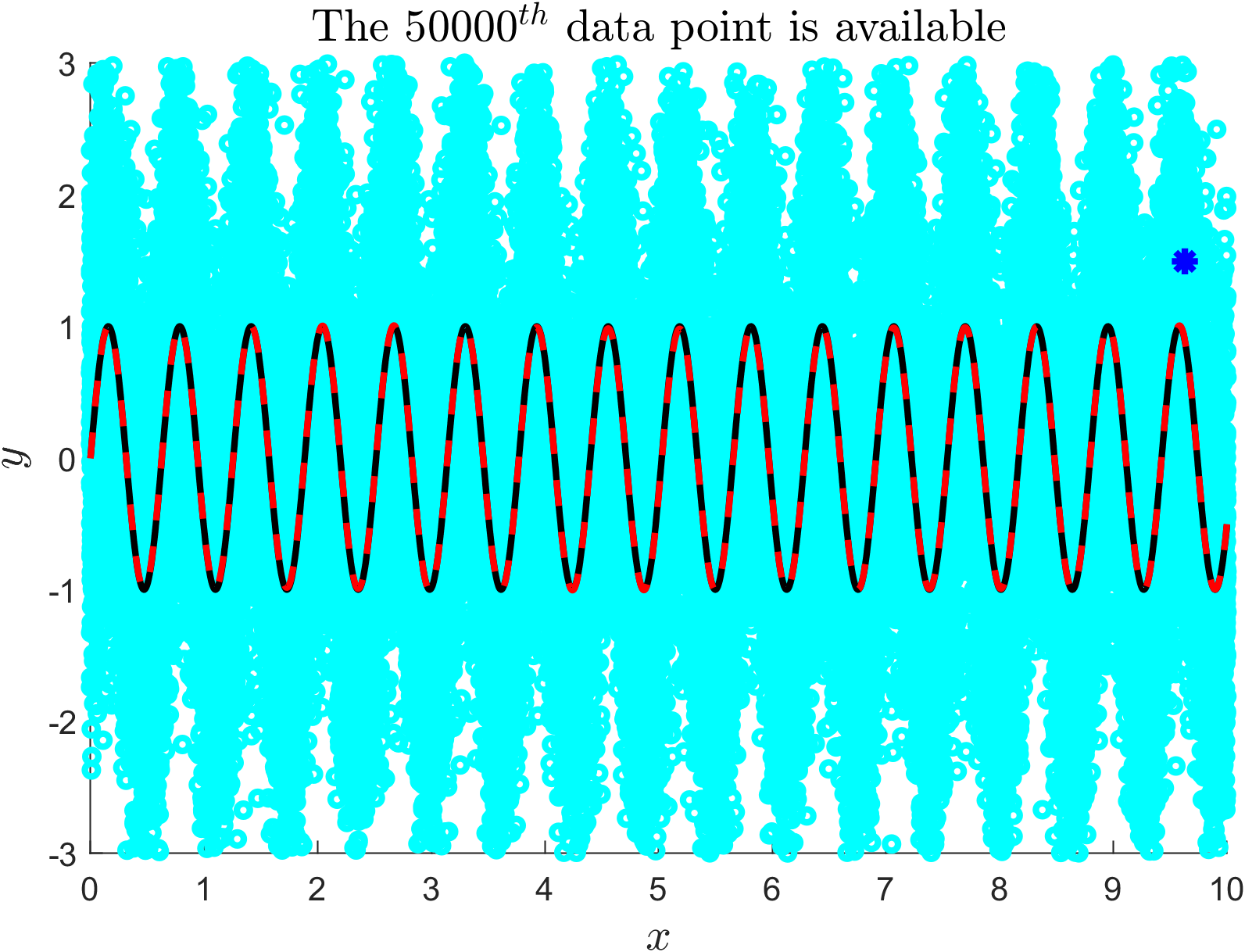}
        \caption{}
    \end{subfigure}
    \caption{Evolution and continual learning of the function approximation learned using our Riccati-based approach as more data becomes available. (a) shows the evolution of the learned coefficients $\weight_k, k=1,...,n$ and the relative $L^2$ error of the inferred $y$ as more data is incorporated into the model; the horizontal dotted lines denote the exact reference values. (b) shows the inferences of $y$ after the 200th, 1,000th, and 50,000th noisy data point becomes available. Our Riccati-based approach allows us to incrementally update the learned coefficients as more data becomes available without accessing the previous data or re-training on the entire dataset, which provides both memory and computational advantages over conventional learning methods.}
    \label{fig:continual}
\end{figure}

\begin{table}[ht]
    \footnotesize
    \centering
    \begin{tabular}{|c|c|c|c|}
    \hline
         & $h=0.001$ & $h=0.0005$ & $h=0.0001$ \\
       \hline
       $\ell_1$ error of $\weightvec^*$ & $2.7814\times 10^{-7}$ & $4.9461\times 10^{-10}$ & $3.0801\times 10^{-12}$ \\
       \hline
       $\ell_1$ relative error of $\weightvec^*$ & $2.7196\times 10^{-7}$ & $4.8361\times 10^{-10}$ & $3.0116\times 10^{-12}$\\
       \hline
    \end{tabular}
\caption{Errors in computing the minimizer $\weightvec^*$ of the function approximation loss~\eqref{eq:continual:loss} using our Riccati-based approach. We use RK4 to solve the Riccati ODEs~\eqref{eqt:sequentialRiccatiODEs} and \eqref{eqt:regression_1Riccati} with double precision and various step sizes $h$. The reference is obtained by using the method of least squares to minimize the loss function~\eqref{eq:continual:loss} directly. }
    \label{tab:continual}
\end{table}

Although this loss function~\eqref{eq:continual:loss} could be minimized using conventional machine learning techniques (e.g., the method of least squares), these methods typically require access to and training on the entire dataset $\{(x_i, y_i)\}_{i=1}^N$, which conflicts with the assumptions in continual learning.
However, note that this loss function~\eqref{eq:continual:loss} is of the form~\eqref{eq:loss_function}. Thus, we can instead apply our Riccati-based approach (as described in Sections~\ref{subsec:method_1} and~\ref{subsec:method_2}) to solve this learning problem. In particular, the methodology described in Section~\ref{subsec:method_2} matches the data streaming paradigm of continual learning; we incrementally update our learned coefficients by considering each new data point as the time evolution of a corresponding multi-time HJ PDE. 
As a result, in contrast to conventional machine learning methods, our Riccati-based approach does not require storage of previous data points, and its memory and computational complexity is constant for each added data point. As such, our Riccati-based methodology may be well-suited to online learning applications.
Additionally, since this time evolution of the HJ PDE (i.e., the addition of a new data point) requires knowledge about the solution to the HJ PDE at the previous time (i.e., the results of the previous training), our Riccati-based approach also inherently avoids catastrophic forgetting.

The results of applying our Riccati-based method to solve this learning problem~\eqref{eq:calibration:loss} are shown in Figure~\ref{fig:continual} and Table~\ref{tab:continual}. 
Figure~\ref{fig:continual}(a) depicts the evolution of the coefficients $\{\weight_k\}_{k=1}^n$ as more data points become available. We observe that the learned coefficients $\weight_k, k=1,...,n$ converge to their true values as more data points are incorporated into the model.
Figure~\ref{fig:continual}(b) displays the inferences obtained when the $200$th, $1,000$th, and $50,000$th noisy data point become available, which demonstrates that our Riccati-based approach is capable of real-time inferences without catastrophic forgetting, even though each inference is made using only one data point. 
As shown, the regression using data with high-level noise becomes more accurate as more data is incorporated into the learning process.

Table~\ref{tab:continual} displays the numerical errors of the minimizer $\weightvec^*$ of the loss function~\eqref{eq:continual:loss} obtained using the Riccati-based methodology from Section~\ref{sec:method} after the last data point becomes available. We observe that the accuracy of our approach increases as we decrease the step size $h$ of RK4, which indicates that the errors of $\weightvec^*$ stem from the accuracy of RK4 in solving the corresponding Riccati ODEs.
The reference solution is obtained by minimizing~\eqref{eq:continual:loss} directly using the method of least squares and assuming that all $N=50,000$ data points are accessible.

%% file: example_calibration.tex
\subsection{1D steady-state reaction-diffusion equation and post-training calibration}\label{subsec:example_2}
In this example, we use our Riccati-based approach to apply post-training calibrations when solving a PDE. Specifically, we leverage the methodology in Section~\ref{subsec:method_2} to add or remove data and the methodology in Section~\ref{subsec:method_3} to enforce the boundary conditions of the PDE without retraining the entire learned model.
Consider the following 1D steady-state reaction-diffusion equation:
\begin{equation}\label{eq:reaction}
\begin{dcases}
  D\frac{\partial^2 u}{\partial x^2}(x) + \kappa u(x) = f(x), x\in [0, 1],\\
  u(0) = u(1) = 0,
\end{dcases}
\end{equation}
where $D=0.01$ is the diffusion coefficient, $\kappa=-1$, and $f(x)$ is the source term of which noisy measurements $\{(x_i, f_i\approx f(x_i))\}_{i=1}^N\subset\R\times\R$ are available.
We consider the scenario where regular training has been employed but with either insufficient data or sufficient data with outliers, both of which yield inaccurate inferences of the solution. 
We further assume that extra information is provided after the regular training and seek to perform post-training calibrations to incorporate this extra information into the already-trained models without losing information from the original training.
In the literature, it is well-established that post-training calibrations can significantly increase the performance of deployed machine learning methods \cite{psaros2023uncertainty, zou2022neuraluq}.
However, designing computationally efficient methods for performing these post-calibrations is still of great interest.

In this example, we solve this PDE~\eqref{eq:reaction} by reformulating the PDE as an optimization problem \cite{raissi2019physics, sirignano2018dgm, han2018solving}.
We use a linear model to approximate the solution, i.e. $u(x) = \sum_{k=1}^n \weight_k\phi_k(x)$, where $n=21$ and $\{\phi_k(x)\}_{k=1}^n=\{1\}\cup\{\sin(2l\pi x),$ $ \cos(2l\pi x)\}_{l=1}^{(n-1)/2}$ are the truncated Fourier basis functions on $[0, 1]$. 
We learn the coefficients $\weightvec = [\weight_1, \dots, \weight_n]^T$ of the linear model by minimizing the following loss:
\begin{equation}\label{eq:calibration:loss}
\begin{aligned}
    \mathcal{L}(\weightvec) & = \frac{1}{2}\sum_{i=1}^N \lambda_i \Bigg|D\sum_{k=1}^{n} \weight_k\frac{\partial^2 \phi}{\partial x^2}(x_i) + \kappa \sum_{k=1}^{n} \weight_k \phi_k(x_i) - f_i \Bigg|^2 \\
    & + \frac{1}{2}\lambda_b\left|\sum_{k=1}^{n} \weight_k \phi_k(0) - 0\right|^2 + \frac{1}{2}\lambda_b\left|\sum_{k=1}^{n} \weight_k \phi_k(1) - 0\right|^2 + \frac{1}{2}\sum_{k=1}^{n} \MLreg_k|\weight_k|^2,
\end{aligned}
\end{equation}
where $\lambda_i, i=1,...,N$, $\lambda_b$, and $\MLreg_k, k=1,..,n$ are balancing weights for the PDE residual, the boundary conditions, and the regularization term, respectively. In our numerical experiments, we assume the exact solution to be $u(x) = \sin^3(2\pi x)$ (and $f$ to be defined by~\eqref{eq:reaction}, accordingly) and the noise to be additive Gaussian with zero mean and standard deviation $0.1$. For the regular training, we set $\lambda_i=1$, $\lambda_b=1$, and $\MLreg_k=1$ and apply our Riccati-based approach (see Section~\ref{subsec:method_1}) to get our original estimate of the minimizer $\weightvec^*$ of the loss function~\eqref{eq:calibration:loss}. Using our Riccati-based approach yields an  $\ell_1$ error of $8.9950\times 10^{-10}$ and relative $\ell_1$ error of $1.4288\times 10^{-9}$ in $\weightvec^*$, where the reference is obtained by minimizing~\eqref{eq:calibration:loss} directly using the method of least squares.

\begin{figure}[ht]
    \begin{subfigure}[b]{\textwidth}
        \centering
        \includegraphics[width = 0.3\textwidth]{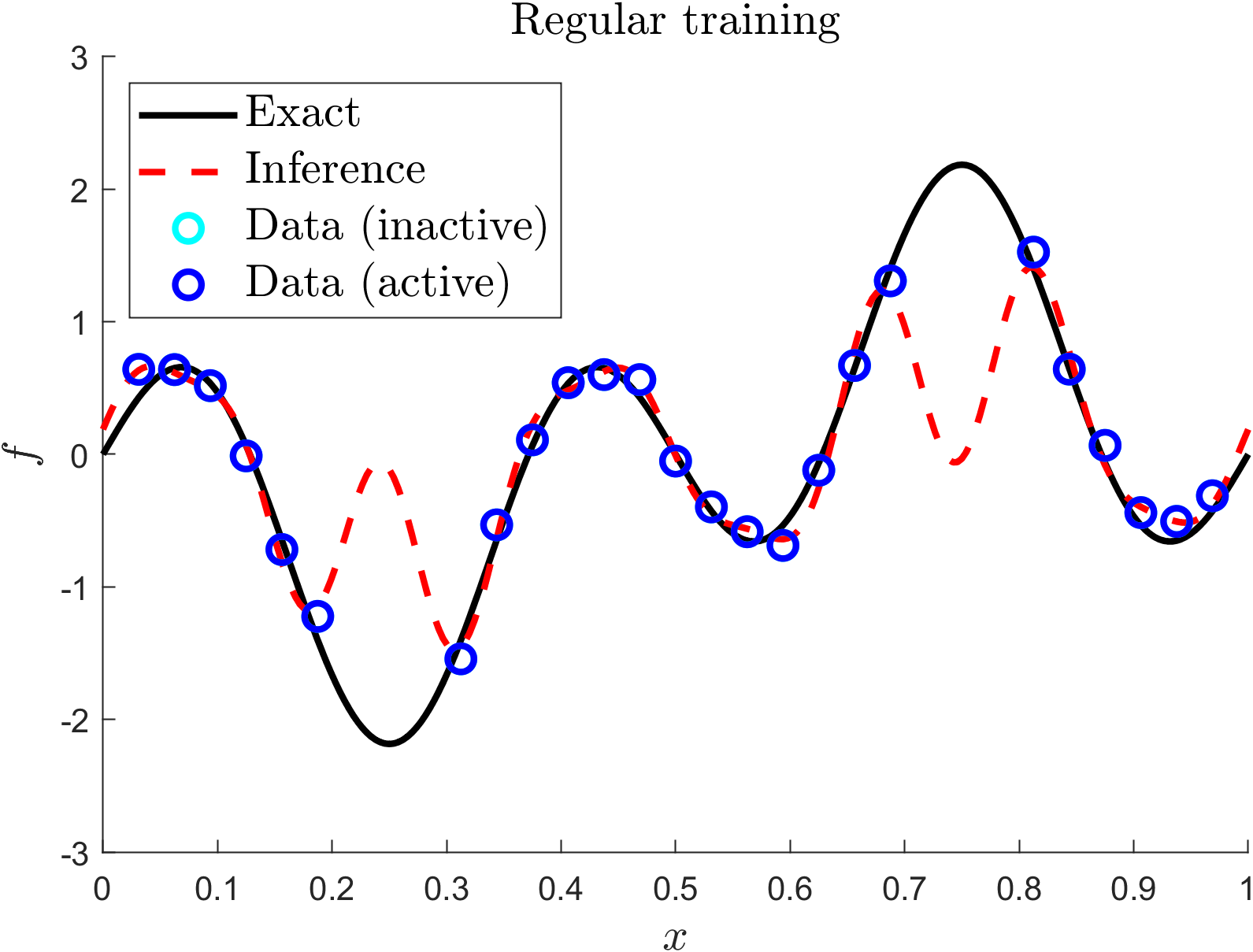}
        \includegraphics[width = 0.3\textwidth]{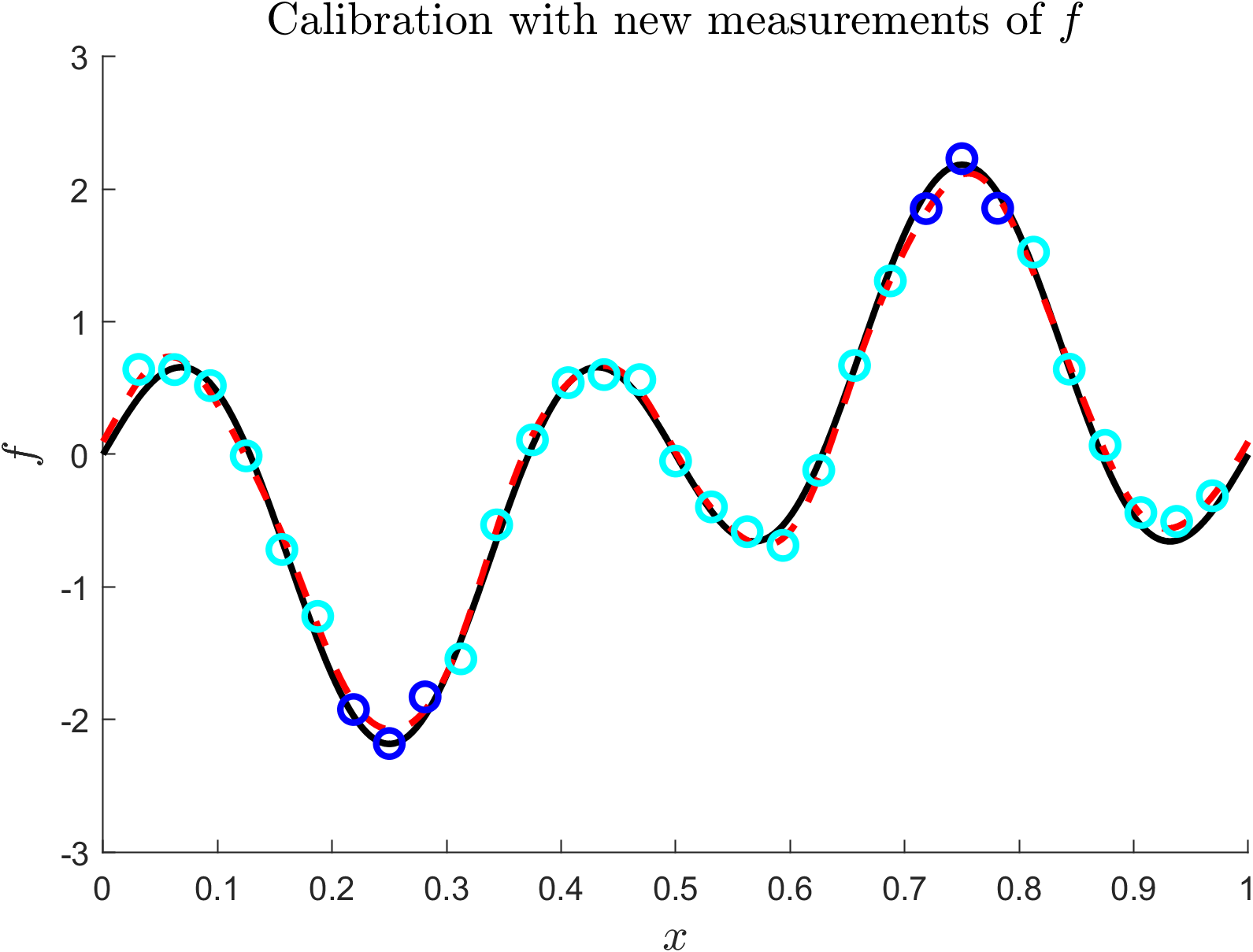}
        \includegraphics[width = 0.3\textwidth]{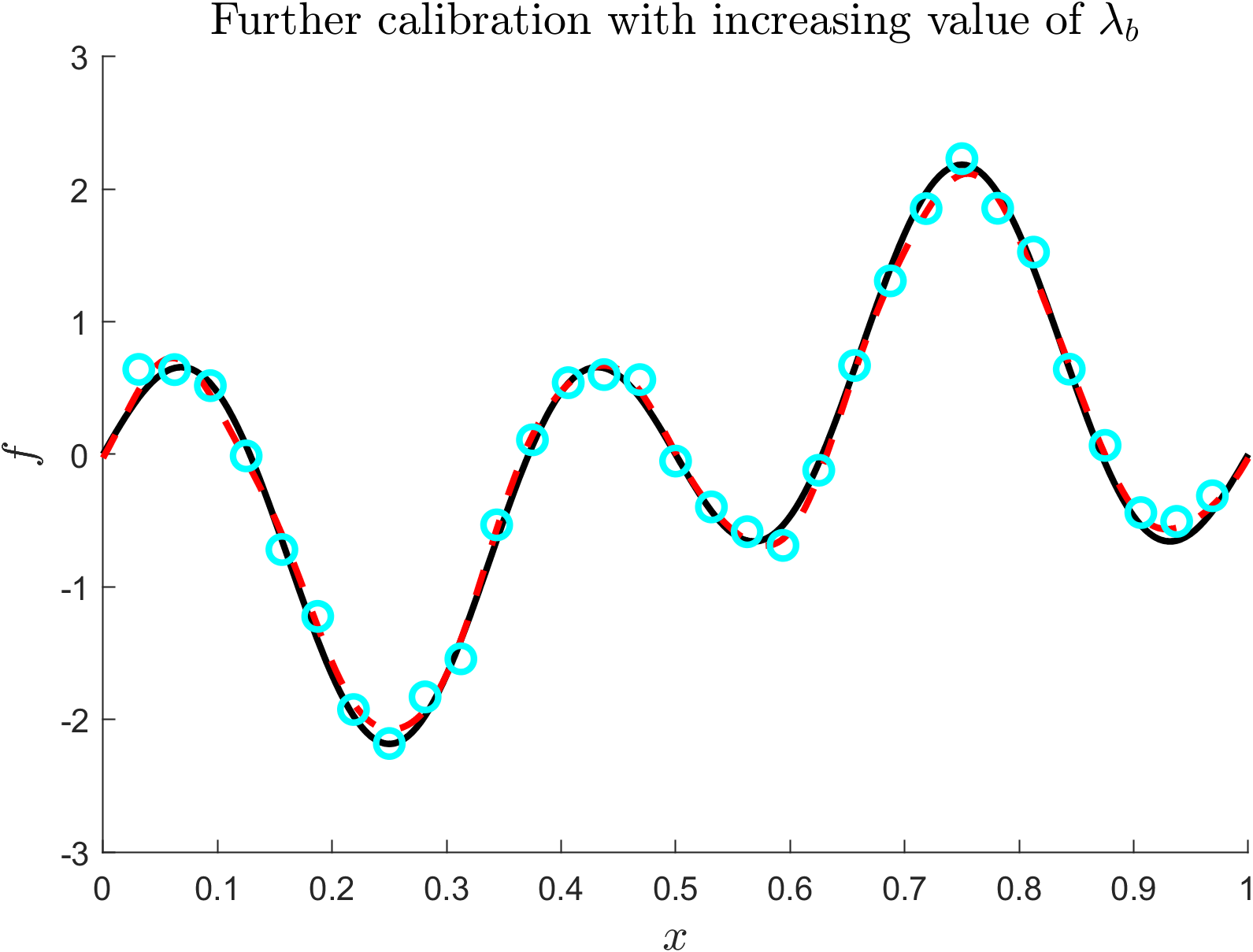}
        \caption{}
    \end{subfigure}
    \begin{subfigure}[b]{\textwidth}
        \centering
        \includegraphics[width = 0.3\textwidth]{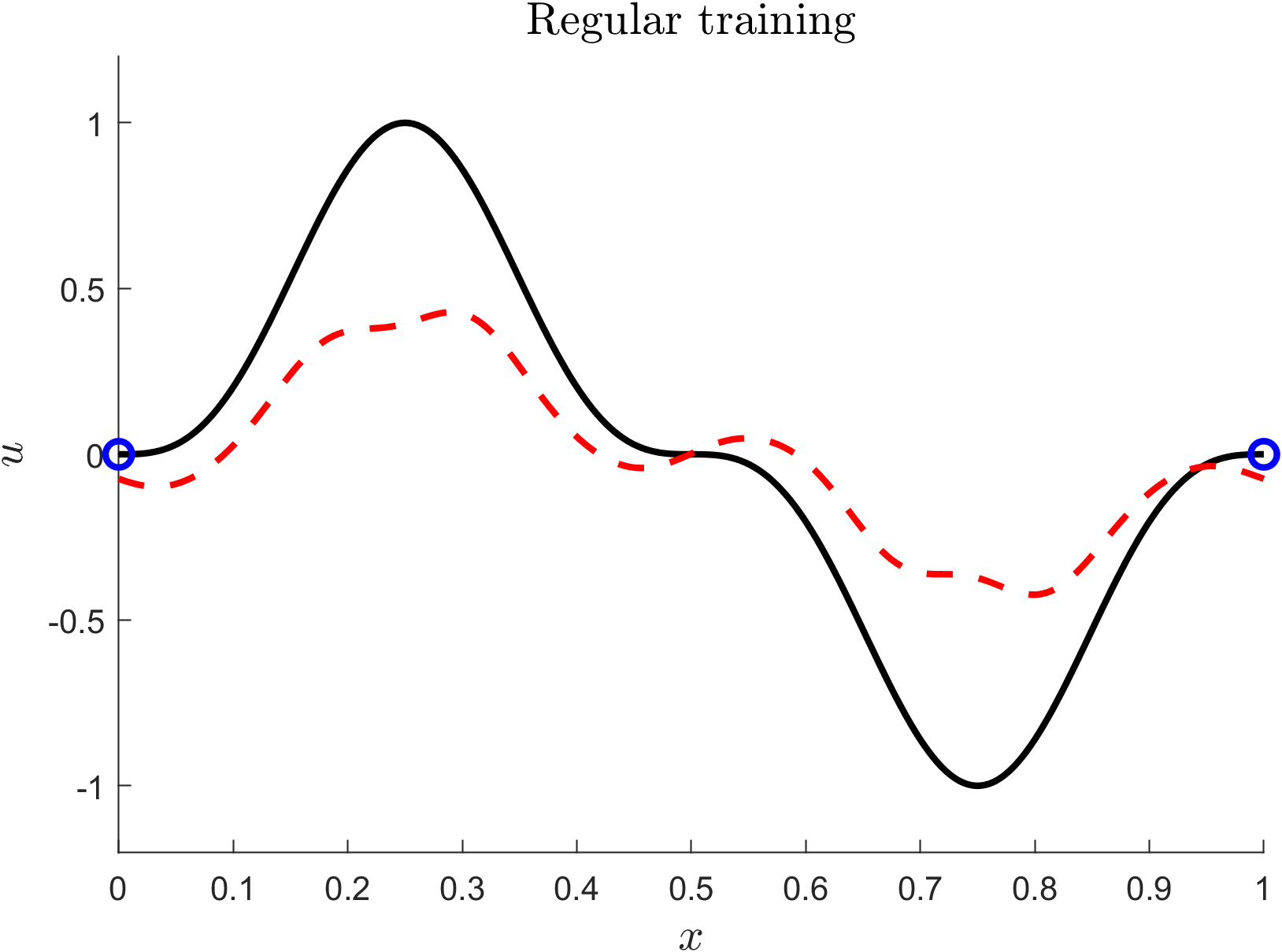}
        \includegraphics[width = 0.3\textwidth]{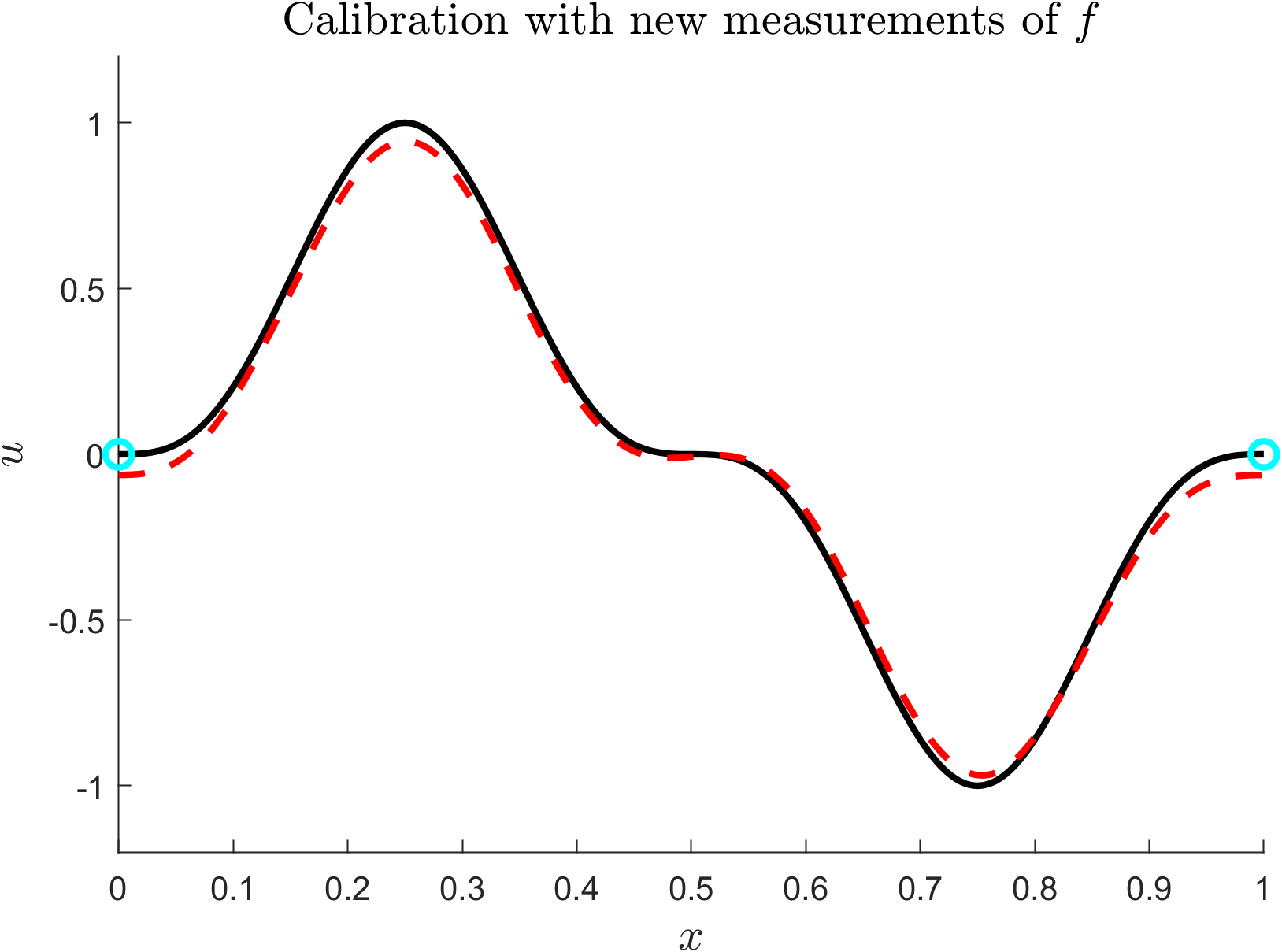}
        \includegraphics[width = 0.3\textwidth]{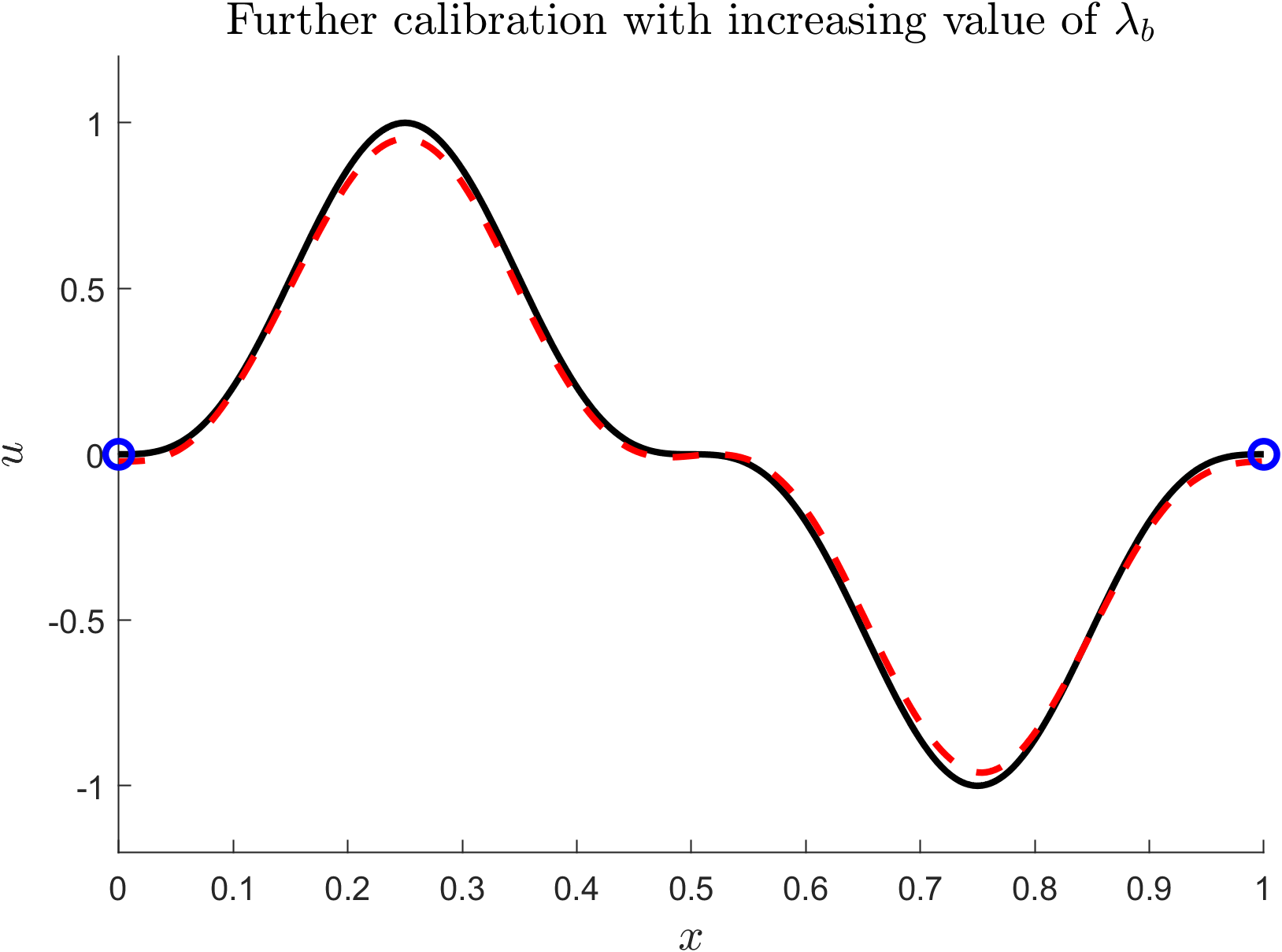}
        \caption{}
    \end{subfigure}
    \caption{Results of solving the 1D steady-state reaction-diffusion equation~\eqref{eq:reaction} with noisy measurements of the source term $f$ in the domain and noiseless measurements of the solution $u$ on the boundary. (a) results for $f$; (b) results for $u$. \textbf{Left}: results of regular training; \textbf{middle}: calibrating the results with some additional noisy measurements of $f$; \textbf{right}: calibrating the results further by enforcing the boundary conditions by increasing the value of $\lambda_b$ in the loss function~\eqref{eq:calibration:loss}. The regular training uses our Riccati-based method in Section~\ref{subsec:method_1} to minimize~\eqref{eq:calibration:loss}, while the calibrations use the adaptations of our method in Section~\ref{subsec:method_2}. Calibrations are employed without re-training or access to the data from the previous training, which demonstrates the advantages in both memory storage and computational complexity of our Riccati-based approach over conventional machine learning methods.}
    \label{fig:calibration:1}
\end{figure}

In the leftmost column of Figure~\ref{fig:calibration:1}, we see that the accuracy of both $u$ and $f$ as inferred by the regular training is impaired by a lack of data around the highest peak and lowest valley of the exact functions. 
To compensate, we first calibrate our model by adding some new noisy measurements of $f$ in these regions where the data points are sparse. This calibration uses the methodology described in Section~\ref{subsec:method_2}, and the results are shown in the middle column of Figure~\ref{fig:calibration:1}. 
Next, we note that the inferred $u$ still disagrees with the exact solution at the boundary points. 
Hence, we further calibrate our model by increasing the value of the boundary weight $\lambda_b$ from $1$ to $10$ to enforce the boundary conditions of the PDE. This calibration is done using the methodology described in Section~\ref{subsec:method_3} and the results of this second calibration are presented in the rightmost column of Figure~\ref{fig:calibration:1}. In both cases, we observe that the calibrations successfully improve the accuracy of the learned model. \updatethree{These results are also reflected in the relative $L^2$ errors shown in Table~\ref{tab:calibration:1}.}

\begin{table}[ht]
    \footnotesize
    \centering
    \begin{tabular}{|c|c|c|c|}
    \hline
         & Regular training & First calibration & Second calibration \\
       \hline
       relative $L^2$  error of $f$ & $66.40\%$ & $7.15\%$ & $6.49\%$\\
       \hline
      relative $L^2$  error of $u$ & $57.29\%$ & $7.35\%$ & $5.57\%$\\
       \hline
    \end{tabular}
\caption{Errors in solving the 1D steady-state reaction-diffusion equation~\eqref{eq:reaction} and regressing $f$ at different stages of training, using our Riccati-based approach. Regular training is done with insufficient data and provides inaccurate inferences. The first calibration adds new measurements of $f$, and the second calibration increases the weights $\lambda_b$ of the boundary conditions in \eqref{eq:calibration:loss}. Both calibration steps successfully improve the accuracy of our inferences without requiring retraining on or access to the previous data. Qualitative results can be found in Figure~\ref{fig:calibration:1}.}
\label{tab:calibration:1}
\end{table}

Note that our Riccati-based approach allows us to perform each of these calibration steps using only the new or changed values in that step and the results of the previous training step. In other words, each step of this training process (including the original training and each subsequent calibration step) is done without sharing data between training steps, which exactly matches the framework of federated learning \cite{li2020federated}. Thus, our methodology may be relevant to distributed training or collaborative learning applications, where data privacy is of concern.

Next, we discuss another post-training calibration technique.
In this case, we assume the data for the regular training is sufficient but contains outliers due to large noise. We again assume the regular training is performed using the Riccati-based approach from Section~\ref{subsec:method_1}. Then, we eliminate these outliers using the methodology described in Section~\ref{subsec:method_2}, which only requires knowledge about the outliers to be removed and the results of the regular training.
The results of this post-training outlier removal show that eliminating these points successfully improves the accuracy of the learned models (see Appendix \ref{appendix:1}, Figure~\ref{fig:calibration:2}). \updatethree{Finally, in Appendix~\ref{appendix:1}, Figure~\ref{fig:example_2:3}, we also provide a large-scale continual learning example for solving~\eqref{eq:reaction} to further demonstrate the performance of our Riccati-based approach in large data settings.}

%% file: example_poisson.tex
\subsection{Poisson equation using PINNs and transfer learning}\label{subsec:example_3}
In this example, we demonstrate the versatility of our Riccati-based method by combining it with existing machine learning techniques to fit the last layer of a PINN. We also show that when we perform hyper-parameter tuning by solving the associated Riccati ODEs (Section~\ref{subsec:method_3}), we not only provide the solution to the updated problem but also a continuum of solutions along a 1D curve on the Pareto front of the data fitting losses and regularization.
Consider the 2D Poisson equation with Dirichlet boundary conditions, which is given by
\begin{equation}\label{eq:poisson}
\begin{dcases}
  \frac{\partial^2 u}{\partial x^2}(x,y) + \frac{\partial^2 u}{\partial y^2}(x,y) = f(x,y) & (x, y) \in \Omega,\\
  u(x, y) = 0 & (x, y) \in \partial \Omega,
\end{dcases}
\end{equation}
where $\Omega := [0, 1]^2$ and $f$ is a source term. We solve this equation using transfer learning and PINNs. Consider the scenario where we only have access to measurements $\{(x_i, y_i, f_i = f(x_i, y_i))\}_{i=1}^N$ of $f$ at limited sampling points. Transfer learning compensates for this lack of knowledge by transferring the knowledge from models pre-trained on similar problems to  solve this new problem of interest. Here, we learn the linear model $u(x,y) = \sum_{k=1}^n\weight_k\phi_k(x,y)$, where each basis function $\phi_k(x, y), k=1, \dots, n$ is the PINN solution of a neural network 
pre-trained to solve the 2D Poisson equation~\eqref{eq:poisson} with similar source terms. 
Transfer learning using pre-trained neural networks as basis functions, as we do here, has recently grown in popularity in the scientific machine learning community \cite{desai2021one, zou2023hydra, goswami2022deep} and has been shown to provide efficient yet accurate inferences even given very limited data \cite{zou2023hydra}. 
We learn the coefficients $\weightvec$ of our linear model by minimizing the following PINN-type loss:
\begin{equation}\label{eq:loss:poisson}
    \mathcal{L}(\weightvec) = \frac{1}{2}\sum_{i=1}^N \lambda_i \left|\sum_{k=1}^n \weight_k\left(\frac{\partial^2\phi_k}{\partial x^2} + \frac{\partial^2\phi_k}{\partial y^2}\right)(x_i, y_i) - f_i \right|^2 + \frac{1}{2}\sum_{k=1}^n \MLreg_k|\weight_k|^2,
\end{equation}
where $\lambda_i=1, i=1,...,N$ and $\MLreg_k = \MLreg, k=1,...,n$ are weights for the data fitting and $\ell_2$-regularization terms, respectively. Note that minimizing~\eqref{eq:loss:poisson} with respect to $\weightvec$ is equivalent to fitting the last layer of a neural network given previous, pre-trained  nonlinear layers as the basis functions.

In our numerical experiments, we use transfer learning to solve the 2D Poisson equation~\eqref{eq:poisson}  with source term  $f(x, y) = \sin(2.5\pi x)\sin(2.5\pi y)$ using $n=100$ basis functions and $N=100$ random measurements of $f$.
We use the multi-head PINN method \cite{zou2023hydra} to obtain the basis functions, which correspond to the shared nonlinear hidden layers of the pre-trained PINN solutions to~\eqref{eq:poisson} with source terms $f(x, y) = \sin(k\pi x)\sin(k\pi y), k=1,2,3,4$. \updatetwo{We note that the boundary condition is hard-encoded in the basis functions, and hence, in \eqref{eq:loss_function}, no penalty term involving the boundary condition is included.} We then solve the learning problem~\eqref{eq:loss:poisson} using our Riccati-based method from Section~\ref{subsec:method_1}.

In Table~\ref{tab:poisson}, we compare the errors of the minimizer $\weightvec^*$ of~\eqref{eq:loss:poisson} and the solution $u$ of the PDE~\eqref{eq:poisson} as we decrease the weight $\MLreg (=\MLreg_k,\forall k)$ of the regularization term from 1 to  1e-5. The reference for $\weightvec^*$ is obtained from minimizing~\eqref{eq:loss:poisson} directly for each value of $\MLreg$ using the method of least squares. The reference for $u$ is computed using a finite difference method with a five-point stencil \updatetwo{and a $257\times257$ uniform grid on $\Omega$} to solve~\eqref{eq:poisson}. The same grid is used to evaluate our trained models. 
We originally minimize~\eqref{eq:loss:poisson} using $\MLreg = 1$ and the Riccati-based approach in Section~\ref{subsec:method_1}. We then compute the solutions for the other values of $\MLreg$ by incrementally decreasing $\MLreg$ by a factor of 10 and using the methodology from Section~\ref{subsec:method_3} to reuse the results of training with the previous value of $\MLreg$ to compute the solution for the new value of $\MLreg$. Consequently, we see that the error in $\weightvec^*$ increases as we decrease $\MLreg$ due to error accumulation from repeated applications of RK4. However, the error of $u$ generally decreases as we decrease $\MLreg$ with the lowest error being achieved when $\MLreg=$ 1e-4.

\begin{table}[ht]
    \footnotesize
    \centering
    \begin{adjustbox}{width=\textwidth}
    \begin{tabular}{|c|c|c|c|c|c|c|}
    \hline
         & $\MLreg = \updatetwo{10^0}$ & $\MLreg = \updatetwo{10^{-1}}$ & $\MLreg =\updatetwo{10^{-2}} $ & $\MLreg =\updatetwo{10^{-3}} $ & $\MLreg = \updatetwo{10^{-4}}$ & $\MLreg=\updatetwo{10^{-5}}$  \\
         \hline
       $\ell_1$ error of $\weightvec^*$ & $5.9707\times 10^{-10}$ & $4.2905\times 10^{-8}$ & $1.0725\times 10^{-6}$ & $1.7235 \times 10^{-5}$ & $2.0490\times 10^{-4}$ & $5.8749\times 10^{-3}$\\
       \hline
       $L^2$ relative error of $u$ & $6.5247\%$ & $5.6793\%$ & $3.0390\%$ & $1.7280\%$ & $1.1662\%$ & $1.4736\%$ \\
       \hline
    \end{tabular}
    \end{adjustbox}
    \caption{Errors of the minimizer $\weightvec^*$ of~\eqref{eq:loss:poisson} and the solution $u$ to the 2D Poisson equation~\eqref{eq:poisson} using transfer learning and our Riccati-based approach. The reference for $\weightvec^*$ is given by minimizing~\eqref{eq:loss:poisson} with the corresponding value of $\MLreg$ directly using the method of least squares, and the reference for $u$ is given by solving~\eqref{eq:poisson} using a finite difference method. Since we decrease $\MLreg$ incrementally from 1 to 1e-5, the error of $\weightvec^*$ accumulates due to successive applications of RK4.} 
    \label{tab:poisson}
\end{table}

\begin{figure}[htbp]
    \begin{subfigure}[b]{\textwidth}
        \centering
        \includegraphics[width = 0.4\textwidth]{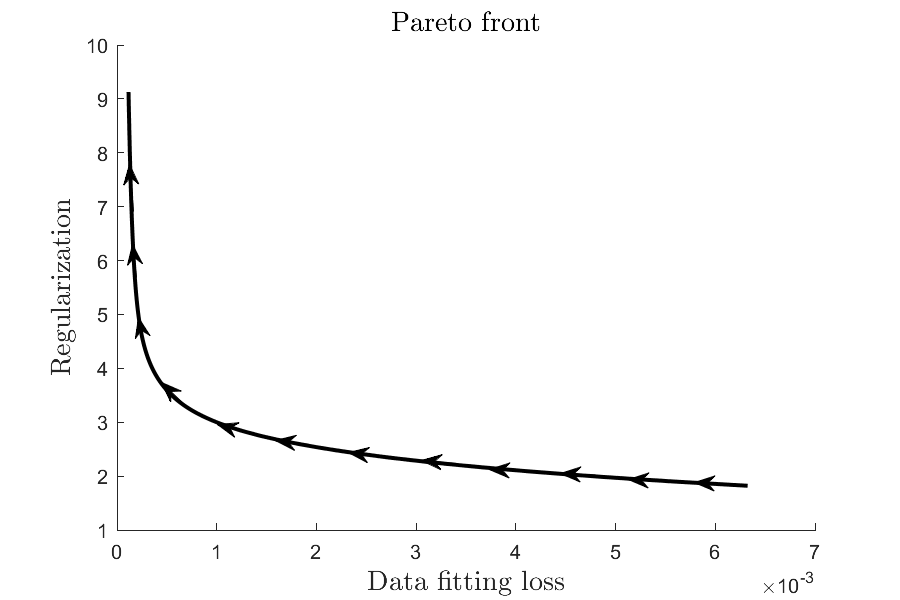}
        \includegraphics[width = 0.4\textwidth]{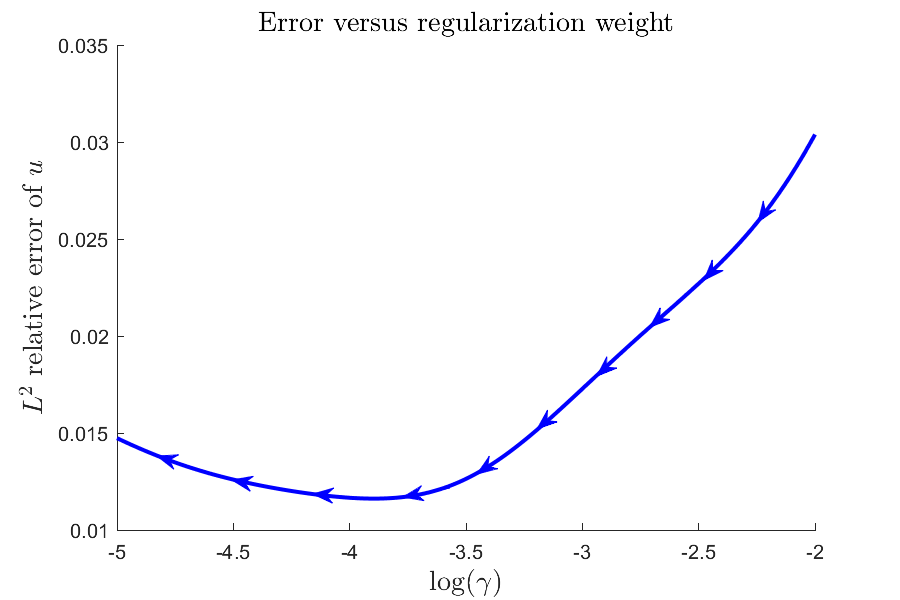}
    \end{subfigure}
    \caption{Results of changing the regularization weight $\MLreg$ when solving the 2D Poisson equation using PINNs and transfer learning. We incrementally decrease the value of $\MLreg$ by evolving the corresponding Riccati ODEs backwards in time. As a result, we obtain a flow of solutions, the direction of which is represented by the arrows in each figure. In the left figure, this flow of solutions gives us that each change in the value of $\MLreg$ from $\hat\MLreg$ to $\tilde\MLreg$ results in the recovery of every point of the Pareto front along the one-dimensional curve parameterized by $\MLreg\in[\hat\MLreg, \tilde\MLreg]$.}
    \label{fig:poisson}
\end{figure}

From the results in Table~\ref{tab:poisson}, we see that our choice of the hyper-parameter $\MLreg$ can greatly influence the accuracy of our learned model. Note that since we fix $\lambda_i = 1, \forall i$ and 
$\MLreg_k = \MLreg, \forall k$, we can view~\eqref{eq:loss:poisson} as a bi-objective loss, where the two objectives are the weighted data fitting term $\frac{1}{2}\sum_{i=1}^N \left|\sum_{k=1}^n \weight_k\left(\frac{\partial^2\phi_k}{\partial x^2} + \frac{\partial^2\phi_k}{\partial y^2}\right)(x_i, y_i)- f_i\right|^2$ and the regularization term $ \frac{1}{2}\sum_{k=1}^n|\weight_k|^2$. 
To better understand the effects of tuning $\MLreg$, in Figure~\ref{fig:poisson}, we explore the Pareto front of these two objectives. Traditional scalarization-based approaches for computing the Pareto front typically rely on discrete samplings of the Pareto front corresponding to discrete choices of $\MLreg$ \cite{Jin2008pareto}.
While our Riccati-based methodology from Section~\ref{subsec:method_3} also recovers discrete points on the Pareto front corresponding to particular choices of $\MLreg$, note that when we change $\MLreg$, e.g., from $\MLreg=1$ to $\MLreg=0.1$, we also recover a one-dimensional curve along the Pareto front corresponding to every  $\MLreg\in[0.1, 1]$. We obtain this 1D curve theoretically via the flow of solutions obtained from the corresponding Riccati ODEs and numerically via the intermediate steps of RK4.
The left plot of Figure~\ref{fig:poisson} shows the 1D curve along the Pareto front recovered by our Riccati-based approach (although note that in this example, the Pareto front is also one-dimensional and hence is equivalent to the exposed 1D curve), where the flow of solutions corresponding to decreasing $\MLreg$ is represented by the arrows.
Thus, although in general our methodology cannot compute the entire Pareto front, every time we change the value of the hyper-parameters, our approach recovers a continuous 1D curve along the Pareto front. In the right plot of Figure~\ref{fig:poisson}, we also visualize how the $L^2$ error of our learned solution $u$ changes as we decrease $\MLreg$.

%% file: example_sindy.tex
\subsection{Identifying the dynamics of the Kraichnan-Orszag system from data}\label{subsec:example_4}
In this example, we demonstrate the versatility of our Riccati-based approach by showing how it can be combined with existing methods to solve more general problems (see Section~\ref{subsec:method_4}).
Consider the Kraichnan-Orszag (K-O) system \cite{wan2006multi, zou2022neuraluq, zhang2023discovering}
\begin{equation}\label{eq:ko}
\begin{dcases}
  \frac{dx_1}{d\tau} = x_2x_3,\\
  \frac{dx_2}{d\tau} = x_1x_3,\\
  \frac{dx_3}{d\tau} = -2x_1x_2,
\end{dcases}
\end{equation}
with initial conditions $x_1(0) = 1, x_2(0) = 0.8, x_3(0) = 0.5$. Our goal is to identify the dynamics (the right-hand side of~\eqref{eq:ko}) of the K-O system  using measurements of $x_i$ and $\frac{d x_i}{d\tau}, i=1,2,3$ at different times. We identify the dynamics by learning the linear models $\frac{dx_i}{d\tau} = \sum_{k=1}^n\weight_k^i\phi_k, i = 1, 2, 3$. Following the general framework of the SINDy method \cite{brunton2016discovering}, we use the following quadratic basis functions ($n = 10$) for the dynamics:
\begin{equation*}\label{eq:ko:basis}
    \{\MLbasis_k(x_1, x_2, x_3)\}_{k=1}^n = \{1, x_1, x_2, x_3, x_1^2, x_2^2, x_3^2, x_1x_2, x_2x_3, x_1x_3\},
\end{equation*}
and impose $\ell_1$-regularization on $\weightvec$ to promote sparse identification of the dynamics. 
Then, we learn each $\weightvec^i$ independently and in parallel by minimizing the loss functions
\begin{equation}\label{eq:ko:loss}
    \mathcal{L}_i(\weightvec^i) = \frac{1}{2}\sum_{j=1}^N \lambda_j \left[\left(\frac{dx_i}{d\tau}\right)_j - \sum_{k=1}^{n} \weight^i_k \phi_k((x_1)_j, (x_2)_j, (x_3)_j)\right]^2 + \sum_{k=1}^n \MLreg_k|\weight_k^i|,
\end{equation}
where $\mathcal{L}_i$ denotes the loss function for equation $i$, $(x_i)_j$ and $(\frac{dx_i}{dt})_j$ denote the measurements of $x_i$ and $\frac{dx_i}{d\tau}$, respectively, at time $\tau_j$, $i=1,2,3,j=1,...,N$. Note that \eqref{eq:ko:loss} corresponds to setting $R = \|\cdot\|_1$ and $\HJx = 0$ in the linear regression problem~\eqref{eqt:loss_fn_general_reg}. Thus, solving this learning problem is equivalent to evaluating the solution to the corresponding multi-time HJ PDE at $(0, \lambda_1, \dots, \lambda_n)$. 
In our numerical experiments, we generate data points for training and testing by solving~\eqref{eq:ko} numerically for $x_1, x_2, x_3\in[0, 10]$ \updatetwo{using MATLAB \textit{ode45} \cite{MATLAB} and then} using a central finite difference scheme to approximate the time derivatives. We set $\lambda_j=1, \forall j$ and $\MLreg_k=0.1, \forall k$.

Instead of the sparse regression techniques employed by SINDy, 
we use PDHG to minimize~\eqref{eq:ko:loss} (see Section~\ref{subsec:method_4}). Each iteration of PDHG involves minimizing a loss function of the form~\eqref{eq:ko:loss}, but with $\ell_2$-regularization instead of $\ell_1$. Hence, this sub-problem can be solved using our Riccati-based methods. As discussed in Section~\ref{subsec:method_4}, note that we only need to apply RK4 for the first iteration of PDHG, and every subsequent iteration can be solved using a change of bias. As such, we do not suffer from any error accumulation related to repeated applications of RK4.
In Table~\ref{tab:sindy:1}, we see that we do indeed recover a sparse identification of the dynamics. However, we also incorrectly identify non-zero coefficients for the basis functions $x_2$ and $x_3$. We note that this misidentification may be the result of a lack of unique identifiability of the system from the data points sampled. In fact, Table~\ref{tab:sindy:3} shows that the errors in the solution $x_1,x_2,x_3$ of the identified system versus the solution of the true system~\eqref{eq:ko} are relatively small, which corroborates that identifiability may have been an issue.

\begin{table}[ht]
    \footnotesize
    \centering
    \begin{tabular}{|c|c|c|c|c|c|c|c|c|c|c|}
    \hline
         & $1$ & $x_1$ & $x_2$ & $x_3$ & $x_1^2$ & $x_2^2$ & $x_3^2$ & $x_1x_2$ & $x_2x_3$ & $x_1x_3$ \\
         \hline
         $\weightvec^{1, *}$ & $0$ & $0$ & $0$ & $0$ & $0$ & $0$ & $0$ & $0$ & $0.9931$ & $0$\\
         \hline
         $\weightvec^{2, *}$ & $0$ & $0$ & $0$ & $0.0160$ & $0$ & $0$ & $0$ & $0$ & $0$ & $0.9761$\\
         \hline
         $\weightvec^{3, *}$ & $0$ & $0$ & $-0.0165$ & $0$ & $0$ & $0$ & $0$ & $-1.9777$ & $0$ & $0$\\
         \hline
    \end{tabular}
\caption{Results of sparse identification of the K-O system~\eqref{eq:ko} using PDHG to minimize the $\ell_1$-regularized losses~\eqref{eq:ko:loss}. The true solution is 0 for all entries, except $\weightvec^{1, *} = 1$ for $x_2x_3$, $\weightvec^{2, *} = 1$ for $x_1x_3$, and $\weightvec^{3, *} = -2$ for $x_1x_2$. We recover the dynamics reasonably well, albeit with some slight misidentification of $\weightvec^{2, *}, \weightvec^{3, *}$.}
    \label{tab:sindy:1}
\end{table}

\begin{table}[ht]
    \footnotesize
    \centering
    \begin{tabular}{|c|c|c|c|}
    \hline
         & $x_1$ & $x_2$ & $x_3$ \\
       \hline
       relative $L^2$ error (\%) & $0.2144$ & $0.3718$ & $0.3152$\\
       \hline
    \end{tabular}
    \caption{Errors of the solution $x_1, x_2, x_3$ of the system identified using PDHG compared to the true solution of the K-O system. The reference is obtained by numerically solving the true system~\eqref{eq:ko} using a central finite difference scheme. These errors indicate that the errors in the system identification in Table~\ref{tab:sindy:1} may be due to a lack of unique identifiability of the system using the given data points.}
    \label{tab:sindy:3}
\end{table}

%% file: conclusion.tex
\section{Summary}\label{sec:conclusion}
In this paper, we established a novel theoretical connection between \updateone{regularized} learning problems and the multi-time Hopf formula. In doing so, we showed that when we solve these learning problems, we actually solve certain multi-time HJ PDEs and their corresponding optimal control problems. In this work, we focused on the development of the connection between regularized linear regression and the LQR problem. By leveraging this connection, we developed new methodology based on solving Riccati ODEs that allows us to design new training approaches for certain machine learning applications, such as continual learning, post-training calibration, hyper-parameter tuning and exploration of the associated Pareto front, and sparse dynamics identification. We also showed that our Riccati-based approach yields some promising computational advantages over conventional learning methods; after the original training, the models learned using our approach can be continually updated without having to retrain the entire model or having access to all of the previous data, which could be particularly useful in continual learning \cite{parisi2019continual, kirkpatrick2017overcoming, van2019three} and federated learning \cite{li2020federated, kairouz2021advances}.
We note that our connection shares some similarities to the approach of NeuralODEs \cite{chen2018neural}, which also connects ODEs to machine learning; however, NeuralODEs focuses on establishing new neural networks architectures using ODEs, whereas we make connections between (Riccati) ODEs (that arise from optimal control theory) and the training/optimization process of existing machine learning methods and neural networks.
\updateeditor{Our connection is also related to~\cite{todorov2008general}, which connects LQR problems with Kalman filters; however,~\cite{todorov2008general} does not make any connections to PDEs or learning problems.}

Thus, our theoretical connection and Riccati-based methodology present many exciting opportunities. Some possible future directions are as follows. While our Riccati-based methodology allows us to alter the hyper-parameters and data points used in the learning problem without having to retrain on all previous data, it also requires that the original training be done using our Riccati-based methodology. It would allow for increased versatility if we could more easily combine our Riccati-based approach with other training methods. 
Additionally, in Sections~\ref{subsec:method_4} and~\ref{subsec:example_4}, we showed that our Riccati-based approach allows computations to be reused when using non-quadratic regularizations. However, in this case, the training process still had to be restarted if the hyper-parameters, dataset, or regularization type is changed. It would allow for more flexibility if we could develop more adaptive processes for changing these aspects of the learning problem when the regularization is not quadratic. 

In Section~\ref{sec:LQR}, we focused on LQR problems, where the dynamics are independent of the trajectory, but it would be worthwhile to investigate what connections LQR problems with  general linear dynamics may yield (e.g., LQR problems with general linear dynamics may be reformulated as LQR problems with state-independent dynamics using a change of variable \cite{Darbon2016Algorithms}). 
Another natural extension would be to consider nonlinear models, which currently pose challenges in both scientific machine learning and optimal control and hence, connections drawn in this case would benefit both fields.
\updateone{It would also be interesting to explore applications of our connection between general learning problems and general optimal control problems (i.e., the connection in Section~\ref{sec:general_connection_Hopf}). 
In general, these optimal control problems can be tackled by solving characteristic line equations, which are represented by two-point boundary value problems (TPBVPs) and thus are more challenging to solve than initial value problems like the Riccati equation. However, working with these TPBVPs may be bypassed in some cases by considering numerical solvers such as those in~\cite{chen2021hopf,chen2021lax,darbon2015convex,darbon2019decomposition,Darbon2016Algorithms,yegorov2017perspectives}.}

Alternatively, if we relax our assumption on convexity by allowing for nonconvex loss functions (or equivalently, nonconvex Hamiltonians), we could also extend our connection to be between learning problems and differential games \cite{evans1984differentialgames} instead of optimal control. However, even with convex losses, there are already many interesting potential applications in machine learning to pursue. 
\updateeditor{For example, in the regularized linear regression case, it may be of interest to explore adjusting methods such as recursive least squares to perform hyperparameter estimation, e.g., similarly to~\cite{chung2020sampled}.} 
More generally, the theoretical connection presented in Section~\ref{sec:general_connection_Hopf} for general convex losses provides a formula for the optimal learning parameters $\weightvec^*$ in terms of the hyper-parameters. This connection could be leveraged to simplify the bi-level optimization in meta-learning applications to a single-level, constrained optimization problem. 
Finally, we could generalize our theory to instead consider viscous HJ PDEs, which add a Laplacian term to the right-hand side of the HJ PDEs~\eqref{eqt:singletimeHJPDE} and~\eqref{eqt:multitimeHJPDE}. Then, by leveraging the recently established connection between viscous HJ PDEs and Bayesian modeling \cite{langlois2021HJvariational}, we could extend to applications in Bayesian inference.

%% file: appendix_short.tex
\def\LQRu{\ba_u}
\def\LQRx{\ba_x}
\def\LQRTx{\bb}
\def\Cx{\bv_x}
\def\Cp{\bv_p}
\newpage
\section{Glossary of related terms in each problem}\label{appendix:glossary}
\begin{table}[h!]
    \centering
    \begin{adjustbox}{width=\textwidth}
    \begin{tabular}{c|c|c|c}
         \textbf{Learning Problem} & \textbf{Multi-Time HJ PDE} & \textbf{Optimal Control} & \textbf{Relation} \\
         \hline \hline
         \begin{tabular}{c}
              minimal loss function \\ $\min_{\weightvec\in\weightspace}\mathcal{L}(\weightvec)$
         \end{tabular} & \begin{tabular}{c}
              solution \\ $S(\HJx, \HJt_1, \dots, \HJt_N)$
         \end{tabular} & \begin{tabular}{c}
              value function \\ $S(\HJx, \HJt_1, \dots, \HJt_N)$
         \end{tabular} & \begin{tabular}{c}
              $\min_{\weightvec\in\weightspace}\mathcal{L}(\weightvec) =-S(\HJx, \HJt_1, \dots, \HJt_N)$ \update{$+c(\HJx)$}
         \end{tabular} \\
         \hline
         \begin{tabular}{c}
              optimal parameters $\weightvec^*$
         \end{tabular} & \begin{tabular}{c}
              spatial gradient \\ $\nabla_\HJx S(\HJx, \HJt_1, \dots, \HJt_N)$; \\
              maximizer $\HJmom^*$ \\ in the Hopf formula
         \end{tabular} & \begin{tabular}{c}
              optimal control $\HJu^*(\cdot)$
         \end{tabular} & \begin{tabular}{c}
              $\weightvec^* = \HJmom^* = \nabla_\HJx S(\HJx, \HJt_1, \dots, \HJt_N)$; \\
              $\HJu^*(s) \in \argmax_{\HJu\in\R^m} \{\langle -f_i(\HJu), \HJmom^*\rangle - L_i(\HJu)\},$ \\ $ s \in \left(\sum_{j=1}^{i-1}\HJt_j, \sum_{j=1}^{i}\HJt_j\right]$
         \end{tabular} \\
         \hline
         \begin{tabular}{c}
              data fitting weights $\param_i$
         \end{tabular} & \begin{tabular}{c}
              time $\HJt_i$ 
         \end{tabular} & \begin{tabular}{c}
              piecewise time horizons $T_j$ 
         \end{tabular} & \begin{tabular}{c}
              $\param_i = \HJt_i, i = 1, \dots, \numt$; \\ $T_j = \sum_{i=1}^j \HJt_i, j = 0, \dots, \numt$
         \end{tabular}\\
         \hline
         \begin{tabular}{c}
              data fitting loss \\ $\lossfunc_i(\mathcal{A}F(\bz_i;\cdot), \by_i)$
         \end{tabular} & \begin{tabular}{c}
              Hamiltonian $\Hamiltonian_i(\cdot)$ 
         \end{tabular} & \begin{tabular}{c}
              running cost $L_i(\HJu)$, \\ dynamics $f_i(\HJu)$ 
         \end{tabular} & \begin{tabular}{c}
              $\lossfunc_i(\mathcal{A}F(\bz_i;\cdot), \by_i) = \Hamiltonian_i(\cdot) =$ \\ $\sup_{\HJu\in\R^m} \{\langle -f_i(\HJu), \cdot\rangle - L_i(\HJu)\}$
         \end{tabular}\\
         \hline
         \begin{tabular}{c}
              regularization $\regfunc(\cdot)$
         \end{tabular} & \begin{tabular}{c}
              initial data $\HJIC(\cdot)$, \\ spatial variable $\HJx$
         \end{tabular} & \begin{tabular}{c}
              terminal cost $\HJIC(\cdot)$, \\ initial position $\HJx$
         \end{tabular} & \begin{tabular}{c}
              $\regfunc(\HJmom) = \HJIC^*(\HJmom) - \langle \HJx,\HJmom\rangle + c(\HJx)$, \\ where $c(\HJx)$ is a constant (possibly 0) \\ that may depend on $\HJx$
         \end{tabular}\\
         \hline
         \begin{tabular}{c}
              \# of learnable \\ parameters $n$
         \end{tabular} & \begin{tabular}{c}
              spatial dimension $n$
         \end{tabular} & \begin{tabular}{c}
              dimension of \\ state space $n$
         \end{tabular} & \begin{tabular}{c}
              same number $n$
         \end{tabular} \\
         \hline
         \begin{tabular}{c}
              \# of data points $N$
         \end{tabular} & \begin{tabular}{c}
              \# of times $N$
         \end{tabular} & \begin{tabular}{c}
              \# of time intervals $N$
         \end{tabular} & \begin{tabular}{c}
              same number $N$
         \end{tabular} \\
         \hline
         \begin{tabular}{c}
              dimension of measured data $m$ \\ (e.g., $\mathcal{A}F(\cdot;\weightvec), \by_i\in \R^m$)
         \end{tabular} & \begin{tabular}{c}
              Hamiltonian $H_i(\cdot)$
         \end{tabular} & \begin{tabular}{c}
              dimension of \\ control space $m$
         \end{tabular} & \begin{tabular}{c}
              $\Hamiltonian_i(\cdot) =\sup_{\HJu\in\R^m} \{\langle -f_i(\HJu), \cdot\rangle - L_i(\HJu)\}$
         \end{tabular}
    \end{tabular}
    \end{adjustbox}
    \caption{\updateone{Glossary of the mathematical relations between the quantities in the regularized learning problem~\eqref{eqt:general_learning}, the multi-time HJ PDE~\eqref{eqt:multitimeHJPDE}, and the associated optimal control problem~\eqref{eqt:optimal_control_standardform}. These mathematical relations allow us to form our theoretical connection in Section~\ref{sec:general_connection_Hopf}. Note that the variables $\HJx, \HJt$ in the HJ PDE become hyper-parameters in the learning problem, and we can treat them as constants when optimizing the learning problem~\eqref{eqt:general_learning} with respect to $\weightvec = \HJmom$.}}
    \label{tab:glossary}
\end{table}

\section{A more general version of LQR}\label{appendix:general_LQR}
In Section~\ref{sec:intro_LQR}, we presented a concise overview of the LQR problem in its canonical form. However, this paper establishes a linkage between regression problems and LQR problems with lower-order terms. Thus, in this section, we discuss an LQR problem with lower-order terms that more closely aligns with those used to establish our connections. We note that the notation employed in this section differs slightly from the notation utilized in the main body of the paper. The finite-horizon, continuous-time LQR with lower-order terms can be expressed as follows:
\begin{multline}\label{eqt:LQR_general_lower_order}
    S(\HJx,\HJt) = \min_{\HJu(\cdot)} \Bigg\{\int_0^\HJt \Big(\frac{1}{2}\bx(s)^T\LQRxx\bx(s) + \frac{1}{2}\HJu(s)^T\LQRuu\HJu(s) + \bx(s)^T\LQRxu\HJu(s) \\
    - \LQRu^T\bu(s) - \LQRx^T\bx(s) \Big) ds  + \frac{1}{2}\bx(\HJt)^T\LQRTC\bx(\HJt) + \LQRTx^T \bx(\HJt): \\
    \dot\bx(s) = \LQRA\bx(s) + \LQRB\HJu(s) \forall s\in(0,t], \bx(0) = \HJx\Bigg\},
\end{multline}
where $\LQRxx, \LQRTC \in\R^{n\times n}$ and $\LQRuu\in\R^{m\times m}$ are symmetric positive definite, $\LQRxu\in\R^{n\times m}$, $\LQRA\in\R^{n\times n}$, $\LQRB\in\R^{n\times m}$, $\LQRx,\LQRTx\in \Rn$, and $\LQRu\in\R^m$. The corresponding HJ PDE is 
\begin{equation}
    \begin{dcases}
    \frac{\partial S(\HJx,\HJt)}{\partial \HJt} + \Hamiltonian(\HJx,\nabla_\HJx S(\HJx,\HJt)) = 0 & \HJx\in\R^n, \HJt > 0, \\
    S(\HJx,0) = \HJIC(\HJx)  & \HJx\in\R^n,
    \end{dcases}
\end{equation}
where the initial data of the HJ PDE is given by the terminal cost $\HJIC(\HJx) := \frac{1}{2}\HJx^T\LQRTC\HJx + \LQRTx^T \bx$ of the optimal control problem~\eqref{eqt:LQR_general_lower_order} and the Hamiltonian $\Hamiltonian$ is defined by
\begin{equation}
    \begin{aligned}
    \Hamiltonian&(\HJx,\HJmom)  = \sup_{\HJu\in\R^m} \langle -f(\HJx,\HJu), \HJmom \rangle - L(\HJx,\HJu) \\
     &=  -\langle \LQRA\HJx, \HJmom\rangle - \frac{1}{2}\langle \HJx, \LQRxx\HJx\rangle + \LQRx^T \bx+ \frac{1}{2}\langle \LQRB^T\HJmom + \LQRxu^T\HJx - \LQRu, \LQRuu^{-1}(\LQRB^T\HJmom + \LQRxu^T\HJx  - \LQRu) \rangle,
    \end{aligned}
\end{equation}
where $f(\HJx,\HJu) = \LQRA\HJx + \LQRB\HJu$ is the source term of the dynamics and $L(\HJx,\HJu) = \frac{1}{2}\HJx^T\LQRxx\HJx + \frac{1}{2}\HJu^T\LQRuu\HJu + \HJx^T\LQRxu\HJu - \LQRu^T \bu - \LQRx^T \bx$ is the running cost.

It is well-known that this LQR problem can be solved using the Riccati equation as follows. Define $\Cpp = \LQRB \LQRuu^{-1}\LQRB^T$, $\Cxx = -\LQRxu\LQRuu^{-1}\LQRxu^T + \LQRxx$, $\Cxp = \LQRA - \LQRB \LQRuu^{-1}\LQRxu^T$, $\Cx = \LQRx - \LQRxu \LQRuu^{-1}\LQRu$, and $\Cp = \LQRB\LQRuu^{-1} \LQRu$. 
Then, the solution is given by
$S(\HJx,\HJt) = \frac{1}{2} \HJx^T \Sxx(\HJt)\HJx + \Sx(\HJt)^T\HJx + \Sc(\HJt)$, 
where the function $\Sxx: [0,\infty)\to\R^{n\times n}$ which takes values in the space of symmetric positive definite matrices, $\Sx\colon [0,+\infty)\to\Rn$, and $\Sc\colon[0,+\infty)\to\R$ solve the following Riccati ODE system:
\begin{equation} 
{\small
    \begin{dcases}
    \dot{\Sxx}(\HJt) =  -\Sxx(\HJt)^T\Cpp\Sxx(\HJt) + \Sxx(\HJt)^T\Cxp + \Cxp^T\Sxx(\HJt) + \Cxx &\HJt\in(0,+\infty),\\
    \dot{\Sx}(\HJt) = -\Sxx(\HJt)^T \Cpp \Sx(\HJt) + \Cxp^T \Sx(\HJt) - \Cx + \Sxx(\HJt)^T \Cp &\HJt\in(0,+\infty),\\
    \dot{\Sc}(\HJt) = -\frac{1}{2}\|\LQRuu^{-1/2}(\LQRB^T\Sx(\HJt) - \LQRu)\|^2 &\HJt\in(0,+\infty),\\
    \Sxx(0) = \LQRTC, \Sx(0) = \LQRTx, \Sc(0) = 0.
    \end{dcases}
    }
\end{equation}

\section{Details of the methodology}
\subsection{Algorithm for deleting one data point} \label{appendix:method_delete_data}
Here, we provide details for the algorithm for deleting one data point from Section~\ref{subsec:method_2}.
Removing the $j$-th data point corresponds to removing the term $\frac{1}{2}\lambda_{j}\|\Phi_{j} \weightvec - \MLy_{j}\|_2^2 $ in the loss function~\eqref{eq:loss_function} or, equivalently, removing the Hamiltonian $\frac{1}{2}\|\Phi_{j} \weightvec - \MLy_{j}\|_2^2$ from the multi-time HJ PDE and removing the pieces $L_{j}(s, \HJu) = \frac{1}{2}\HJu^T\HJu - \MLy_{j}$ and $f(s,\HJu) = \Phi_{j}^T\HJu$ from the running cost and dynamics, respectively, of the corresponding piecewise LQR problem.
Hence, numerically, we can remove the $j$-th data point by solving the following Riccati ODE
\begin{equation}\label{eqt:regression_2Riccati}
    \begin{dcases}
    \dot{\tilde\Sxx}(\HJt) =  -\tilde\Sxx(\HJt)^T\MLbasismat_{j}^T\MLbasismat_{j}\tilde\Sxx(\HJt) &\HJt<\lambda_j,\\
    \dot{\tilde\Sx}(\HJt) = -\tilde\Sxx(\HJt)^T\MLbasismat_{j}^T(\MLbasismat_{j}\tilde\Sx(\HJt) - \MLy_{j})&\HJt<\lambda_j
    \end{dcases}
\end{equation}
with terminal condition $\tilde\Sxx(\lambda_j) = \Sxx\left(T_N\right)$ and $\tilde\Sx(\lambda_j) = \Sx\left(T_N\right)$, where $\Sxx\left(T_N\right)$ and $\Sx\left(T_N\right)$ are obtained from solving the learning problem~\eqref{eq:loss_function} with all $N$ data points. Then, the solution to the new learning problem with the $j$-th point removed is given by~\eqref{eqt:sec42_newoptimizer}, where $\tilde\Sxx = \tilde\Sxx(0)$ and $\tilde \Sx = \tilde \Sx(0)$ are the solution to~\eqref{eqt:regression_2Riccati}.

\subsection{Algorithm for tuning the regularization weights}\label{appendix:hyperparam_tuning}
Here, we provide details for the algorithm for \updatetwo{tuning the regularization weights} from Section~\ref{subsec:method_3}.
We consider the case where we change each regularization parameter $\MLreg_k$ to $\tilde\MLreg_k$. This change can be regarded as two steps: first, we change all parameters $\MLreg_k$ for the indices $k$ such that $\tilde\MLreg_k > \MLreg_k$, and then we change the other parameters. Define the index set $\mathcal{K} $ to be $\mathcal{K} = \{k\colon \tilde\MLreg_k > \MLreg_k\}$. 

The first step is equivalent to adding the term $\sum_{k\in \mathcal{K}}\frac{\tilde\MLreg_k - \MLreg_k}{2}(\weight_k - \MLcenter_k)^2$ to the loss function~\eqref{eq:loss_function}. We can interpret this as adding an $(N+1)$-th Hamiltonian $\frac{1}{2}\weightvec^T\MLregmat_{+}\weightvec$ with corresponding time variable $\HJt_{N+1} = 1$ to the multi-time HJ PDE, where $\MLregmat_+$ is a diagonal matrix whose $k$-th diagonal element is $\tilde\MLreg_k - \MLreg_k$ if $k\in\mathcal{K}$ and $0$ otherwise.
Therefore, the solution to this new multi-time HJ PDE can be solved by the following Riccati equation:
\begin{equation}\label{eqt:riccati_increase_regweight}
\begin{dcases}
\dot{\Sxx}_+(\HJt) =  -\Sxx_+(\HJt)^T\MLregmat_+\Sxx_+(\HJt) &\HJt\in \left(0,1\right),\\
\dot{\Sx}_+(\HJt) = -\Sxx_+(\HJt)^T\MLregmat_+\Sx_+(\HJt)&\HJt\in \left(0, 1\right),
\end{dcases}
\end{equation}
with initial condition $\Sxx_+(0)$ and $\Sx_+(0)$, which are the corresponding solutions to the Riccati equations before changing the weights $\MLreg_k, k\in\mathcal{K}$. In other words, we set $\Sxx_+(0) = \Sxx\left(T_N\right)$ and $\Sx_+(0) = \Sx\left(T_N\right)$, where $\Sxx(T_N), \Sx(T_N)$ are obtained from solving the original learning problem~\eqref{eq:loss_function} with the original values of $\MLreg_k$.

The second step is equivalent to removing the term $\sum_{k\not\in\mathcal{K}}\frac{\MLreg_k - \tilde\MLreg_k}{2}(\weight_k - \MLcenter_k)^2$ from the loss function~\eqref{eq:loss_function}. This is equivalent to solving a single-time HJ PDE with a terminal condition at time $1$ and Hamiltonian $\frac{1}{2}\weightvec^T\MLregmat_{-}\weightvec$, where $\MLregmat_-$ is a diagonal matrix whose $k$-th diagonal element is $\MLreg_k - \tilde\MLreg_k$ if $k\not\in\mathcal{K}$ and $0$ otherwise.
Then, the solution can be obtained by solving the following Riccati equation:
\begin{equation}\label{eqt:riccati_decrease_reg_weight}
\begin{dcases}
\dot{\Sxx}_-(\HJt) =  -\Sxx_-(\HJt)^T\MLregmat_-\Sxx_-(\HJt) &\HJt\in \left(0,1\right),\\
\dot{\Sx}_-(\HJt) = -\Sxx_-(\HJt)^T\MLregmat_-\Sx_-(\HJt)&\HJt\in \left(0, 1\right)
\end{dcases}
\end{equation}
with terminal condition $\Sxx_-(1) = \Sxx_+(1)$ and $\Sx_-(1) = \Sx_+(1)$, where $\Sxx_+, \Sx_+$ are obtained from the solution to~\eqref{eqt:riccati_increase_regweight}.

Finally, the minimizer of the new learning problem after changing all of the weights $\MLreg_k$ in the regularization term is given by ${\Sxx}_-(0) \tilde\MLregmat \MLcentervec+ {\Sx}_-(0)$, where ${\Sxx}_-, {\Sx}_-$ are obtained from the solution to~\eqref{eqt:riccati_decrease_reg_weight} and $\tilde\MLregmat$ is a diagonal matrix whose $k$-th diagonal element is the new regularization parameter $\tilde\MLreg_k$.

\section{Computational complexity of the methodology}\label{appendix:comp_complexity}
\updatethree{In this section, we compare the computational complexity of our Riccati-based methodology (using RK4) from Section~\ref{sec:method} to that of the method of least squares.}

\subsection{Initial training}\label{sec:computational_complexity_initial_training}
\updatethree{First, we consider solving the learning problem~\eqref{eq:loss_function} with $N$ points, which may be considered as an initial training step with $N$ data points.}

\updatethree{In this case, our Riccati-based methodology requires us to solve a sequence of $N$ Riccati ODEs of the form~\eqref{eqt:sequentialRiccatiODEs}. Using RK4 to solve each ODE involves computing $\Phi_i^T\Phi_i$ (recall that $\Phi_i\in\R^{m\times n}$), which requires $O(mn^2)$ operations, and multiplying several $n\times n$ matrices together, which requires $O(n^3)$ operations. In other words, solving one step of Riccati ODEs (i.e., adding one data point to our training set) requires $O(mn^2 + n^3)$ operations. Thus, the overall computational complexity of our Riccati-based methodology for training on $N$ points is $O(Nmn^2 + Nn^3)$.}

\updatethree{In contrast, using the method of least squares to solve~\eqref{eq:loss_function} with $N$ points involves computing $\Phi_i^T\Phi_i$ for each point, which requires $O(mn^2)$ operations for each point or $O(Nmn^2)$ operations for all $N$ points, and then solving an $n\times n$ linear system, which requires $O(n^3)$ operations. Thus, the overall computational complexity of the method of least squares for training on $N$ points is $O(Nmn^2 + n^3)$.}

\updatethree{As expected, the method of least squares beats our Riccati-based methodology in this case, but as we discussed previously, our methodology is advantageous when changing the training set or hyper-parameters.}

\subsection{Continual learning} 
\updatethree{In this section, we discuss adding data points to the training set (as in Section~\ref{subsec:method_2}), where we update our learned models as soon as a new data point becomes available (e.g., as in continual learning). We consider the initial training set to have $N$ points and then sequentially incorporate $K$ more data points.}

\updatethree{As discussed in Section~\ref{sec:computational_complexity_initial_training}, adding one data point using our Riccati-based methodology requires $O(mn^2 + n^3)$ operations. Thus, the overall computational complexity of using our methodology to perform the initial training and then to retrain $K$ times is $O((N+K)mn^2 + (N+K)n^3)$. }
\updatethree{Adding the $i$-th data point using the method of least squares requires $O((N+i)mn^2 + n^3)$ operations. Thus, the overall computational complexity of using the method of least squares to perform the initial training and then to retrain $K$ times is $\sum_{i=0}^{K} O((N+i)mn^2 + n^3) = O((KN + K^2)mn^2 + Kn^3)$. }
\updatethree{Thus, our Riccati-based methodology becomes increasingly more advantageous over the method of least squares as the number of points added $K$ increases.}

\subsection{Tuning the bias}
\updatethree{As discussed in Section~\ref{subsec:method_3}, tuning the bias $\MLcentervec$ using our Riccati-based methodology is accomplished via the update~\eqref{eqt:minimizer_changeofbias}. This update only involves a matrix-vector multiplication (recall that $\Gamma$ is diagonal) and vector addition. Thus, tuning the bias using our methodology requires $O(n^2)$ operations. In contrast, tuning the bias using the method of least squares requires retraining on the entire dataset. Thus, if the training set has size $N$, then the computational complexity of the method of least squares in this case is $O(Nmn^2 + n^3)$. Hence, our methodology is much more efficient than the method of least squares in this case. Note that these same calculations also apply to using our methodology vs the method of least squares to perform the first step in each iteration of PDHG~\eqref{eq:pdhg}, which, as discussed previously, is equivalent to performing a change of bias.}

\section{Riccati ODEs and Recursive Least Squares}\label{sec:RLS}
In Section~\ref{subsec:method_2}, we discuss adding or removing data using our Riccati-based methodology. However, data points could also be added or removed incrementally using recursive least squares. In this section, we discuss the connections between our Riccati-based methodology and recursive least squares. Namely, we show that the recursive least squares method for this problem can be considered as a special case of our Riccati-based methodology. Note that in this section, we focus on adding data points, but the results for removing data points can be derived similarly. For generality, we consider the more general learning problem~\eqref{eqt:regression_multidata} in Section~\ref{sec:LQR_multiptregression}, and the results for the more specific learning problem in Section~\ref{subsec:method_2} can be recovered using the identifications listed in Section~\ref{subsec:method_1}.

The analytical solution to the Riccati ODEs~\eqref{eqt: odeP_piecewise},~\eqref{eqt: odeq_piecewise} is given by
\begin{equation*}
\Sxx_N = \left(\LQRTC^{-1} + \sum_{i=1}^N t_i\LQRB_i\LQRuu_i^{-1}\LQRB_i^T\right)^{-1}, 
\end{equation*}
\begin{equation*}
    \Sx_N = \left(\LQRTC^{-1} + \sum_{i=1}^N t_i\LQRB_i\LQRuu_i^{-1}\LQRB_i^T\right)^{-1}\left(\LQRTC^{-1}\bb + \sum_{i=1}^N t_i\LQRB_i\LQRuu_i^{-1} \ba_i\right),
\end{equation*}
where $\Sxx_N = \Sxx\left(\sum_{i=1}^N t_i\right)$ and $\Sx_N = \Sx\left(\sum_{i=1}^N t_i\right)$. 
Then, using the Woodbury matrix identity, the recursive least squares solution is given by
\begin{equation}\label{eq:RLS}
\begin{aligned}
    \Sxx_N &= \Sxx_{N-1} - t_N\Sxx_{N-1}\LQRB_N\left(\LQRuu_N + t_N\LQRB_N^T \Sxx_{N-1} \LQRB_N\right)^{-1}\LQRB_N^T\Sxx_{N-1}, \\
    \Sx_N &= \Sx_{N-1} + t_N\Sxx_N\LQRB_N\LQRuu_N^{-1}\ba_N - t_N\Sxx_{N-1}\LQRB_N\left(\LQRuu_N + t_N\LQRB_N^T \Sxx_{N-1} \LQRB_N\right)^{-1}\LQRB_N^T\Sx_{N-1}.
\end{aligned}
\end{equation}

\begin{prop}
    The recursive least squares solution~\eqref{eq:RLS} is a discretization of the Riccati ODEs~\eqref{eqt: odeP_piecewise},~\eqref{eqt: odeq_piecewise}.
\end{prop}

\begin{proof}
    We begin by proving the result for $\Sxx$. Let $\Sxx(\cdot)$ be the solution to the Riccati ODEs. Then $\Sxx(s)$ is invertible for all $s\in \left[0, \sum_{i=1}^N t_i\right]$, which gives us that 
    $\frac{d\Sxx^{-1}}{ds} = -\Sxx^{-1}\frac{d\Sxx}{ds}\Sxx^{-1}.$
    Plugging this identity into~\eqref{eqt: odeP_piecewise}, we have that
    \begin{equation}\label{eq:RLS_dPinv}
        \frac{d\Sxx^{-1}(s)}{ds} = \LQRB_i\LQRuu_i^{-1}\LQRB_i^T \quad s\in[T_{i-1}, T_i), \quad \Sxx^{-1}(0) = \LQRTC^{-1}.
    \end{equation}
    The solution to this ODE is given by 
    $$\Sxx^{-1}(s) = \LQRTC^{-1} + \sum_{j = 1}^i t_j\LQRB_j\LQRuu_j^{-1}\LQRB_j^T + (s - T_{i-1})\LQRB_i\LQRuu_i^{-1}\LQRB_i^T, \forall s\in[T_{i-1}, T_i), i = 0, \dots, N.$$
    If we only consider the values of $\Sxx^{-1}(s)$ at the discrete times $s = T_0, T_1, \dots, T_N$, then this solution can be defined recursively as 
    \begin{equation*}
        \Sxx^{-1}(T_i) = \Sxx^{-1}(T_{i-1}) + t_i\LQRB_i\LQRuu_i^{-1}\LQRB_i^T,
    \end{equation*}
    or, in other words,  we have that
    \begin{equation*}
    \Sxx(T_i) = (\Sxx^{-1}(T_{i-1}) + t_i\LQRB_i\LQRuu_i^{-1}\LQRB_i^T)^{-1}.
    \end{equation*}
    By applying the Woodbury matrix identity, we recover the recursive least squares solution for $\Sxx$ in~\eqref{eq:RLS}.
    Next, we prove the result for $\Sx$. Consider 
    \begin{equation*}\label{eq:RLS_dPinvq}
        \frac{d(\Sxx^{-1}\Sx)}{ds}(s) = \frac{d\Sxx^{-1}}{ds}\Sx + \Sxx^{-1}\frac{d\Sx}{ds} = \LQRB_i\LQRuu_i^{-1}\LQRB_i^T\Sx + \Sxx^{-1}\frac{d\Sx}{ds} = \LQRB_i\LQRuu_i^{-1}\ba_i
    \end{equation*}
    for $s\in[T_{i-1}, T_i)$ and with initial condition $(\Sxx^{-1}\Sx) (0) = -\LQRTC^{-1}\bb$, where the second equality follows from~\eqref{eq:RLS_dPinv} and the third equality follows from multiplying~\eqref{eqt: odeq_piecewise} by $\Sxx^{-1}$ on the left. Then, the solution to this ODE gives us that
    $$\Sx(s) = \Sxx(s)\Sxx^{-1}(T_{i-1})\Sx(T_{i-1}) + (s-T_{i-1})\Sxx(s)\LQRB_i\LQRuu_i^{-1}\ba_i,  \forall s\in[T_{i-1}, T_i), i = 0, \dots, N,$$
    which defines $\Sx(s)$ recursively. If we only consider the values of $\Sx(s)$ at the discrete times $s = T_0, T_1, \dots, T_N$, then using this recursive definition of $\Sx$ simplifies to
    \begin{equation*}
        \Sx(T_i) = \Sxx(T_i)\left(\Sxx^{-1}(T_{i-1})\Sx(T_{i-1}) + t_i\LQRB_i\LQRuu_i^{-1}\ba_i\right).
    \end{equation*}
    Finally, applying the recursive least squares solution for $\Sxx(T_i)$ in~\eqref{eq:RLS} to the definition of $\Sx(T_i)$ above gives us the recursive least squares solution for $\Sx$ in~\eqref{eq:RLS}.
\end{proof}

Thus, we have shown that the recursive least squares solution represents a special case (in the form of a discretization) of the Riccati ODEs, and hence, we may regard the recursive least squares update~\eqref{eq:RLS} as an alternative method for solving the Riccati ODEs~\eqref{eqt: odeP_piecewise},~\eqref{eqt: odeq_piecewise}, if we only care about the solution at the discrete times $s = T_0, \dots, T_N$. For instance, in Section~\ref{subsec:example_3}, we show how using more general numerical methods like RK4 naturally expose 1D curves along the Pareto front of solutions, which aligns with the Riccati ODEs as a continuous formulation; in contrast, the discretization used by the recursive least squares update would only allow for discrete points on the Pareto front. Moreover, we note that there are some cases in which recursive least squares is known to be numerically unstable~\cite{ljung1985error, liavas1998numerical, liavas1999numerical}, whereas Runge-Kutta methods have well-established stability and accuracy results \cite{butcher2016numerical}.

\section{Additional results for example 2}\label{appendix:1}
\begin{figure}[ht]
    \begin{subfigure}[b]{\textwidth}
        \centering
        \includegraphics[width = 0.3\textwidth]{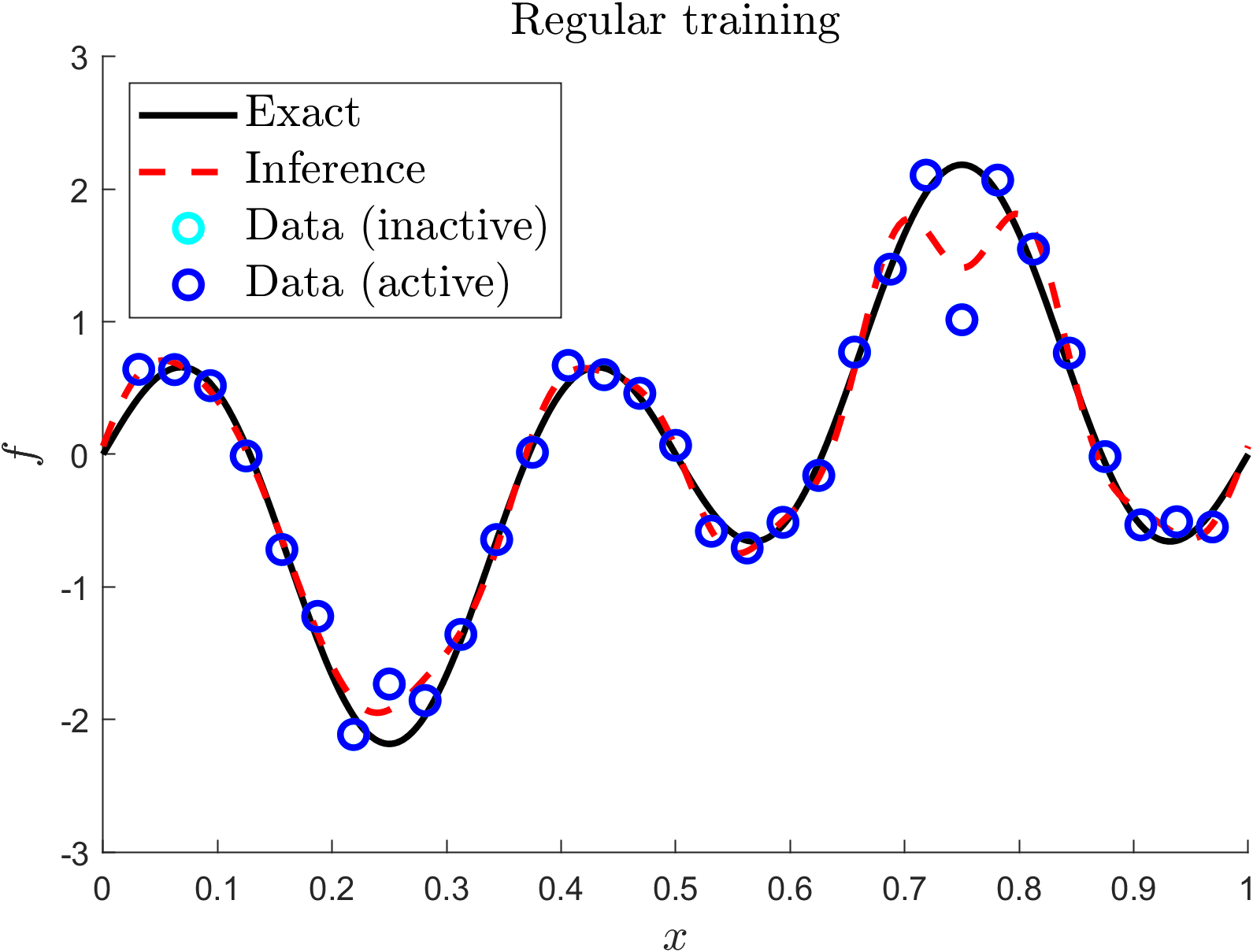}
        \includegraphics[width = 0.3\textwidth]{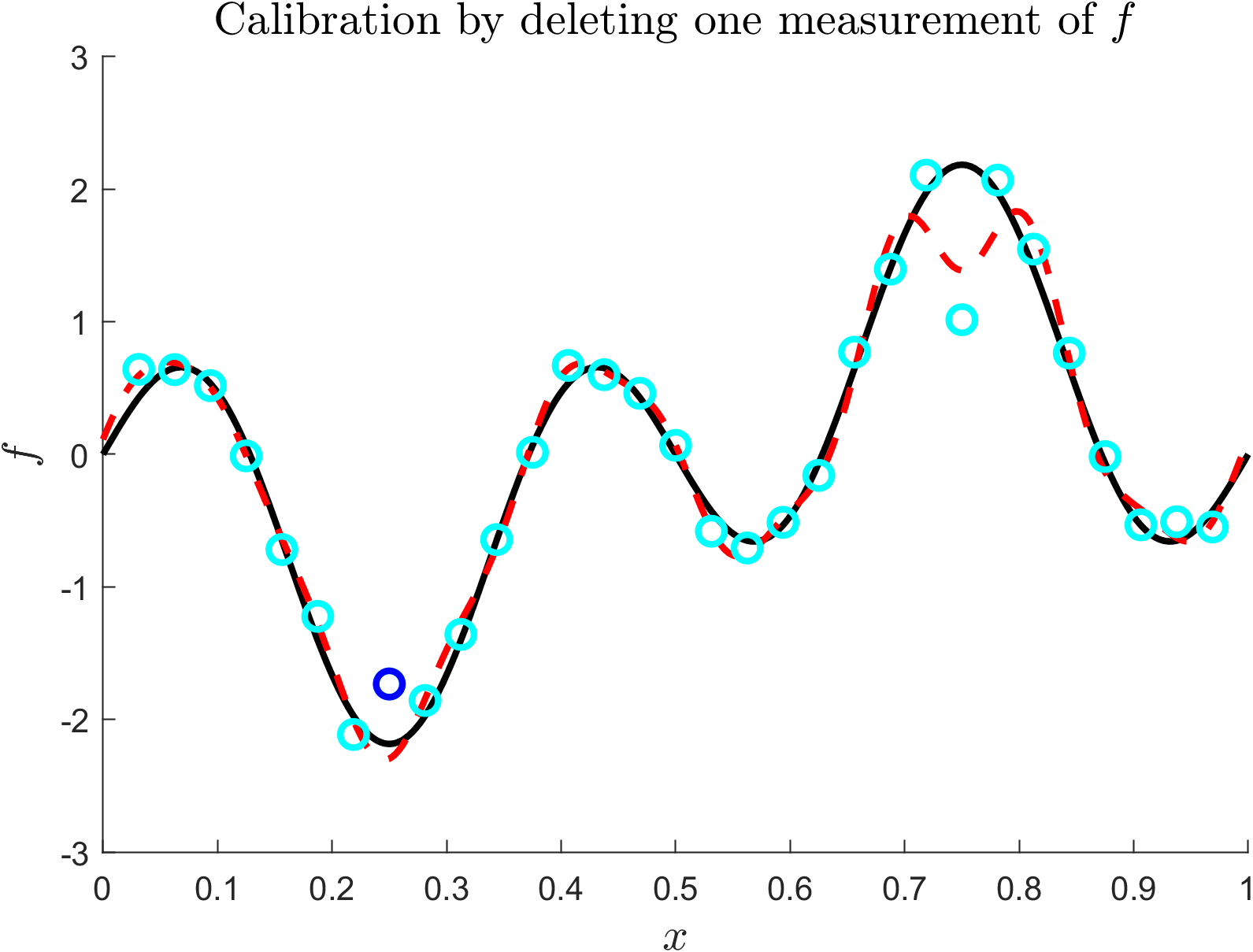}
        \includegraphics[width = 0.3\textwidth]{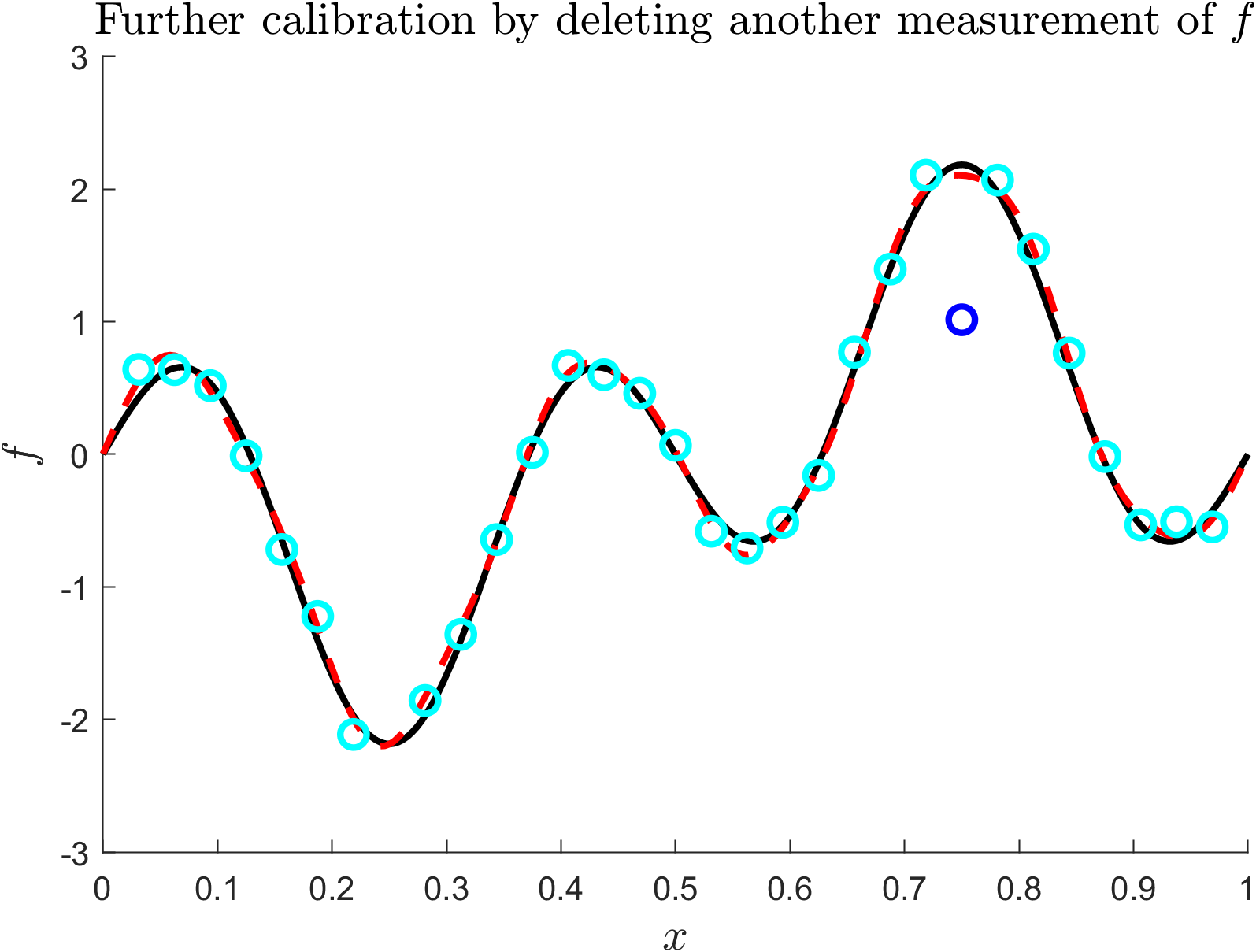}
        \caption{}
    \end{subfigure}
    \begin{subfigure}[b]{\textwidth}
        \centering
        \includegraphics[width = 0.3\textwidth]{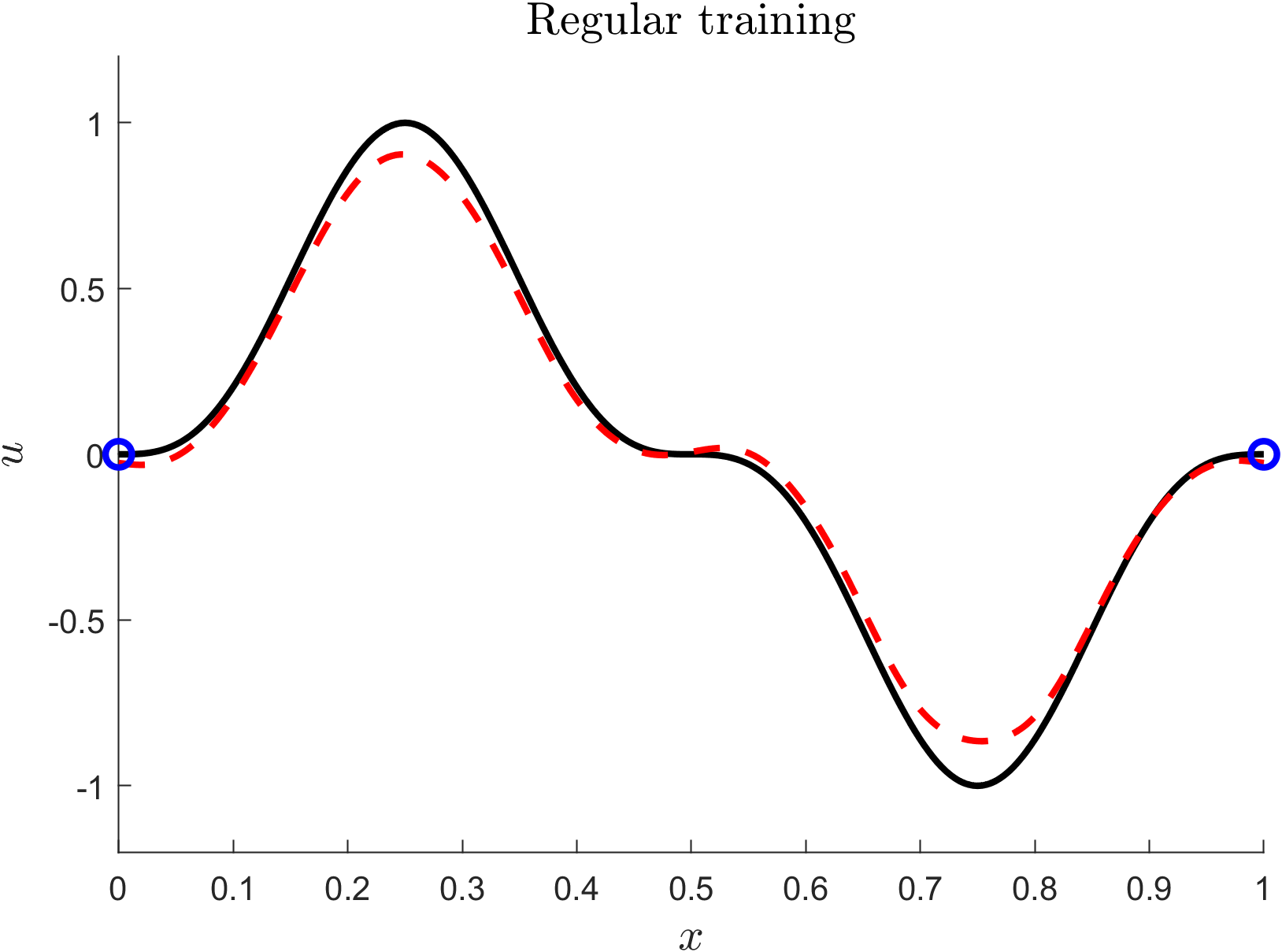}
        \includegraphics[width = 0.3\textwidth]{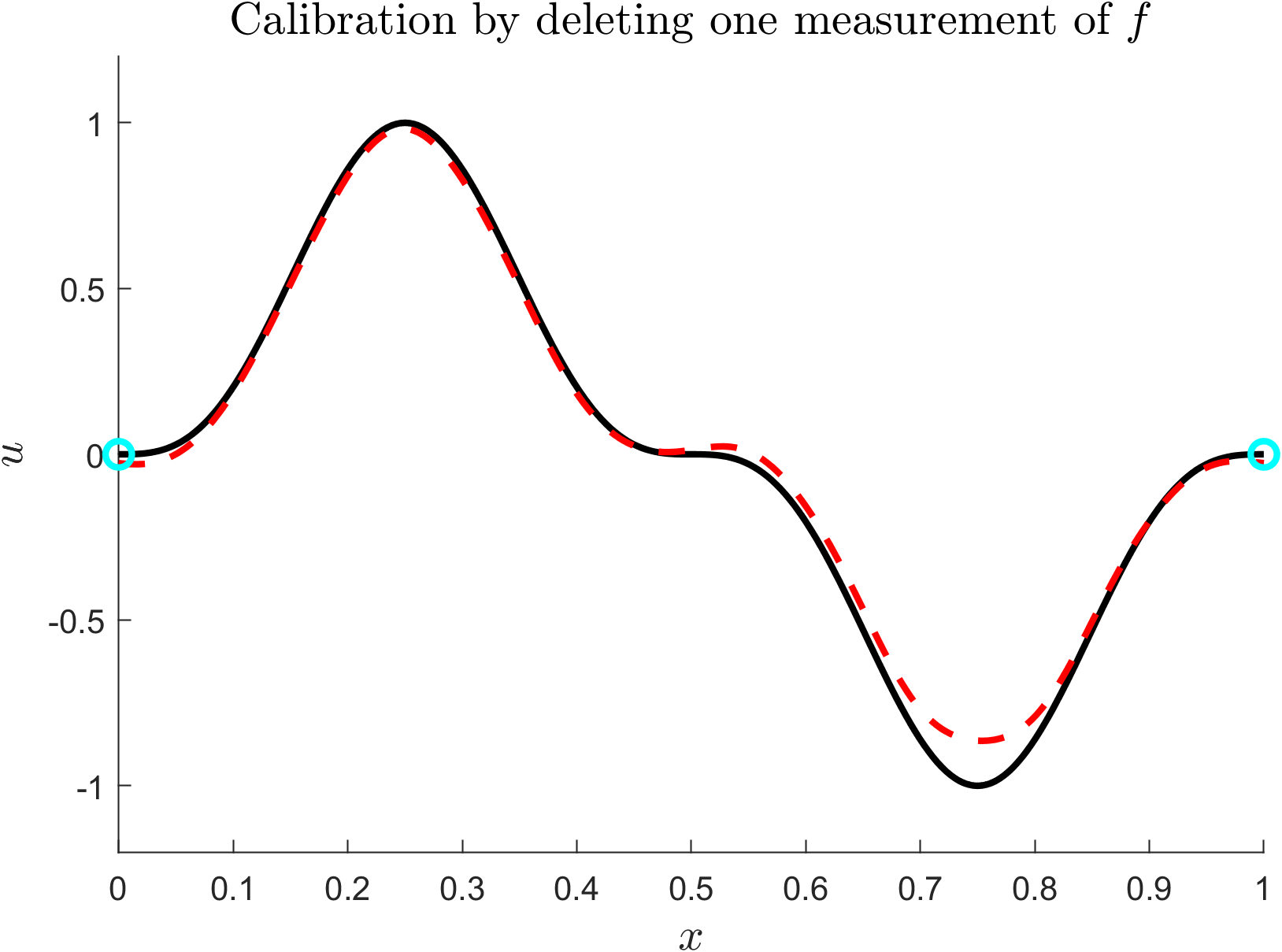}
        \includegraphics[width = 0.3\textwidth]{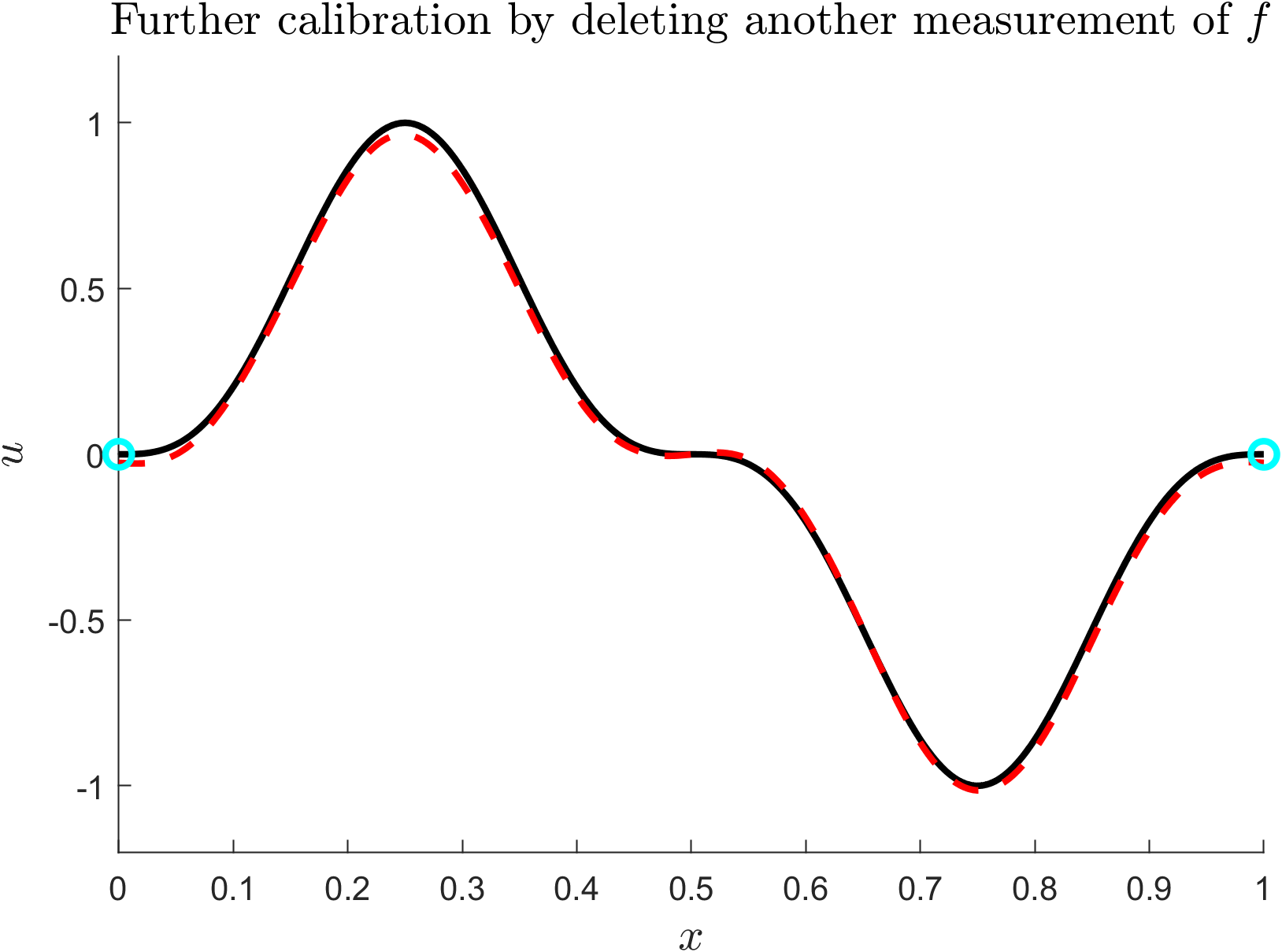}
        \caption{}
    \end{subfigure}
    \caption{Results of solving the 1D steady-state reaction-diffusion equation~\eqref{eq:reaction} with noisy measurements of the source term $f$ in the domain and noiseless measurements of the solution $u$ on the boundary. (a) results for $f$; (b) results for $u$. \textbf{Left}: results of regular training using our Riccati-based method in Section~\ref{subsec:method_1}; \textbf{middle} and \textbf{right}: calibrating the results of regular training by eliminating two outlier measurements of $f$ using the methodology described in Section~\ref{subsec:method_2}. Calibrations are employed without re-training or access to the data from the previous training.}
    \label{fig:calibration:2}
\end{figure}

In Section~\ref{subsec:example_2}, we consider a 1D steady-state reaction-diffusion equation~\eqref{eq:reaction} and discuss three different types of post-training calibrations: adding new data points to compensate for a lack of knowledge in the regular training, enforcing the fitting of some data points by increasing the weights $\lambda_i$ of their respective terms in the loss function~\eqref{eq:calibration:loss}, and removing some data points so that their  effects are eliminated. In this section, we present the results for the last case \updatethree{and then provide a large-scale continual learning example for solving~\eqref{eq:reaction}}. In Figure~\ref{fig:calibration:2}, we remove two outliers one-by-one and observe that their removal does successfully increase the accuracy of the learned model. Again, using our Riccati-based approach, the removal of these points is done using only knowledge about the point to be removed and the results of the previous training step. 

\begin{table}[ht]
    \footnotesize
    \centering
    \begin{tabular}{|c|c|c|c|}
    \hline
         & $N=100$ & $N=1000$ & $N=50000$ \\
       \hline
       Error of $u$ & $72.59\%$ & $22.55\%$ & $1.76\%$\\
       \hline
       Error of $f$ & $89.65\%$ & $29.76\%$ & $3.35\%$\\
       \hline
    \end{tabular}
\caption{Errors of $u$ and $f$ when solving the 1D steady-state reaction-diffusion equation \eqref{eq:reaction} with different numbers $N$ of noisy measurements of $f$ using our Riccati-based approach. The errors represent the relative $L^2$ errors on a uniform grid of $[0, 1]$. New measurements are incorporated following the continual learning framework, and, using our Riccati-based methodology, incorporating new data points does not require retraining on or access to any of the previous data points. Qualitative results can be found in Figure \ref{fig:example_2:3}.}
\label{tab:example_2:2}
\end{table}

\begin{figure}[ht]
    \begin{subfigure}[b]{\textwidth}
        \centering
        \includegraphics[width = 0.3\textwidth]{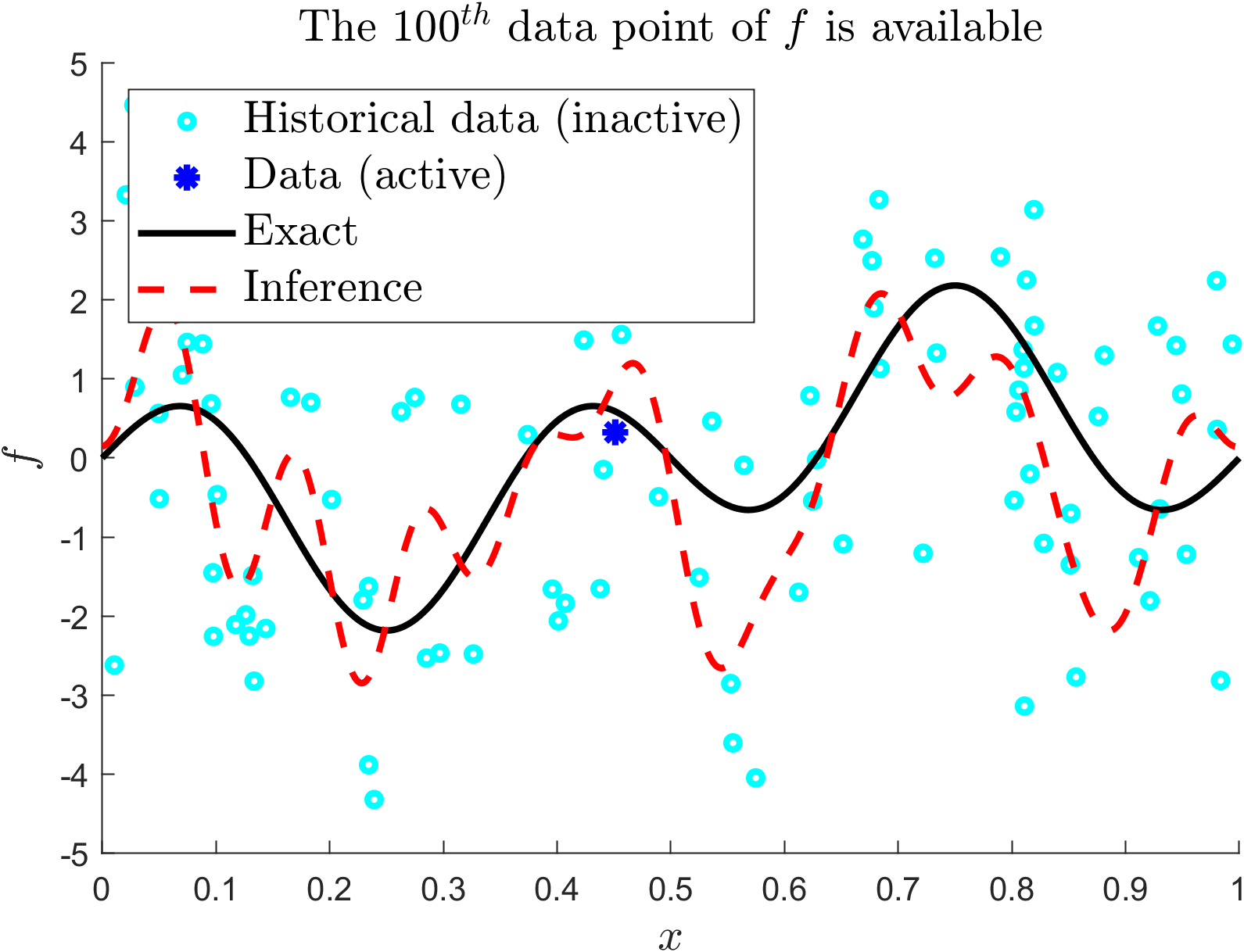}
        \includegraphics[width = 0.3\textwidth]{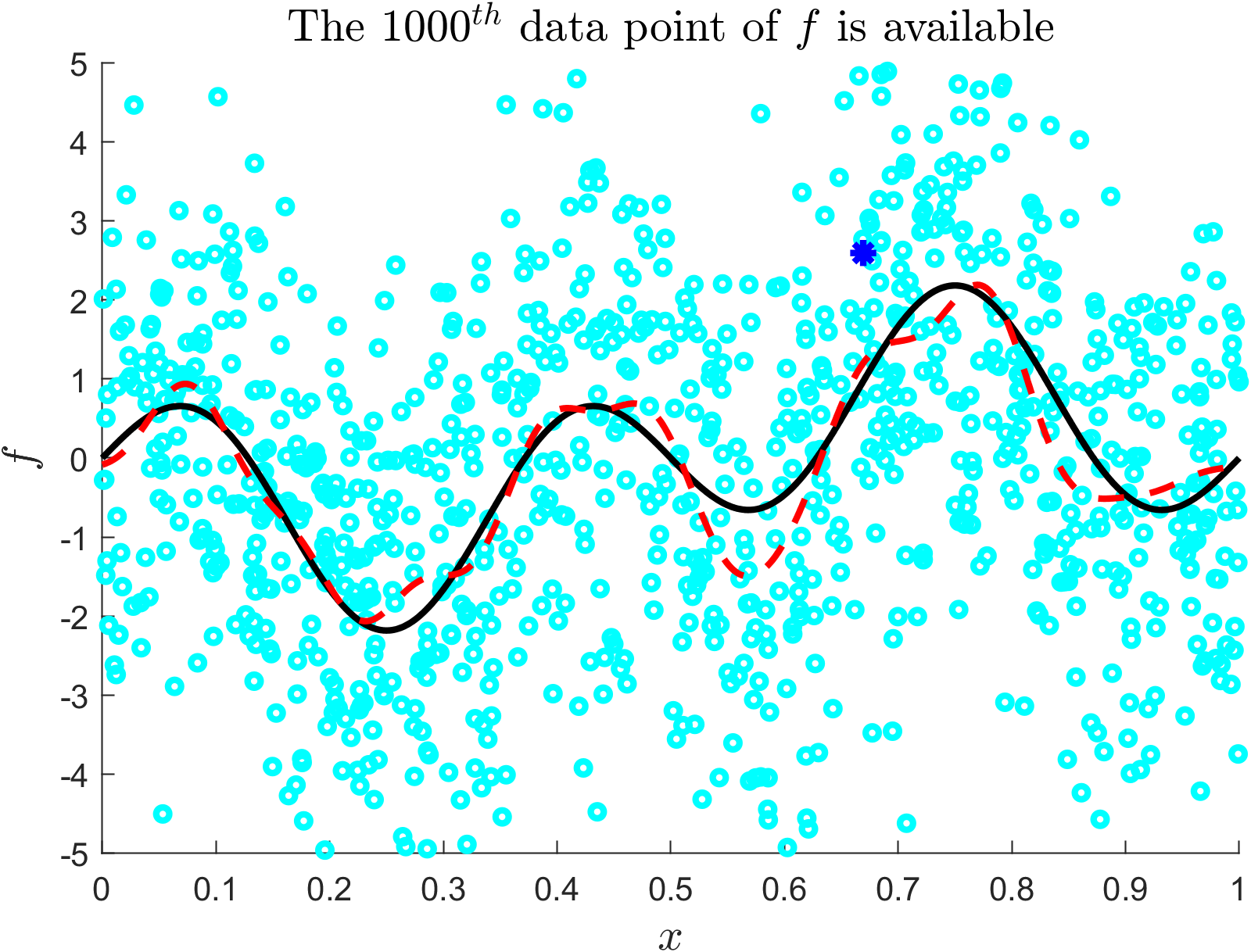}
        \includegraphics[width = 0.3\textwidth]{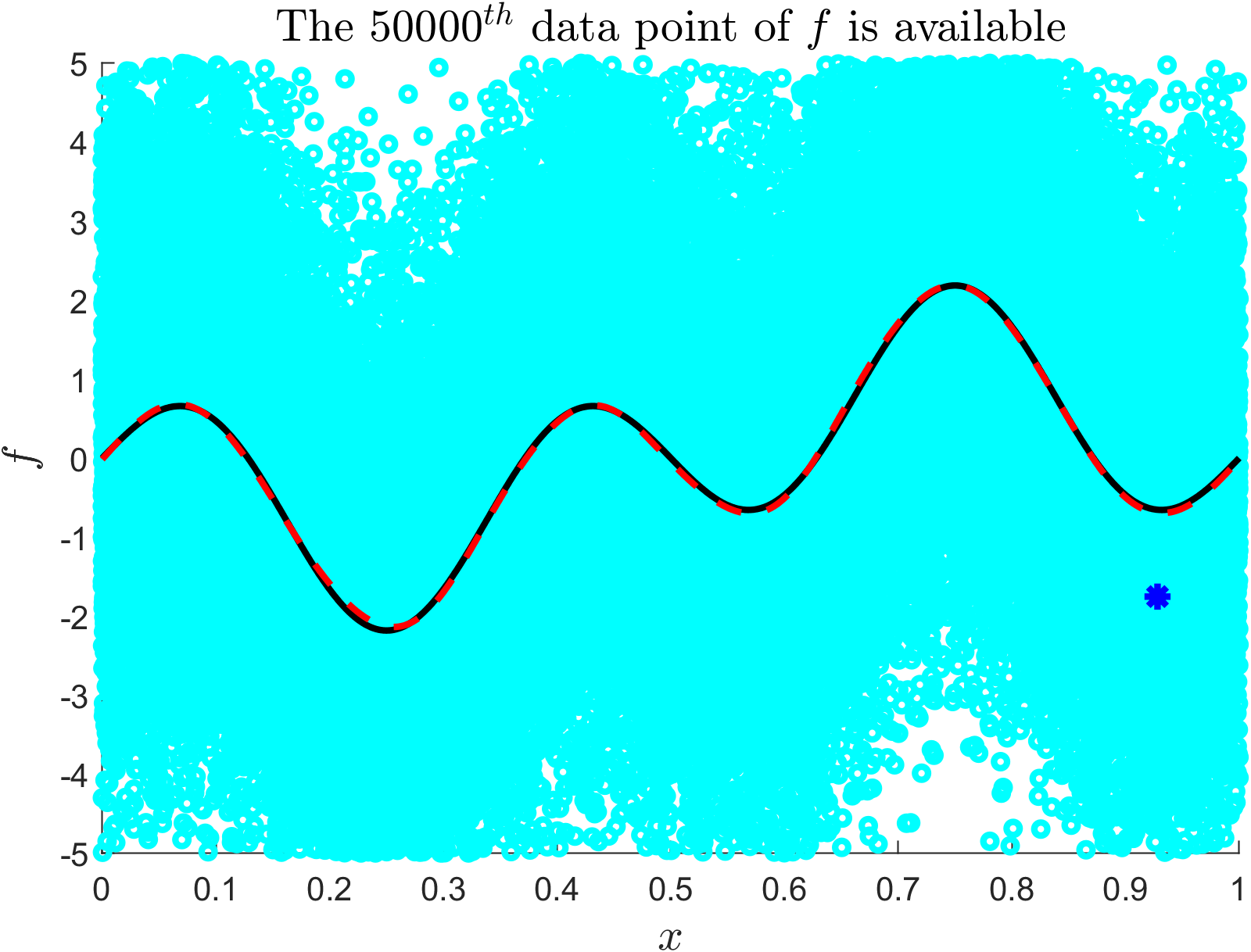}
        \caption{}
    \end{subfigure}
    \begin{subfigure}[b]{\textwidth}
        \centering
        \includegraphics[width = 0.3\textwidth]{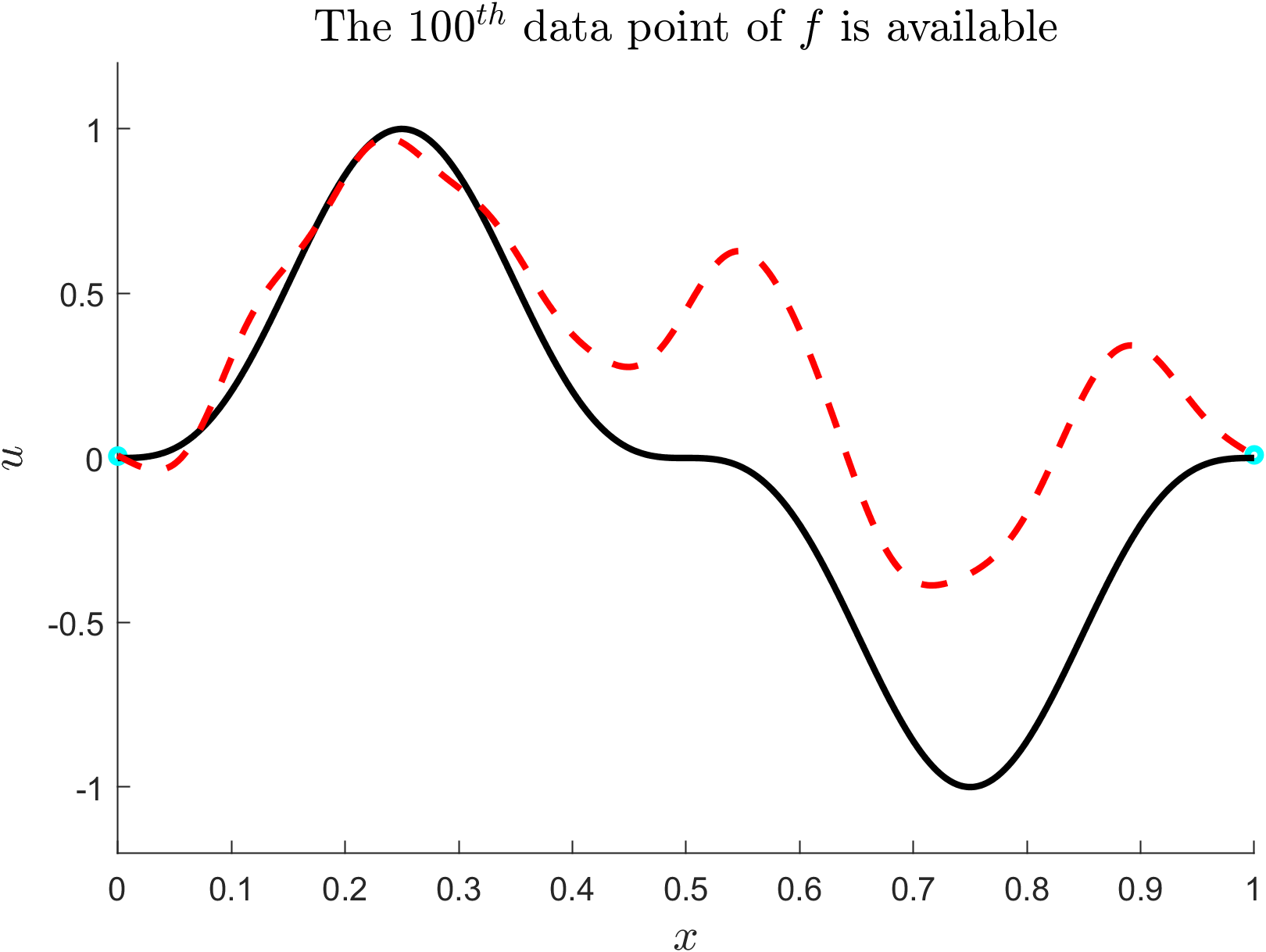}
       \includegraphics[width = 0.3\textwidth]{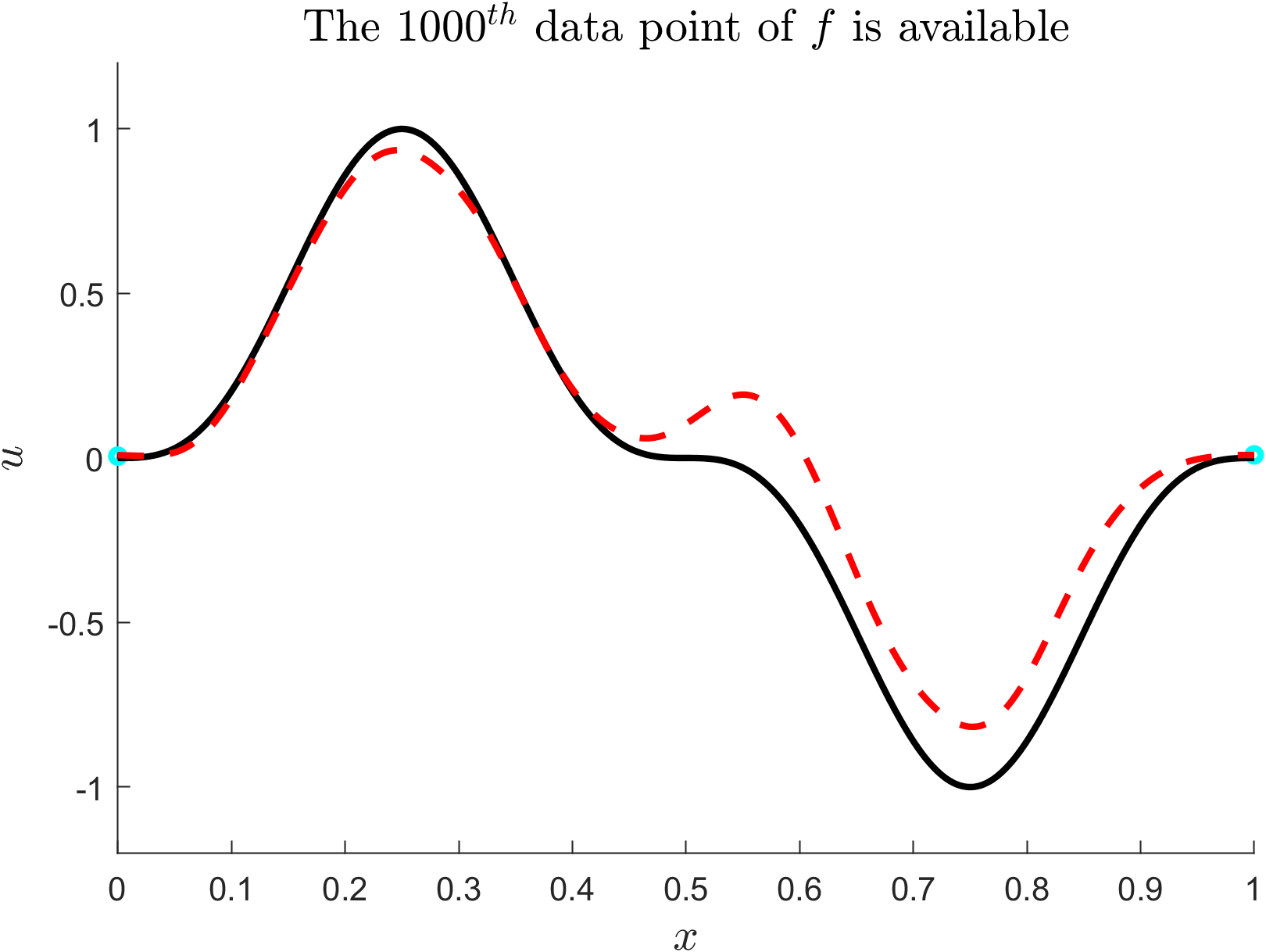}
       \includegraphics[width = 0.3\textwidth]{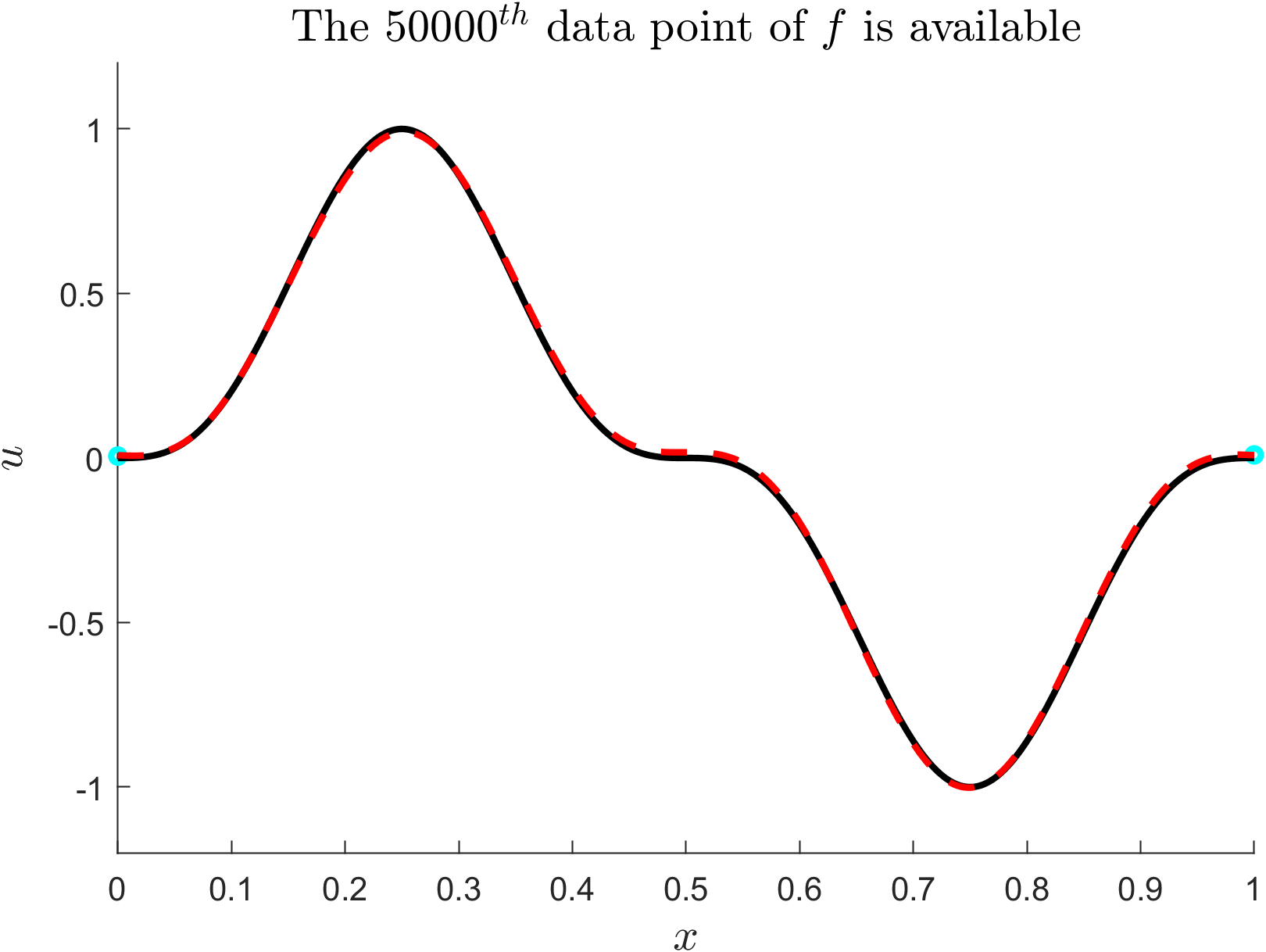}
        \caption{}
    \end{subfigure}
    \caption{Continual learning of the source term $f$ and solution $u$ to the 1D steady-state reaction-diffusion equation~\eqref{eq:reaction} our Riccati-based approach as more data becomes available. Measurements of the source term $f$ are corrupted by additive Gaussian noise with noise scale two. Our Riccati-based approach allows us to incrementally update the learned coefficients as more data becomes available without requiring access to the previous data or re-training on the entire dataset, which provides advantages in both memory and computations over conventional learning methods.}
    \label{fig:example_2:3}
\end{figure}

Next, we provide a large-scale continual learning example for solving the 1D steady-state reaction-diffusion equation~\eqref{eq:reaction}. Similar to Section \ref{subsec:example_1}, the measurements of the source term $f$ are corrupted by additive Gaussian noise with a relatively large noise scale. Hence, we need a large amount of data to solve~\eqref{eq:reaction} accurately. We again follow the continual learning setup and assume that the measurements of $f$ must be accessed in a stream, such that the historical data is not available after it is incorporated into our learned model. Specifically, each new data point of $f$ comes from uniformly randomly sampling a point $x\in[0, 1]$ and is corrupted by additive Gaussian noise with noise scale two. The same basis functions and loss function listed in Section~\ref{subsec:example_2} are used in this case. The results are presented in Figure \ref{fig:example_2:3} and Table \ref{tab:example_2:2}. We observe that the accuracy of our inferences of $u$ and $f$ is improved continuously as new measurements of $f$ are incorporated. This process does not require storage or use of historical data, which again highlights the computational and memory advantages of our Riccati-based approach.

\section{Details of the hyper-parameters in the numerical examples}\label{sec:details}
\updatetwo{In this section, we include additional details of the numerical examples from Section~\ref{sec:numerics}. Recall that RK4 is used to numerically solve the associated Riccati ODEs. The step size of RK4, denoted as $h$, is chosen based on three principles: high accuracy, efficient computations, and avoiding computational overflow. Unless otherwise stated, we use Python, the NumPy library \cite{harris2020array}, the TensorFlow library \cite{tensorflow2015-whitepaper}, and double precision in all of the numerical examples.}

\updatetwo{In the first example (Section~\ref{subsec:example_1}), $\lambda_i=1, \forall i$, $\MLreg_k=0.1, \forall k$, and the step size $h$ of RK4 is set to be $0.001, 0.0005, 0.0001$. A uniform grid with size $1001$ on $[0, 10]$ is used for evaluation (computing relative $L^2$  error). 
In the second example (Section~\ref{subsec:example_2}), we change $\lambda_b$ from $1$ to $10$ and the rest of the hyper-parameters remain the same as the regular training. The time step of RK4 is $h=0.0001$. A uniform grid with size $257$ on $[0, 1]$ is used for evaluation.
In the third example (Section~\ref{subsec:example_3}), the multi-head PINN used to obtain the basis functions has three hidden layers, each of which is equipped with hyperbolic tangent activation functions and $100$ neurons. The multi-head PINN is trained with the Adam optimizer \cite{kingma2014adam}, where the learning rate is $1\times 10^{-3}$ and the other hyperparameters are set to their default TensorFlow values. The training data for $f$ are evaluated on a $33\times 33$ uniform grid on $[0, 1]^2$ and the inference of $u$ is evaluated on a $257\times 257$ uniform grid.
The results of the regular training are obtained using $\lambda_i=1, \forall i$, $\MLreg_k=1, \forall k$, and $h=0.001$. When tuning $\MLreg$ and visualizing the Pareto front, we use the values of $\MLreg$ and the step size $h$ of RK4 listed in Table~\ref{tab:hyperparameter}. In the last example (Section~\ref{subsec:example_4}), we set $\lambda_i=1, \forall i$, $\MLreg_k=0.1, \forall k$, and $h=0.001$. In our implementation of the PDHG method, we set $\sigma_\weightvec = \sigma_{\pdhgdual} = 0.5$. The solution $x_1, x_2, x_3$ of the system identified is evaluated on a uniform grid of $10001$ on $[0, 10]$.}

\begin{table}[ht]
    \footnotesize
    \centering
    
    \begin{tabular}{|c|c|c|c|c|c|}
    \hline
    $\MLreg$ & $1\rightarrow 10^{-1}$ & $10^{-1}\rightarrow 10^{-2}$ & $10^{-2}\rightarrow 10^{-3}$ & $10^{-3}\rightarrow 10^{-4}$ & $10^{-4}\rightarrow 10^{-5}$ \\
    \hline
    $h$ & $10^{-2}$ & $10^{-3}$ & $10^{-4}$ & $10^{-6}$ & $10^{-7}$ \\
    \hline
    \end{tabular}
    \caption{\updatetwo{Step size $h$ of RK4 when tuning $\MLreg$ in the hyper-parameter tuning/Pareto front visualization example in Section~\ref{subsec:example_3}. $\MLreg_0 \rightarrow \MLreg_1$ represents the case when tuning $\MLreg$ from $\MLreg_0$ to $\MLreg_1$. }}
    \label{tab:hyperparameter}
\end{table}

%% file: main_sisc_template.bbl
\begin{thebibliography}{10}

\bibitem{tensorflow2015-whitepaper}
{\sc M.~Abadi, A.~Agarwal, P.~Barham, and et. al.}, {\em {TensorFlow}: Large-scale machine learning on heterogeneous systems}, 2015, \url{https://www.tensorflow.org/}.
\newblock Software available from tensorflow.org.

\bibitem{anderson2007optimal}
{\sc B.~D. Anderson and J.~B. Moore}, {\em Optimal control: linear quadratic methods}, Courier Corporation, 2007.

\bibitem{Bardi1997Optimal}
{\sc M.~Bardi and I.~Capuzzo-Dolcetta}, {\em Optimal control and viscosity solutions of {H}amilton-{J}acobi-{B}ellman equations}, Systems \& Control: Foundations \& Applications, Birkh\"{a}user Boston, Inc., Boston, MA, 1997, \url{https://doi.org/10.1007/978-0-8176-4755-1}, \url{https://doi.org/10.1007/978-0-8176-4755-1}.
\newblock With appendices by Maurizio Falcone and Pierpaolo Soravia.

\bibitem{bardi1984hopf}
{\sc M.~Bardi and L.~Evans}, {\em On {H}opf's formulas for solutions of {H}amilton-{J}acobi equations}, Nonlinear Analysis: Theory, Methods \& Applications, 8 (1984), pp.~1373 -- 1381, \url{https://doi.org/10.1016/0362-546X(84)90020-8}.

\bibitem{brunton2016discovering}
{\sc S.~L. Brunton, J.~L. Proctor, and J.~N. Kutz}, {\em Discovering governing equations from data by sparse identification of nonlinear dynamical systems}, {Proceedings of the National Academy of Sciences}, 113 (2016), pp.~3932--3937.

\bibitem{butcher2016numerical}
{\sc J.~C. Butcher}, {\em Numerical methods for ordinary differential equations}, John Wiley \& Sons, 2016.

\bibitem{chambolle2011pdhg}
{\sc A.~Chambolle and T.~Pock}, {\em {A first-order primal-dual algorithm for convex problems with applications to imaging}}, Journal of Mathematical Imaging and Vision, 40 (2011), pp.~120--145.

\bibitem{chen2021hopf}
{\sc P.~Chen, J.~Darbon, and T.~Meng}, {\em Hopf-type representation formulas and efficient algorithms for certain high-dimensional optimal control problems}, arXiv preprint arXiv:2110.02541,  (2021).

\bibitem{chen2021lax}
{\sc P.~Chen, J.~Darbon, and T.~Meng}, {\em Lax-oleinik-type formulas and efficient algorithms for certain high-dimensional optimal control problems}, arXiv preprint arXiv:2109.14849,  (2021).

\bibitem{chen2018neural}
{\sc R.~T. Chen, Y.~Rubanova, J.~Bettencourt, and D.~K. Duvenaud}, {\em Neural ordinary differential equations}, Advances in neural information processing systems, 31 (2018).

\bibitem{chung2020sampled}
{\sc J.~Chung, M.~Chung, J.~T. Slagel, and L.~Tenorio}, {\em Sampled limited memory methods for massive linear inverse problems}, Inverse Problems, 36 (2020), p.~054001.

\bibitem{darbon2015convex}
{\sc J.~Darbon}, {\em On convex finite-dimensional variational methods in imaging sciences and {H}amilton--{J}acobi equations}, SIAM Journal on Imaging Sciences, 8 (2015), pp.~2268--2293, \url{https://doi.org/10.1137/130944163}, \url{https://arxiv.org/abs/https://doi.org/10.1137/130944163}.

\bibitem{Darbon2023Neural}
{\sc J.~Darbon, P.~M. Dower, and T.~Meng}, {\em Neural network architectures using min-plus algebra for solving certain high-dimensional optimal control problems and {H}amilton-{J}acobi {PDE}s}, Math. Control Signals Systems, 35 (2023), pp.~1--44, \url{https://doi.org/10.1007/s00498-022-00333-2}, \url{https://doi.org/10.1007/s00498-022-00333-2}.

\bibitem{langlois2021HJvariational}
{\sc J.~Darbon and G.~Langlois}, {\em {On Bayesian Posterior Mean Estimators in Imaging Sciences and Hamilton–Jacobi Partial Differential Equations}}, J. Math Imaging Vis., 63 (2021), p.~821–854.

\bibitem{darbon2019decomposition}
{\sc J.~Darbon and T.~Meng}, {\em On decomposition models in imaging sciences and multi-time {H}amilton--{J}acobi partial differential equations}, SIAM Journal on Imaging Sciences, 13 (2020), pp.~971--1014, \url{https://doi.org/10.1137/19M1266332}, \url{https://doi.org/10.1137/19M1266332}, \url{https://arxiv.org/abs/https://doi.org/10.1137/19M1266332}.

\bibitem{darbon2022imagedenoising}
{\sc J.~Darbon, T.~Meng, and E.~Resmerita}, {\em {On Hamilton–Jacobi PDEs and Image Denoising Models with Certain Nonadditive Noise}}, Journal of Mathematical Imaging and Vision, 64 (2022), p.~408–441.

\bibitem{Darbon2016Algorithms}
{\sc J.~Darbon and S.~Osher}, {\em Algorithms for overcoming the curse of dimensionality for certain {H}amilton-{J}acobi equations arising in control theory and elsewhere}, Res Math Sci Research in the Mathematical Sciences, 3 (2016), pp.~1--26, \url{https://doi.org/10.1186/s40687-016-0068-7}, \url{https://doi.org/10.1186/s40687-016-0068-7}.

\bibitem{desai2021one}
{\sc S.~Desai, M.~Mattheakis, H.~Joy, P.~Protopapas, and S.~Roberts}, {\em One-shot transfer learning of physics-informed neural networks}, arXiv preprint arXiv:2110.11286,  (2021).

\bibitem{evans1984differentialgames}
{\sc L.~C. Evans and P.~E. Souganidis}, {\em {Differential games and representation formulas for solutions of Hamilton-Jacobi-Isaacs equations}}, Indiana University mathematics journal, 33 (1984), pp.~773--797.

\bibitem{goodfellow2016deep}
{\sc I.~Goodfellow, Y.~Bengio, and A.~Courville}, {\em Deep learning}, MIT press, 2016.

\bibitem{goswami2022deep}
{\sc S.~Goswami, K.~Kontolati, M.~D. Shields, and G.~E. Karniadakis}, {\em Deep transfer operator learning for partial differential equations under conditional shift}, Nature Machine Intelligence,  (2022), pp.~1--10.

\bibitem{han2018solving}
{\sc J.~Han, A.~Jentzen, and W.~E}, {\em Solving high-dimensional partial differential equations using deep learning}, {Proceedings of the National Academy of Sciences}, 115 (2018), pp.~8505--8510.

\bibitem{harris2020array}
{\sc C.~R. Harris, K.~J. Millman, S.~J. Van Der~Walt, R.~Gommers, P.~Virtanen, D.~Cournapeau, E.~Wieser, J.~Taylor, S.~Berg, N.~J. Smith, et~al.}, {\em Array programming with {NumPy}}, Nature, 585 (2020), pp.~357--362.

\bibitem{hopf1965hopfformula}
{\sc E.~Hopf}, {\em {Generalized Solutions of non-linear Equations of First Order}}, Journal of Mathematics and Mechanics, 14 (1965), pp.~951--973.

\bibitem{Jin2008pareto}
{\sc Y.~Jin and B.~Sendhoff}, {\em Pareto-based multiobjective machine learning: An overview and case studies}, IEEE Transactions on Systems, Man, and Cybernetics, Part C (Applications and Reviews), 38 (2008), pp.~397--415, \url{https://doi.org/10.1109/TSMCC.2008.919172}.

\bibitem{kairouz2021advances}
{\sc P.~Kairouz, H.~B. McMahan, B.~Avent, A.~Bellet, M.~Bennis, A.~N. Bhagoji, K.~Bonawitz, Z.~Charles, G.~Cormode, R.~Cummings, et~al.}, {\em Advances and open problems in federated learning}, Foundations and Trends{\textregistered} in Machine Learning, 14 (2021), pp.~1--210.

\bibitem{kingma2014adam}
{\sc D.~P. Kingma and J.~Ba}, {\em Adam: A method for stochastic optimization}, arXiv preprint arXiv:1412.6980,  (2014).

\bibitem{kirkpatrick2017overcoming}
{\sc J.~Kirkpatrick, R.~Pascanu, N.~Rabinowitz, J.~Veness, G.~Desjardins, A.~A. Rusu, K.~Milan, J.~Quan, T.~Ramalho, A.~Grabska-Barwinska, et~al.}, {\em Overcoming catastrophic forgetting in neural networks}, {Proceedings of the National Academy of Sciences}, 114 (2017), pp.~3521--3526.

\bibitem{lecun2015deep}
{\sc Y.~LeCun, Y.~Bengio, and G.~Hinton}, {\em Deep learning}, Nature, 521 (2015), pp.~436--444.

\bibitem{li2020federated}
{\sc T.~Li, A.~K. Sahu, A.~Talwalkar, and V.~Smith}, {\em Federated learning: Challenges, methods, and future directions}, IEEE Signal Processing Magazine, 37 (2020), pp.~50--60.

\bibitem{liavas1998numerical}
{\sc A.~P. Liavas and P.~A. Regalia}, {\em Numerical stability issues of the conventional recursive least squares algorithm}, in Proceedings of the 1998 IEEE International Conference on Acoustics, Speech and Signal Processing, ICASSP'98 (Cat. No. 98CH36181), vol.~3, IEEE, 1998, pp.~1409--1412.

\bibitem{liavas1999numerical}
{\sc A.~P. Liavas and P.~A. Regalia}, {\em On the numerical stability and accuracy of the conventional recursive least squares algorithm}, IEEE Transactions on Signal Processing, 47 (1999), pp.~88--96.

\bibitem{lions1986hopf}
{\sc P.~L. Lions and J.-C. Rochet}, {\em {Hopf Formula and Multitime {H}amilton-{J}acobi Equations}}, Proceedings of the American Mathematical Society, 96 (1986), pp.~79--84, \url{http://www.jstor.org/stable/2045657}.

\bibitem{ljung1985error}
{\sc S.~Ljung and L.~Ljung}, {\em Error propagation properties of recursive least-squares adaptation algorithms}, Automatica, 21 (1985), pp.~157--167.

\bibitem{MATLAB}
{\sc {MathWorks Inc.}}, {\em Matlab version: 9.13.0 (r2022b)}, 2022, \url{https://www.mathworks.com}.

\bibitem{mceneaney2006max}
{\sc W.~McEneaney}, {\em Max-plus methods for nonlinear control and estimation}, Springer Science \& Business Media, 2006.

\bibitem{mohri2018foundations}
{\sc M.~Mohri, A.~Rostamizadeh, and A.~Talwalkar}, {\em Foundations of machine learning}, MIT press, 2018.

\bibitem{Nakamura2021QRnet}
{\sc T.~Nakamura-Zimmerer, Q.~Gong, and W.~Kang}, {\em Q{R}net: optimal regulator design with {LQR}-augmented neural networks}, IEEE Control Syst. Lett., 5 (2021), pp.~1303--1308.

\bibitem{parisi2019continual}
{\sc G.~I. Parisi, R.~Kemker, J.~L. Part, C.~Kanan, and S.~Wermter}, {\em Continual lifelong learning with neural networks: A review}, Neural networks, 113 (2019), pp.~54--71.

\bibitem{psaros2023uncertainty}
{\sc A.~F. Psaros, X.~Meng, Z.~Zou, L.~Guo, and G.~E. Karniadakis}, {\em Uncertainty quantification in scientific machine learning: Methods, metrics, and comparisons}, Journal of Computational Physics,  (2023), p.~111902.

\bibitem{raissi2019physics}
{\sc M.~Raissi, P.~Perdikaris, and G.~Karniadakis}, {\em Physics-informed neural networks: A deep learning framework for solving forward and inverse problems involving nonlinear partial differential equations}, Journal of Computational Physics, 378 (2019), pp.~686--707.

\bibitem{rochet1985multitimeHJ}
{\sc J.-C. Rochet}, {\em {The taxation principle and multi-time {H}amilton-{J}acobi equations}}, Journal of Mathematical Economics, 14 (1985), pp.~113--128.

\bibitem{russell2010artificial}
{\sc S.~J. Russell}, {\em Artificial intelligence: A modern approach}, Pearson Education, Inc., 2010.

\bibitem{sirignano2018dgm}
{\sc J.~Sirignano and K.~Spiliopoulos}, {\em {DGM}: A deep learning algorithm for solving partial differential equations}, Journal of Computational Physics, 375 (2018), pp.~1339--1364.

\bibitem{todorov2008general}
{\sc E.~Todorov}, {\em General duality between optimal control and estimation}, in 2008 47th IEEE Conference on Decision and Control, IEEE, 2008, pp.~4286--4292.

\bibitem{TrentelmanControlLinearSystems}
{\sc H.~L. Trentelman, A.~A. Stoorvogel, and M.~Hautus}, {\em Control {T}heory for {L}inear {S}ystems}, Springer-Verlag London, 2001.

\bibitem{van2019three}
{\sc G.~M. Van~de Ven and A.~S. Tolias}, {\em Three scenarios for continual learning}, arXiv preprint arXiv:1904.07734,  (2019).

\bibitem{wan2006multi}
{\sc X.~Wan and G.~E. Karniadakis}, {\em Multi-element generalized polynomial chaos for arbitrary probability measures}, SIAM Journal on Scientific Computing, 28 (2006), pp.~901--928.

\bibitem{weisberg2005applied}
{\sc S.~Weisberg}, {\em Applied linear regression}, vol.~528, John Wiley \& Sons, 2005.

\bibitem{yegorov2017perspectives}
{\sc I.~Yegorov and P.~M. Dower}, {\em Perspectives on characteristics based curse-of-dimensionality-free numerical approaches for solving {H}amilton--{J}acobi equations}, Applied Mathematics \& Optimization,  (2017), pp.~1--49.

\bibitem{zhang2023discovering}
{\sc Z.~Zhang, Z.~Zou, E.~Kuhl, and G.~E. Karniadakis}, {\em Discovering a reaction-diffusion model for {Alzheimer's} disease by combining {PINNs} with symbolic regression}, arXiv preprint arXiv:2307.08107,  (2023).

\bibitem{zou2023hydra}
{\sc Z.~Zou and G.~E. Karniadakis}, {\em {L-HYDRA: Multi-Head Physics-Informed Neural Networks}}, arXiv preprint arXiv:2301.02152,  (2023).

\bibitem{zou2022neuraluq}
{\sc Z.~Zou, X.~Meng, A.~F. Psaros, and G.~E. Karniadakis}, {\em {NeuralUQ}: A comprehensive library for uncertainty quantification in neural differential equations and operators}, arXiv preprint arXiv:2208.11866,  (2022).

\end{thebibliography}
